\documentclass[nohyperref]{article}

\usepackage{microtype}
\usepackage{graphicx}
\usepackage{booktabs} %
\usepackage{url}
\usepackage{wrapfig}
\usepackage{multirow}
\usepackage{enumitem}

\usepackage{amsmath,amsfonts,bm}
\usepackage{bbm}
\usepackage{amsthm}

\DeclareMathOperator\supp{supp}

\def\eqref#1{equation~\ref{#1}}

\def\1{\bm{1}}

\DeclareMathAlphabet{\mathsfit}{\encodingdefault}{\sfdefault}{m}{sl}
\SetMathAlphabet{\mathsfit}{bold}{\encodingdefault}{\sfdefault}{bx}{n}

\def\gA{{\mathcal{A}}}

\def\gO{{\mathcal{O}}}

\def\gS{{\mathcal{S}}}

\newcommand{\E}[2]{\mathbb{E}_{#1}{\left[#2\right]}} %

\newcommand{\R}{\mathbb{R}}

\DeclareMathOperator*{\argmax}{arg\,max}

\renewcommand{\cite}{\citep}

\newcommand{\defeq}{\mathrel{\mathop:}=}

\usepackage{xspace}
\makeatletter
\DeclareRobustCommand\onedot{\futurelet\@let@token\@onedot}
\def\@onedot{\ifx\@let@token.\else.\null\fi\xspace}

\def\eg{\emph{e.g}\onedot} 
\def\ie{\emph{i.e}\onedot}

\def\wrt{w.r.t\onedot} 

\makeatother

\usepackage{pifont}%
\newcommand{\cmark}{\ding{51}}%
\newcommand{\xmark}{\ding{55}}%

\usepackage{array}
\newcolumntype{P}[1]{>{\centering\arraybackslash}p{#1}}

\usepackage[pagebackref,colorlinks=true,citecolor=blue]{hyperref}

\usepackage[accepted]{icml2022}

\usepackage{amsmath}
\usepackage{amssymb}
\usepackage{mathtools}
\usepackage{amsthm}

\usepackage{nicefrac}
\usepackage{changepage}
\usepackage{caption}
\usepackage{subcaption}

\usepackage[capitalize,noabbrev]{cleveref}

\usepackage[textsize=tiny]{todonotes}

\icmltitlerunning{Recurrent Model-Free RL Can Be a Strong Baseline for Many POMDPs}

\begin{document}

\twocolumn[
\icmltitle{Recurrent Model-Free RL Can Be a Strong Baseline for Many POMDPs}

\icmlsetsymbol{equal}{*}

\begin{icmlauthorlist}
\icmlauthor{Tianwei Ni}{mila}
\icmlauthor{Benjamin Eysenbach}{cmu}
\icmlauthor{Ruslan Salakhutdinov}{cmu}
\\
\url{https://github.com/twni2016/pomdp-baselines}
\end{icmlauthorlist}

\icmlaffiliation{mila}{Université de Montréal \& Mila -- Quebec AI Institute}
\icmlaffiliation{cmu}{Carnegie Mellon University}

\icmlcorrespondingauthor{Tianwei Ni}{tianwei.ni@mila.quebec}
\icmlcorrespondingauthor{Benjamin Eysenbach}{beysenba@cs.cmu.edu}

\icmlkeywords{reinforcement learning, baseline, POMDP, recurrent neural networks, LSTM, GRU, RNN, off-policy, TD3, SAC, meta RL, robust RL, generalization, temporal credit assignment}

\vskip 0.3in
]

\printAffiliationsAndNotice{Work was primarily done when TN was at Carnegie Mellon University.}  %

\newcommand{\codelink}{https://drive.google.com/drive/folders/1I5mLlKPf2Gmdpm0nzy9OkR494nCJll1g?usp=sharing}

\begin{abstract}

Many problems in RL, such as meta-RL, robust RL, generalization in RL, and temporal credit assignment, can be cast as POMDPs. In theory, simply augmenting model-free RL with memory-based architectures, such as recurrent neural networks, provides a general approach to solving all types of POMDPs. However, prior work has found that such recurrent model-free RL methods tend to perform worse than more specialized algorithms that are designed for specific types of POMDPs. This paper revisits this claim. We find that careful architecture and hyperparameter decisions can often yield a recurrent model-free implementation that performs on par with (and occasionally substantially better than) more sophisticated recent techniques. We compare to 21 environments from 6 prior specialized methods and find that our implementation achieves greater sample efficiency and asymptotic performance than these methods on \nicefrac{18}{21} environments. We also release a simple and efficient implementation of recurrent model-free RL for future work to use as a baseline for POMDPs.

\end{abstract}
\vspace{-8mm}

\section{Introduction}

\begin{figure}[h]
    \centering
    \vspace{-1.2mm}
    \includegraphics[width=\linewidth]{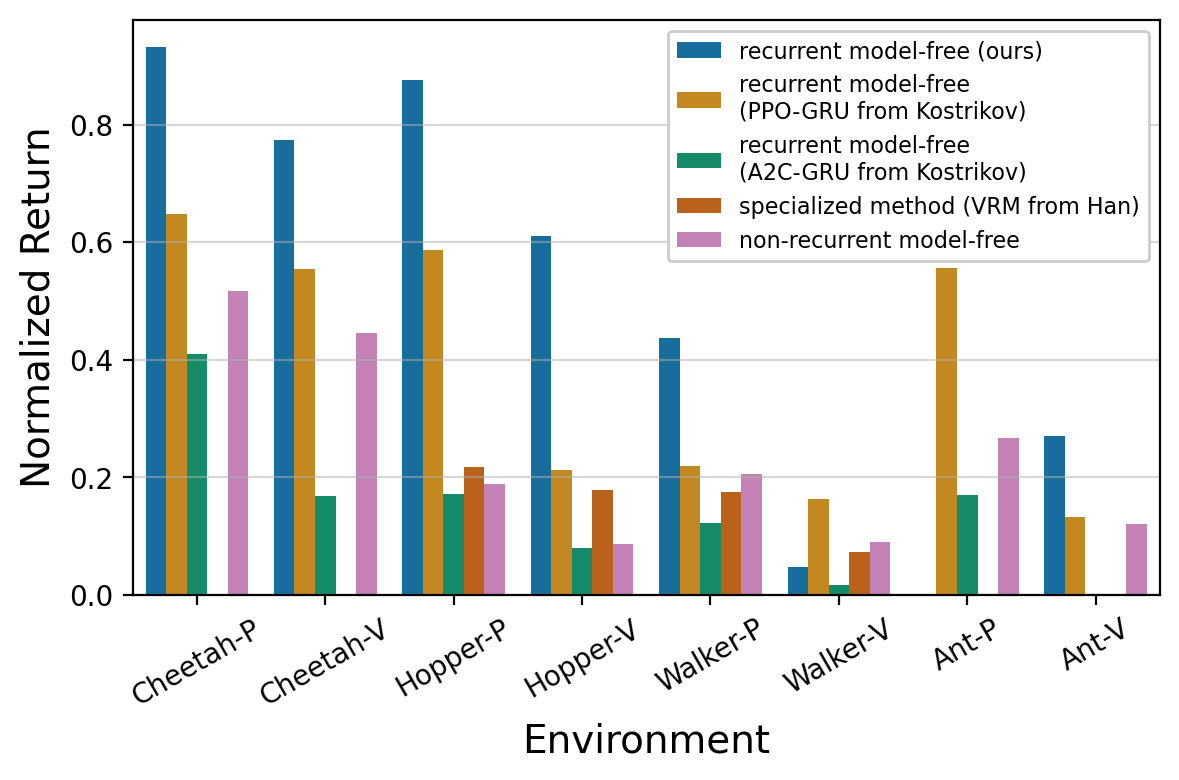}
    \vspace{-8mm}
    \caption{\footnotesize \textbf{The importance of implementation for recurrent model-free RL.}
    This paper identifies important design decisions for recurrent model-free RL. Our implementation outperforms prior implementations (\eg PPO-GRU and A2C-GRU from~\citet{kostrikov2018pytorch}) and purpose-designed methods (\eg VRM from~\citet{han2019variational}) on their respective POMDP benchmarks.}
    \label{fig:trailer}
    \vspace{-6mm}
\end{figure}

Reinforcement learning (RL) is typically cast as a problem of learning a single fully observable task (an MDP), training and testing on that same task. However, most real-world applications of RL demand some degree of transfer and handling of partial observability. For example, visual navigation~\cite{zhu2017target} requires that robots adapt to unseen scenes with occlusion in observations, and human-robot collaboration requires that robots infer the intentions of human collaborators~\cite{chen2018planning}.

Many subareas in RL study problems that are special cases of POMDPs (see Table~\ref{tab:problems}).
For example, meta-RL~\citep{schmidhuber1987evolutionary, thrun2012learning, duan2016rl,wang2016learning}  
is a POMDP where certain aspects of the reward function or (less commonly) dynamics function are unobserved but held constant through one episode. 
The robust RL problem~\citep{bagnell2001solving, rajeswaran2016epopt, pinto2017robust,pattanaik2017robust}
assumes that certain aspects of the dynamics or reward function are unknown, and aims to find optimal policies that perform well against adversarially-chosen perturbations. 
Generalization in RL~\citep{whiteson2011protecting,zhang2018dissection,packer2018assessing,cobbe2019quantifying} 
focuses on unobserved aspects of the dynamics or reward function that are novel during testing, using an average-case objective instead of the worst-case objective of robust RL.
Temporal credit assignment~\citep{sutton1984temporal,arjona2019rudder,hung2019optimizing,ren2021learning} assumes that the reward function is history-dependent and aims to learn to assign credits of current actions to future rewards.  

Recent work has proposed efficient and performant algorithms to solve each of these \emph{specialized} problem settings. However, these algorithms often make assumptions that preclude their application to other POMDPs. For example, methods for robust RL are rarely used for generalization in RL due to objective mismatch (average-case versus worst-case). Similarly, methods for meta-RL are rarely used for temporal credit assignment due to the stationarity assumption in meta-RL.

One method that is applicable to any POMDP is model-free RL equipped with a recurrent policy (actor) and (sometimes) recurrent value function~\cite{duan2016rl,wang2016learning,packer2018assessing, igl2018deep, rakelly2019efficient, fakoor2019meta, yu2020meta}. 
We will refer to this approach as \textbf{recurrent model-free RL}. 
This baseline is simultaneously \textit{simple}, as it requires changing only a few lines of code from a model-free RL algorithm, and \textit{general}, as RNNs~\citep{elman1990finding} are Turing-complete~\citep{siegelmann1995computational} and universal function approximators~\citep{schafer2006recurrent}.

This approach has been used as a baseline in many prior works, but these prior works report that it performs poorly in many problem settings, including 
meta-RL~\citep{rakelly2019efficient,zintgraf2019varibad}, 
general POMDPs~\citep{igl2018deep,han2019variational}, 
robust RL~\citep{zhang2021robust}, 
generalization in RL~\citep{packer2018assessing},
and temporal credit assignment~\citep{arjona2019rudder,raposo2021synthetic}.
Why does recurrent model-free RL perform poorly? 
One common explanation is that specialized algorithms
or more complicated memory architectures~\citep{ritter2018been,parisotto2020stabilizing} (implicitly) encode inductive biases to solve these specific tasks. 
For example, algorithms for meta-RL may assume that the underlying dynamics (while unknown) are fixed, and the underlying goals are fixed within one episode~\citep{rakelly2019efficient, zintgraf2019varibad}. 
Similarly, algorithms for robust RL may assume that the dynamics parameters are known~\citep{rajeswaran2016epopt} and dynamics is Lipschitz continuous~\cite{jiang2021monotonic}.
Algorithms for temporal credit assignment sometimes assume that the history-dependent reward can be decomposed into a sum of Markovian rewards~\citep{ren2021learning}.

This paper challenges the claim that recurrent model-free RL performs poorly. We argue that, contrary to popular belief, recurrent model-free RL can be competitive with recent state-of-the-art algorithms across a range of different POMDP settings.
Similar to the spirit in prior work in Markovian PPO~\cite{engstrom2019implementation, andrychowicz2020matters} and recurrent DQN~\cite{kapturowski2018recurrent}, our experiments show that the implementation of recurrent model-free RL matters.
Through extensive experiments (\eg, Fig.~\ref{fig:trailer} shows results on an occluded locomotion benchmark), we show that the careful design and implementation of recurrent model-free RL is critical to its performance. Design decisions such as the actor-critic architecture, the underlying model-free RL algorithm, and context length in RNNs are especially important.

The main contribution of this paper is a performant implementation of recurrent-model free RL. 
We demonstrate that simple yet important design decisions, such as the underlying RL algorithm and the context length, can often yield a recurrent model-free RL algorithm that performs (at least) on par with prior specialized POMDP algorithms \emph{on the benchmarks those algorithms were designed to solve.} 
Ablation experiments identify the importance of these design decisions. We have released the code that is easy to use and memory-efficient.

\vspace{-3mm}
\section{Background}

\label{sec:prelim}

\vspace{-1mm}
\paragraph{MDP.} A Markov decision process (MDP)~\cite{bellman1957markovian} is a tuple $(\gS, \gA, T, T_0, R,  H, \gamma) $ , where $\gS$ is the set of states, $\gA$ is the set of actions, $T: \gS \times \gA \times \gS \to [0,1]$ is the transition function (dynamics), $T_0:  \gS  \to [0,1] $ is the initial state distribution, $R: \gS \times \gA \times \gS \to \mathbb R $ is the reward function, $H \in \mathbb N$ is the time horizon, and $\gamma \in [0,1)$ is the discount factor. 
Solving an MDP requires learning a policy $\pi: \gS \times \gA \to [0,1] $ that maximizes the expected discounted return: $\pi^* = \argmax_\pi \E{s_t,a_t,r_t\sim T,\pi}{\sum_{t=0}^{H-1} \gamma^t r_{t+1}\mid s_0}$.
For any MDP, there exists an optimal policy that is memoryless~\citep{puterman2014markov}. 
MaxEnt RL algorithms~\cite{ziebart2010thesis,haarnoja2018soft} add an entropy bonus to the RL objective.

\vspace{-3mm}
\paragraph{POMDP.}
A partially observable Markov decision process (POMDP)~\cite{aastrom1965optimal} %
is a tuple $(\mathcal S, \mathcal A, \mathcal O, T, T_0, O, O_0, R, H, \gamma)$, where the underlying process is an MDP $(\gS, \gA, T, T_0, R,  H, \gamma)$. Let $\gO$ be the set of observations and let $O: \gS \times \gA \times \gO \to [0,1]$ be the emission function.
Let the observable trajectory up to time-step $t$ be  $\tau_{0:t} = (o_0, a_0, o_1, r_1,  \dots, a_{t-1},o_t, r_t)$, the memory-based policy in \textit{the most general} form is defined as $\pi(a_t \mid \tau_{0:t})$, conditioning on the whole history.
At the first time step $t=0$, an initial state $s_0\sim T_0(\cdot)$ and an initial observation $o_0\sim O_0(\cdot\mid s_0)$ are sampled. 
At any time-step $t \in \{0,\dots, H-1\}$, the policy emits the action $a_t \in \gA$ to the system, the system updates the state following the dynamics, $s_{t+1} \sim T(\cdot \mid s_{t},a_{t})$, the next observation is sampled $o_{t+1} \sim O(\cdot\mid s_{t+1}, a_t)$ and the reward is computed as $r_{t+1} = R(s_{t}, a_t, s_{t+1})$.

We refer to the part of the state $s_t$ at current time-step $t$ that can be directly unveiled from \textit{current} observation $o_t$ as the \emph{observable state} $s^o_t$, and the rest part of the state as the \emph{hidden state} $s^h_t$. We call the hidden state $s^h_t$ \textit{stationary} if it does not change within an episode.
The average-case and worst-case objectives for POMDPs can be written as:
\vspace{-1mm}
\begin{align*}
& \max_\pi \E{s^h\sim T_0}{\E{\tau}{\sum_{t=0}^{H-1} \gamma^t r_{t+1}\mid s^h}} \tag{average-case}\\
& \max_\pi \min_{s^h \in \supp(T_0)}\E{\tau}{\sum_{t=0}^{H-1} \gamma^t r_{t+1}\mid s^h} \tag{worst-case}
\end{align*}

\section{Related Work}
\label{sec:related_work}

\begin{table*}[t]
    \footnotesize
    \centering
    \caption{{\footnotesize \textbf{Summary of selected POMDP subareas.} For each subarea, we indicate whether the hidden state $s^h$ determines the dynamics or the reward function, and whether it changes within an episode. We indicate the typical inputs to the agent: \texttt{o}bservations, \texttt{a}ctions, \texttt{r}ewards, and \texttt{d}one signals. We indicate whether the subarea uses the average-case or worst-case objective, and whether the evaluation typically includes domain shift. A ``*'' indicates that some prior work violates the trend.
    }}
\begingroup
\renewcommand{\arraystretch}{1.5} %
    \begin{tabular}{p{0.16\textwidth}|p{0.12\textwidth}|p{0.1\textwidth}|p{0.12\textwidth}|p{0.1\textwidth}|p{0.10\textwidth}|p{0.11\textwidth}}
    \toprule
      Subarea & $s^h$ in dynamics? &  $s^h$ in reward?  & Is  $s^h$ stationary? & Agent input  & RL objective & Domain shift? \\
      \midrule
      ``Standard" POMDP & \cmark & \cmark & \xmark & \texttt{oar} & Avg & \xmark \\ \hline
      Meta-RL & \xmark* & \cmark & \cmark & \texttt{oard} &  Avg & \xmark \\  \hline 
        Robust RL & \cmark* & \xmark* & \cmark* & \texttt{oa} & Worst & \xmark \\  \hline
      Generalization in RL & \cmark* & \xmark* & \cmark* & \texttt{oa} & Avg & \cmark* \\   \hline   
      Temporal credit assignment & \xmark & \cmark & \xmark & \texttt{oa} & Avg & \xmark \\ 
      \bottomrule
    \end{tabular}
\endgroup
    \label{tab:problems}
\end{table*}
\vspace{1mm}

We discuss subareas of RL that explicitly or implicitly solve POMDPs, as well as algorithms proposed for these specialized settings. Table~\ref{tab:problems} summarizes these subareas.  

\paragraph{RL for ``standard" POMDPs.} We use the term ``standard'' to refer to prior work that explicitly labels the problems studied as POMDPs. Common tasks include scenarios where the states are partially occluded~\cite{heess2015memory}, different states correspond to the same observation (perceptual aliasing~\cite{whitehead1990active}),  random frames are dropped~\cite{hausknecht2015deep}, observations use egocentric images~\cite{zhu2017target}, or the observations are perturbed with random noise~\cite{meng2021memory}. 
These POMDPs often have hidden states that are non-stationary and affect both the rewards and the dynamics.
POMDPs are hard to solve
because of the curse of dimensionality: the size of the history grows linearly with the horizon length~\cite{papadimitriou1987complexity, littman1996algorithms}. Prior POMDP algorithms~\cite{cassandra1994acting}
attempt to infer the state from the past sequence of observations, and then apply standard RL techniques to that inferred state. Such an inferred state is known as a belief state~\citep{kaelbling1998planning}.
However, exact inference requires the knowledge of the dynamics, emission probabilities, and reward functions, and is intractable in all except the most simple settings.
One strategy for solving these general POMDPs is to use memory-based policies, which take the entire history of past observations as inputs.
Among the memory architectures, RNNs have been widely used to equip RL algorithms~\cite{schmidhuber1991reinforcement,bakker2001reinforcement,wierstra2007solving}, as they have a simpler design without losing the expressivity~\citep{schafer2006recurrent},
compared to more complicated ones, \eg external memory~\citep{graves2016hybrid,oh2016control} and episodic memory~\citep{fortunato2019generalization,zhu2020episodic}.
These recurrent RL strategies can be further subdivided into model-free methods~\cite{heess2015memory, hausknecht2015deep,mirowski2016learning,meng2021memory}, where the single objective is to maximize the return, and model-based methods~\cite{watter2015embed,ha2018recurrent,igl2018deep,Zhang2018SOLARDS,espeholt2018impala,Gregor2019ShapingBS,Hafner2019LearningLD,han2019variational,lee2019stochastic}.%
that have explicitly inferred the belief state and pass it as an additional input to a memoryless policy.
The recurrent model-free RL that we focus on belongs to the class of model-free, off-policy, memory-based algorithms.

\paragraph{Meta-RL.} Meta-RL, also called ``learning to learn"~\cite{schmidhuber1987evolutionary, thrun2012learning}, focuses on POMDPs where some parameters in the rewards or (less commonly) dynamics are varied from episode to episode, but remain fixed within a single episode~\cite{humplik2019meta}. These different values of these parameters represent different tasks. %
The meta-RL setting is almost the same as multi-task RL~\cite{wilson2007multi, yu2020meta}, but differs in that multi-task RL can observe the task parameters, making it an MDP instead of a POMDP. 
Algorithms for meta-RL can be roughly categorized based on how the adaptation step is performed. Gradient-based algorithms~\cite{hochreiter2001learning,finn2017model, fakoor2019meta}
perform adaptation by running a few gradient steps on the pre-trained models. 
Memory or context-based algorithms use memory architectures to implicitly adapt. These memory-based methods which can be further subdivided into implicit and explicit task inference methods.
Implicit task inference methods~\cite{wang2016learning,duan2016rl,ritter2018been,espeholt2018impala,parisotto2020stabilizing} 
use an RL objective only to learn memory-based policies. %
Explicit task inference methods \cite{zintgraf2019varibad,rakelly2019efficient} train an extra inference model to estimate task embeddings (\ie, a representation of the unobserved parameters) by approximate inference. Task embeddings are then used as additional inputs to memoryless policies.

\paragraph{Robust RL.} 
The goal of robust RL is to find a policy that maximizes returns in the worst-case environments.
Prior work designs deep RL algorithms that are robust against perturbations to the dynamics~\cite{khalil1996robust,bagnell2001solving, nilim2005robust, morimoto2005robust, derman2018soft,rajeswaran2016epopt,mankowitz2019robust,jiang2021monotonic}, %
observations~\cite{lin2017tactics,pattanaik2017robust, huang2017adversarial, wang2019robust}, %
and actions~\cite{pinto2017robust, gleave2019adversarial, tessler2019action}. %
Treating the robust RL problem as a POMDP, rather than an MDP (as done in most prior work), unlocks a key capability: using memory to identify the hidden state within a single episode.
While some work find memory-based policies are more robust to adversarial attacks than Markovian policies~\cite{russo2019optimal, zhang2021robust}, they train these baselines in a single MDP without adversaries. In contrast, we will train recurrent model-free RL on a distribution of MDPs. %

\paragraph{Generalization in RL.} 
The goal of generalization in RL is to make RL algorithms perform well in test domains that are unseen during training. This setting differs from robust RL because it uses an average-case objective instead of a worst-case objective. In this sense, meta-RL is closely related to (in-distribution) generalization in RL.
Prior work has studied generalization to
initial states in the same MDP~\cite{whiteson2011protecting, rajeswaran2017towards, zhang2018study}, %
to random disturbance in dynamics~\cite{rajeswaran2017towards}, states~\cite{stulp2011learning}, %
observations~\cite{zhang2018dissection, song2019observational}, %
and actions~\cite{srouji2018structured}, %
and to different modes in procedurally generated games~\cite{justesen2018illuminating, farebrother2018generalization, cobbe2019quantifying}. %
Among them, \citet{packer2018assessing,zhao2019investigating} provide benchmarks on both in-distribution and out-of-distribution generalization to different dynamics parameters.
Algorithms for improving generalization in RL can be roughly divided into regularization-based methods~\cite{farebrother2018generalization, cobbe2020leveraging, igl2019generalization}, %
methods that use special model architectures~\cite{srouji2018structured,raileanu2021decoupling},
and methods that use data augmentation~\cite{tobin2017domain, lee2019network}. %
While randomizing the dynamics or observations implicitly transforms an MDP into a POMDP, these prior methods normally use memoryless RL algorithms. 
By contrast, we consider memory-based RL algorithms that can adapt online for generalization~\citep[Sec.~5.2]{kirk2021survey}.

\paragraph{Temporal credit assignment.} 
POMDPs are sometimes disguised as MDPs. For example, delaying the reward signals~\citep{sutton1984temporal,arjona2019rudder} can make an MDP into a POMDP, as the rewards at current time still depend on previous observations and/or actions, given current observations. 
Similarly, when rewards are defined in terms of trajectories~\citep{liu2019sequence,ren2021learning} (episodic rewards), the problem can likewise become a POMDP.
Algorithms to solve these problems belong to temporal credit assignment subarea~\citep{hung2019optimizing}.
While prior work has applies recurrent model-free RL to these problem settings, it has been reported with poor performance~\citep{hung2019optimizing,liu2019sequence,arjona2019rudder,raposo2021synthetic}.
Methods to tackle delayed rewards include 
stacking recent observations to make the problems as MDPs~\citep{katsikopoulos2003markov},
tuning the discount factor~\citep{fedus2019hyperbolic} and lambda in eligibility traces~\citep{xu2018meta} to increase effective horizons, 
and using hindsight and counterfactuals to reduce the variance of policy gradients~\citep{harutyunyan2019hindsight,mesnard2020counterfactual}. 
A popular branch of specialized methods, is to learn surrogate reward functions for efficient learning, \eg 
return decomposition into a sum of (dense) rewards~\citep{liu2019sequence,ren2021learning,raposo2021synthetic}, 
and reward redistribution across time~\citep{hung2019optimizing,arjona2019rudder,ferret2019self,gangwani2020learning}.

\section{Design Considerations for Recurrent Model-Free RL}
\label{sec:method}

Implementing a recurrent model-free RL algorithm requires making a number of design decisions.
In the following paragraphs, we will describe the important decision factors we find in recurrent model-free RL. 
Table~\ref{tab:baselines} summarizes how prior work and our work make these design decisions.

\begin{table*}[t]
    \footnotesize
    \centering
    \setlength{\tabcolsep}{2.3pt}
    \caption{{\footnotesize \textbf{How does prior work implement recurrent model-free RL?} 
    Almost no two prior methods implement recurrent model-free RL in the same way. Most prior implementations made design choices that led to poor performance. 
    The last rows show the design decisions that we found to work best on each benchmark.
    }}
    \begin{tabular}{cc|ccccc}
    \toprule
     Algorithm    &  Domain / Benchmark & Arch & Encoder & Inputs & Len & RL \\
     \midrule
     \citet{duan2016rl} & Meta-RL  & separate & GRU & \texttt{oard} & 1000 & TRPO, PPO \\ 
     \citet{wang2016learning} & Meta-RL & shared & LSTM & \texttt{oart} & 5-150 & A2C \\ 
     Baseline in~\citet{rakelly2019efficient} & Meta-RL & separate & GRU & \texttt{oard} & 100 & PPO \\ 
     Baseline in~\citet{zintgraf2019varibad} & Meta-RL &  separate & GRU & \texttt{oard} & Max & A2C, PPO \\
     Baseline in~\citet{fakoor2019meta} & Meta-RL &  separate & GRU & \texttt{oar} & 10-25 & TD3 \\ 
     Baseline in~\citet{yu2020meta} & Meta-RL & separate & GRU & \texttt{oard} & 500 & PPO \\ 
     \citet{kostrikov2018pytorch} & POMDP & shared & GRU & \texttt{o} & 5-2048 & PPO, A2C \\
     \citet{ding2019rlalgorithms} & POMDP & separate & LSTM & \texttt{oa} & 150 & TD3, SAC \\ 
     \citet{meng2021memory} & POMDP & separate & LSTM & \texttt{oa} & 1-5 & TD3 \\ 
     \citet{yang2021recurrent} & POMDP & separate & both & \texttt{oa} & Max & TD3, SAC \\ 
     Baseline in~\citet{igl2018deep} & POMDP & shared & GRU & \texttt{oa} & ~25 & A2C \\ 
     Baseline in~\citet{han2019variational} & POMDP & shared & LSTM & \texttt{o} & 64 & SAC \\
     Baseline in~\citet{zhang2021robust} & Robust RL & separate & LSTM & \texttt{o} & 100 & PPO \\ 
     Baseline 1 in~\citet{packer2018assessing} & Generalization in RL & shared & LSTM & \texttt{o} & 128-512  & PPO, A2C \\ %
     Baseline 2 in~\citet{packer2018assessing} & Generalization in RL & separate & LSTM & \texttt{oard} & 128-512   & PPO, A2C \\ %
     Baseline in~\citet{hung2019optimizing} & Temporal credit assignment & shared & LSTM & \texttt{oar} & Max & A3C \\ 
     Baseline in~\citet{liu2019sequence} & Temporal credit assignment & separate & LSTM & \texttt{oa} & Max & PPO \\ 
     Baseline in~\citet{raposo2021synthetic} & Temporal credit assignment & shared & LSTM & \texttt{oar} & 10-60 & IMPALA \\ 
     \midrule
     \multirow{6}{*}{Our work} & Meta-RL~\citep{dorfman2020offline} & separate & LSTM & \texttt{oard} & 64 & TD3 \\ 
      & Meta-RL~\citep{zintgraf2019varibad} & separate & GRU & \texttt{oard} & Max & SAC \\
      & POMDP~\citep{han2019variational} & separate & GRU & \texttt{oa} &  64 & TD3 \\ 
      & Robust RL~\citep{jiang2021monotonic} & separate & LSTM & \texttt{o} & 64  & TD3 \\ 
      & Generalization in RL~\citep{packer2018assessing} & separate & LSTM & \texttt{o} & 64 & TD3 \\ 
      & Temporal credit assignment~\citep{raposo2021synthetic} & separate & LSTM & \texttt{o} & Max & SAC-D \\ 
     \bottomrule
    \end{tabular}
    \label{tab:baselines}
\end{table*}

\paragraph{Recurrent (off-policy) actor-critic architecture.}
The first important design decision is whether the recurrent policy (actor) and the recurrent Q-value function (critic) use a \textit{shared} RNN encoder (and embedders) or use \textit{separate} ones.
In our experiments (Sec.~\ref{sec:ablation}), we show that a shared encoder increases the gradient norm and hinders learning.
While recent work has adopted the design of separate encoders~\citep{fakoor2019meta, ding2019rlalgorithms, meng2021memory, sun2021safe, weng2021tianshou}, some implementations of recurrent model-free RL use the (inferior) shared encoder.
After running some experiments to compare this design decision, we will use the \textit{separate} architecture in the rest of the paper.

\paragraph{Agent inputs.} The next consideration is the choice of inputs for the actor and critic.
While prior work often only conditions the recurrent RL baseline on previous observations (and actions)~\cite{igl2018deep, kostrikov2018pytorch, han2019variational, ding2019rlalgorithms, meng2021memory, yang2021recurrent}, 
our experiments in Sec.~\ref{sec:ablation} find that additionally conditioning on other previous information, such as previous rewards, can increase return by up to $30\%$.
Table~\ref{tab:problems} shows which inputs we found useful for which types of POMDPs.

\paragraph{Model-free RL algorithm.}
Recurrent model-free RL can be understood as applying an off-the-shelf model-free RL algorithm with an actor and a Q function conditioned on \emph{sequences} of inputs. As such, the choice of the underlying model-free RL algorithm is paramount. 
in
While \textit{off-policy} algorithms such as TD3~\cite{fujimoto2018addressing} and SAC~\cite{haarnoja2018soft,haarnoja2018soft2} improve sample efficiency and asymptotic performance in continuous control tasks, these RL algorithms are rarely used in recurrent model-free RL baselines~\citep{rakelly2019efficient, zintgraf2019varibad,zhang2020robust}.
Our experiments show that using these off-policy algorithms for recurrent model-free RL generally works better than recurrent model-free RL implementations that use on-policy algorithms. This result echoes the finding that model-free off-policy TD3-Context~\cite{fakoor2019meta} can be better than the specialized method PEARL~\cite{rakelly2019efficient} in meta-RL.

\paragraph{RNN variants and context length.}
RNN training is known to be unstable, especially with long sequences input~\cite{bengio1994learning}. RNN variants like LSTM~\cite{hochreiter1997long} and GRU~\cite{chung2014empirical} can mitigate the training issues, but still may fail to learn long-term dependencies~\cite{trinh2018learning}.
We study two design decisions here: the RNN architecture (LSTM, GRU) and the context length, \ie the length of sequence fed into RNN for training.
We find that the architecture has a minor effect on the final performance (App.~\ref{sec:single_factor}).
Prior POMDP methods use context lengths ranging from 1 to 2048 (see the ``Len" column of Table~\ref{tab:baselines}), and we select three representatives of short (5), medium (64), and long length (larger than 100) in the experiments for comparison.  We find that the optimal context length is task-specific (Sec.~\ref{sec:ablation}). For example, a POMDP that hides velocities from observations theoretically only requires a short context length to infer velocities through consecutive positions~\cite{meng2021memory}.

\begin{figure*}[ht]
    \centering
    \begin{subfigure}[b]{0.24\textwidth}
        \includegraphics[width=\linewidth]{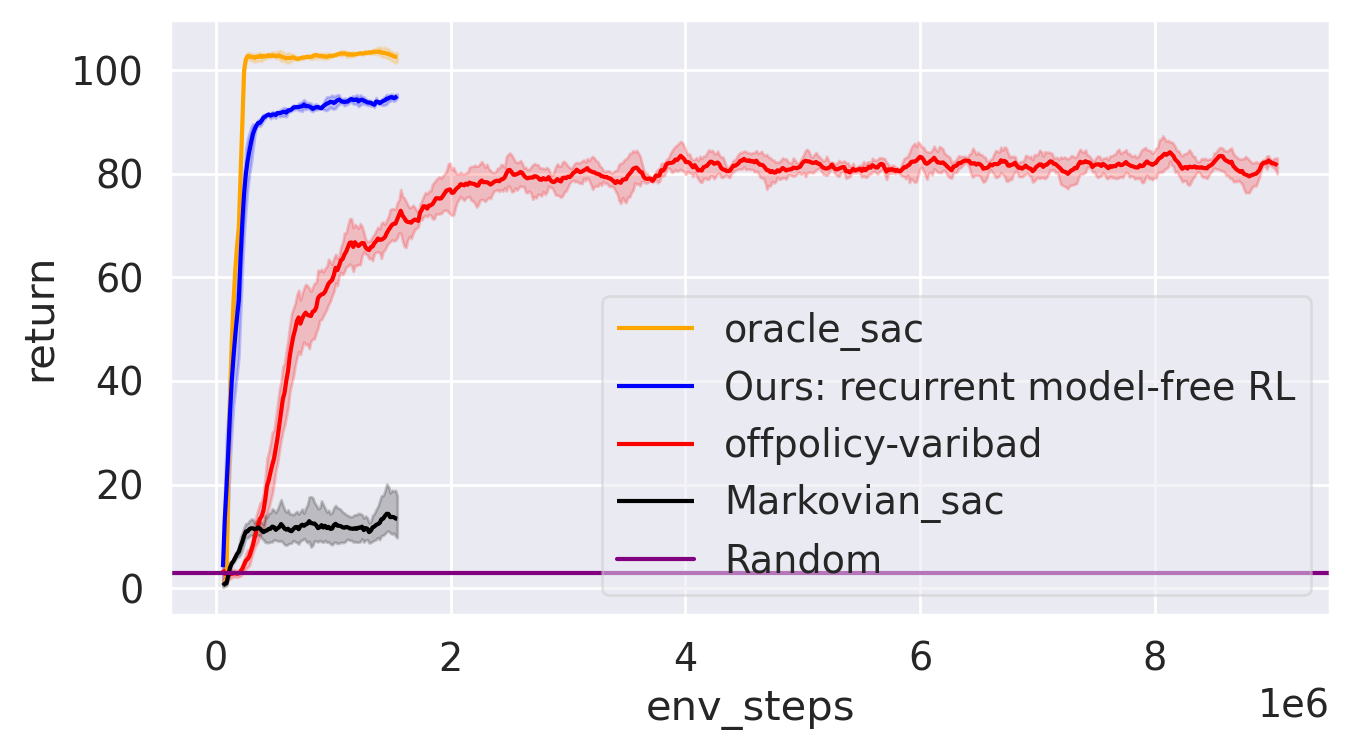}
        \caption*{Semi-Circle}
    \end{subfigure}
    \hfill
    \begin{subfigure}[b]{0.24\textwidth}
        \includegraphics[width=\linewidth]{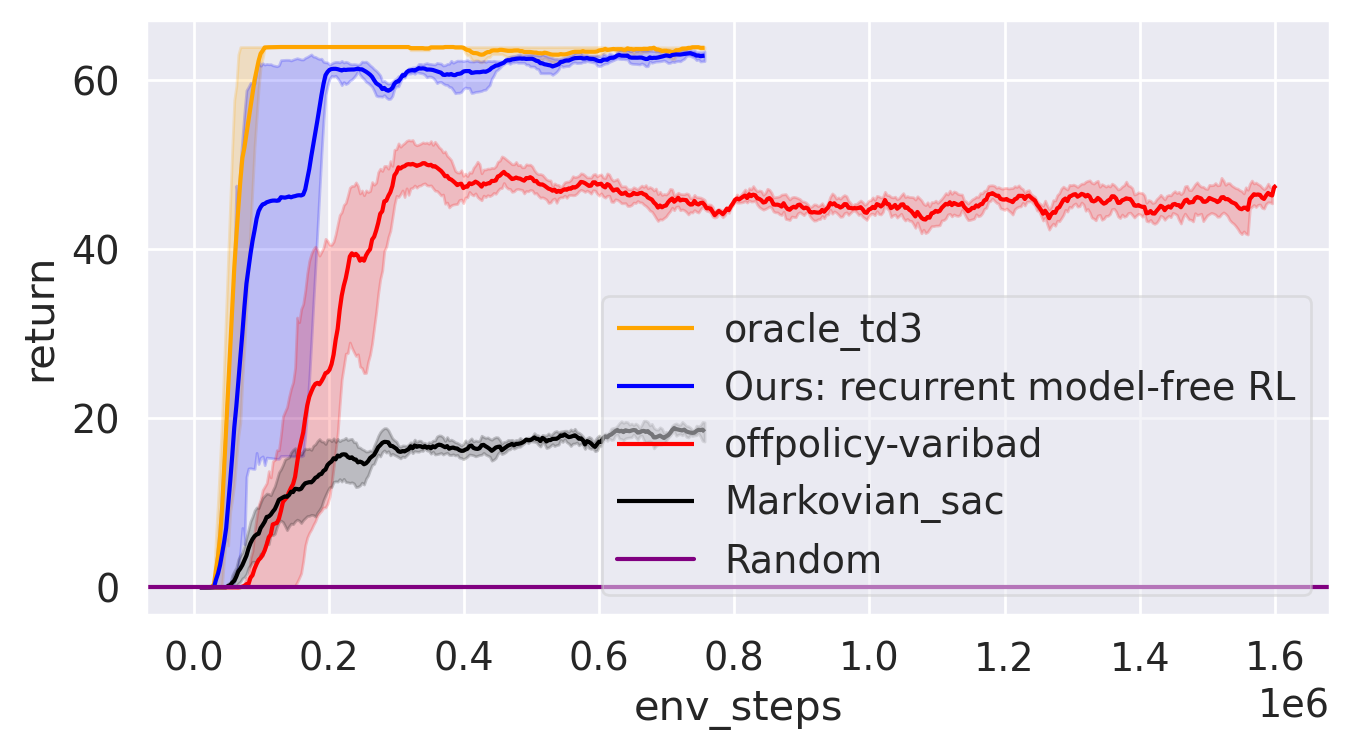}
        \caption*{Wind}
    \end{subfigure}
    \hfill
    \begin{subfigure}[b]{0.24\textwidth}
        \includegraphics[width=\linewidth]{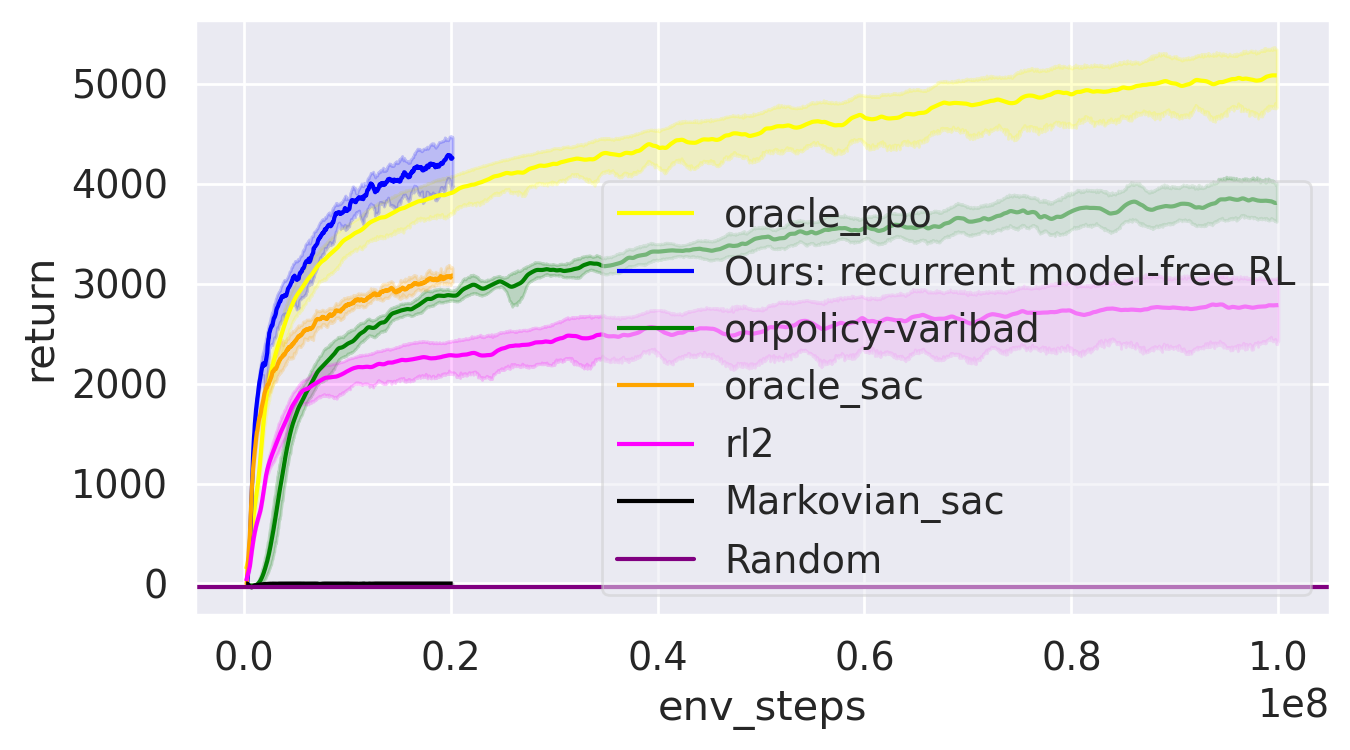} 
        \caption*{Cheetah-Dir}
    \end{subfigure}
    \hfill
    \begin{subfigure}[b]{0.24\textwidth}
        \includegraphics[width=\linewidth]{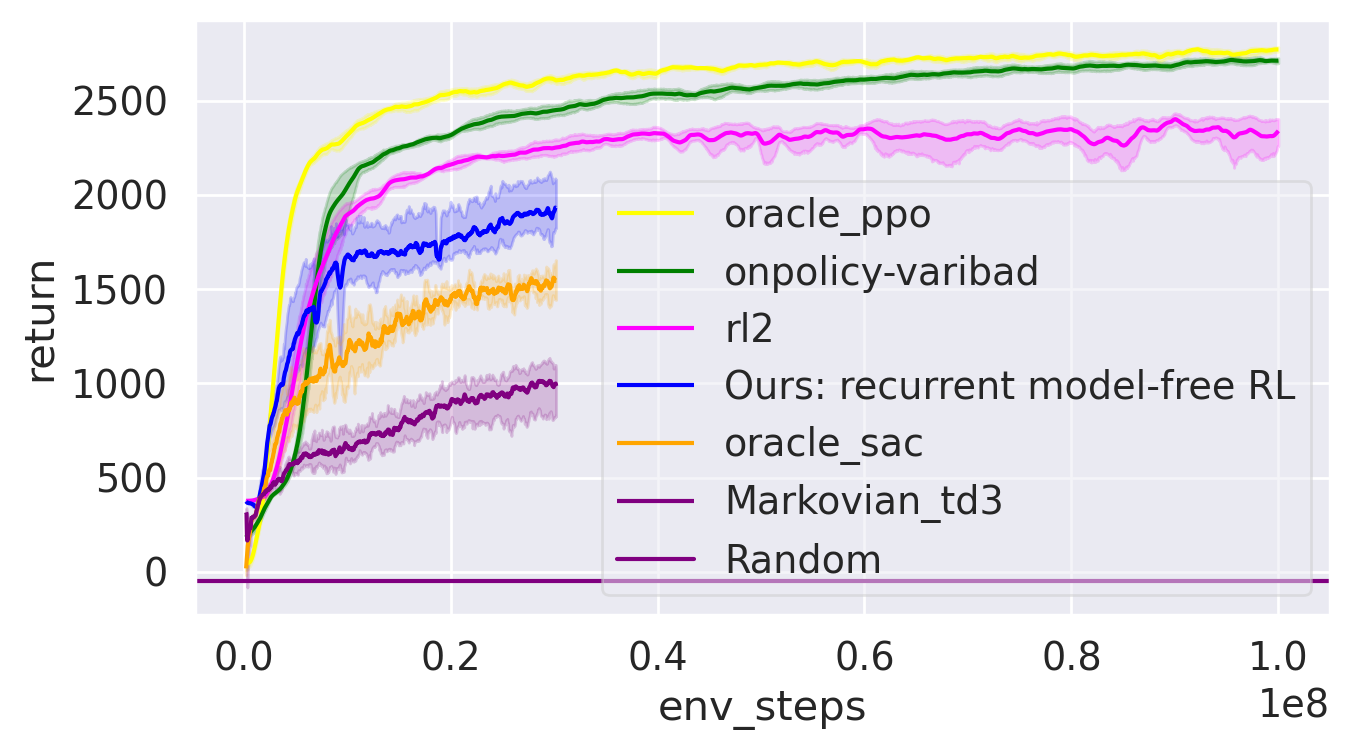} 
        \caption*{Ant-Dir}
    \end{subfigure}
    \vspace{-2mm}
    \caption{\footnotesize \textbf{Learning curves on four meta-RL environments.} 
    Our implementation on recurrent model-free RL can surpass the specialized meta-RL method off-policy variBAD~\citep{dorfman2020offline} on their environments, Semi-Circle and Wind; and greatly outperform on-policy variBAD~\citep{zintgraf2019varibad} on their environment Cheetah-Dir, but fail to match their performance on Ant-Dir.
    On Cheetah-Dir and Ant-Dir, we show the learning curves of the best off-policy oracle and Markovian policies. 
    We copied the data from on-policy variBAD's public github repository\protect\footnotemark\, to plot the learning curves of it, oracle PPO and RL2~\citep{duan2016rl}.}
    \vspace{-4mm}
    \label{fig:meta_best_main}
\end{figure*}

\section{Experiments}
\label{sec:experiments}

Our experiments aim to answer two questions. First, how does a \emph{well-tuned} implementation of recurrent model-free RL compare to specialized POMDP methods?
To give these prior methods the strongest possible footing, we perform the comparison on the benchmarks used by these prior methods.
Second, which design decisions are essential for recurrent model-free RL?
We put the environment details in App.~\ref{sec:env}.

\paragraph{Code implementation.} We release a modular and configurable implementation of recurrent (off-policy) model-free RL in the supplementary material.
Our implementation is efficient in terms of computer memory compared to previous off-policy RL methods for POMDPs. For example, our implementation uses 200x less RAM than \citet{han2019variational} and 9x less GPU memory than \citet{dorfman2020offline}.

\subsection{Recurrent Model-Free RL is Comparable with Prior Specialized Methods on Their Benchmarks}
\label{sec:compare-best}

Recurrent model-free RL is a ubiquitous baseline across a range of different POMDP settings (\eg, meta-RL, occluded observations, delayed rewards)~\citep{rakelly2019efficient,zintgraf2019varibad,humplik2019meta,igl2018deep,han2019variational,arjona2019rudder}. 
This section casts doubt on that claim, showing that a well-tuned implementation of recurrent model-free RL can perform \textit{at least} as well as more complex or specialized methods in \textit{most of} their experimented environments.

We tune a wide range of decision factors, shown in Sec.~\ref{sec:method}. App.~\ref{sec:setting} shows the detailed options of each decision factor.
For each subarea, we select one recent specialized method and use the same benchmark as used in the paper proposing that method. For meta-RL, we perform one additional comparison.

For each benchmark, we show the performance of \textbf{a single variant} among our design combinations that works best across the environments in that benchmark; in other words, we use the same hyperparameters for each task within a benchmark, and do not tune hyperparameters individually for each task.
The exact configurations of each benchmark can be found in the last five rows of Table~\ref{tab:baselines}.
Our implementation is at least comparable to (and sometimes outperforms by a wide margin) prior specialized methods across most tasks (\textbf{18 out of 21 environments}). Our method performs especially well in terms of sample efficiency. 
However, we find one benchmark where it performs worse on 2 out of 3 environments (Ant-Dir and Humanoid-Dir from on-policy variBAD~\citep{zintgraf2019varibad}). This result is not entirely surprising, as on-policy methods typically outperform off-policy methods (which our implementation uses) on the fully-observed versions of these tasks (see Fig.~\ref{fig:meta_on}). 

For each plot of learning curves, we show three approaches as references. First, an \textbf{oracle} policy has access to the POMDP hidden states, turning the POMDP into an MDP. This policy should therefore be treated as an upper bound on the performance that any POMDP method should receive. 
Second, as a lower bound, we use a \textbf{Markovian} policy to solve the POMDP. 
Both oracle policy and Markovian policy are trained with the same hyperparameters as our recurrent model-free RL implementation. 
We also compare to a \textbf{random} policy, which represents a trivial lower bound. 
We show the complete learning curves in App.~\ref{sec:learning_curves} and numerical results in Table~\ref{tab:numerics}, and provide implementation details in App.~\ref{sec:training}.

\paragraph{``Standard" POMDP.} We study ``standard" POMDPs by looking at tasks that typically occlude some part of the observation. We will compare against \textbf{VRM}~\cite{han2019variational}, a recent, state-of-the-art, model-based POMDP algorithm.
We adopt the \textbf{occlusion benchmark} proposed by VRM and there are 8 environments $\{$Hopper, Ant, Walker, Cheetah$\}$-$\{$P, V$\}$, where ``-P" stands for observing positions and angles only, and ``-V" stands for observing velocities only.

Fig.~\ref{fig:trailer} and Fig.~\ref{fig:pomdp_bar} show that the best single variant of our model-free recurrent RL implementation outperforms VRM in 6 out of 8 environments, especially in $\{$Cheetah, Hopper$\}$-P (over $80\%$ of the oracles).
Our results suggest that, while the variational dynamics model used by VRM may be useful for some tasks, this model is not necessary to achieve high results.
While we are primarily interested in sample complexity, but not compute, it is worth noting that our recurrent model-free RL implementation is substantially more efficient than the open-source VRM implementation: our implementation trains $5\times$ faster and can reduce $200\times$ RAM usage (see App.~\ref{sec:code}).

\footnotetext{\url{https://github.com/lmzintgraf/varibad}}

\paragraph{Meta-RL.} We next study the meta-RL setting, where some indicator of the task is unobserved. We compare our implementation of recurrent model-free RL to a specialized, state-of-the-art method, variBAD~\cite{zintgraf2019varibad}. VariBAD explicitly learns task embeddings using a variational, model-based objective. 
While the variBAD was originally proposed using PPO~\citep{zintgraf2019varibad}, recent work has effectively used the same method with SAC~\citep{dorfman2020offline}.
We refer to these methods as \textbf{on-policy variBAD} and \textbf{off-policy  variBAD}, and will compare to both of them.
Importantly, we use the same environments as the papers that proposed these methods.
The environments proposed for off-policy variBAD are relatively easy: Semi-Circle, Wind, and Cheetah-Vel. We adapt Wind to make it harder to solve. Off-policy variBAD is shown to have superior performance over on-policy variBAD in this benchmark~\citep[Fig.~11]{dorfman2020offline}.  
The environments proposed for on-policy one are harder: $\{$Ant, Cheetah, Humanoid$\}$-Dir. On-policy variBAD outperforms RL2~\citep{duan2016rl} in this benchmark~\citep[Fig.~13]{zintgraf2021varibad}.

Fig.~\ref{fig:meta_best_main} shows that our best single variant outperforms off-policy variBAD on their environments (Semi-Circle and Wind), and on-policy variBAD on their environment (Cheetah-Dir), both in terms of sample efficiency and asymptotic return. 
While methods like variBAD disentangle task inference from control, potentially stabilizing training, our experiments suggest that stable training might be  achieved with simple recurrent model-free RL.
As our implementation is off-policy, it has the potential to have better sample efficiency than on-policy variBAD. As our implementation is trained end-to-end, without using pre-trained task representations like off-policy variBAD, it does not have the staleness issue in task representations~\citep{kapturowski2018recurrent}. We believe that these factors may contribute to the relatively good performance of recurrent model-free RL.
Nevertheless, our implementation performs worse than on-policy variBAD on Ant-Dir (Fig.~\ref{fig:meta_best_main}) and Humanoid-Dir (Fig.~\ref{fig:meta_on}). 
These negative results are not entirely surprising, as off-policy methods tend to perform worse than on-policy methods on the fully-observed versions of these tasks (compare oracle SAC/TD3 to oracle PPO in Fig.~\ref{fig:meta_on}). 

\begin{figure}[t]
    \centering
    \footnotesize
    \includegraphics[width=\linewidth]{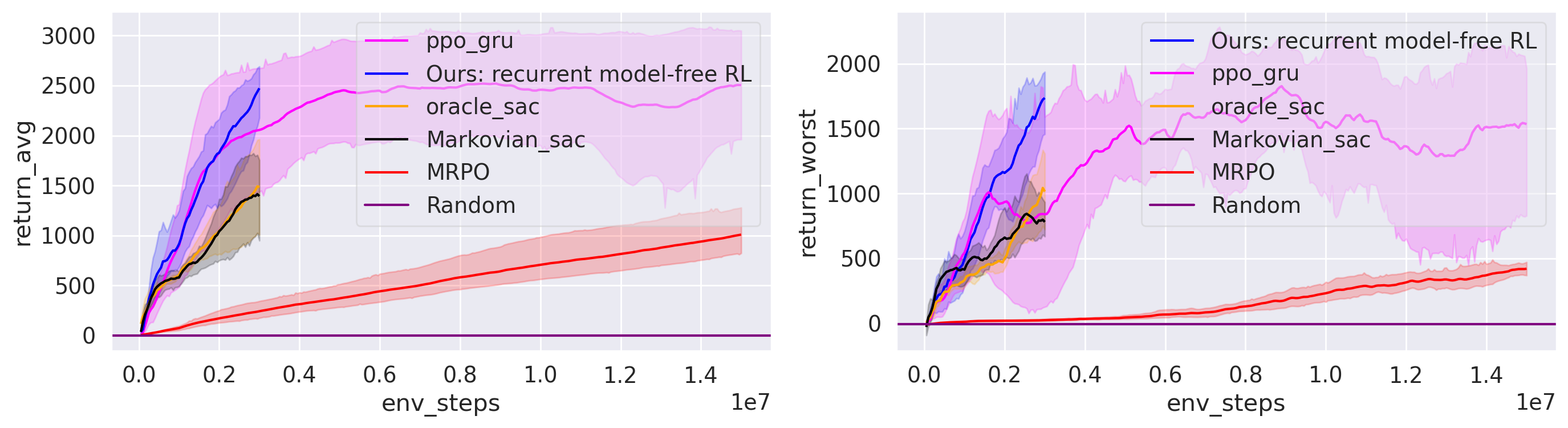}
    \vspace{-5mm}
    \caption{\footnotesize \textbf{Learning curves on one robust RL environment,} Cheetah-Robust. 
    We show the average returns (left figure) and worst returns (right figure) of each method. The \textbf{single best variant} of our implementation on recurrent model-free RL can greatly outperform the specialized robust RL method MRPO~\cite{jiang2021monotonic}, and is more sample-efficient and stable than recurrent PPO.}
    \vspace{-2mm}
    \label{fig:rmdp_best_main}
\end{figure}

\paragraph{Robust RL.} We then study robust RL. We choose the recent, specialized algorithm \textbf{MRPO}~\cite{jiang2021monotonic}, and adopt their benchmark based on \textbf{SunBlaze benchmark}~\cite{packer2018assessing}.
These environments have hidden states that are fixed during one episode, namely $\{$Cheetah, Hopper, Walker$\}$-Robust. The hidden state includes the density and the friction coefficients of the simulated robots.

Fig.~\ref{fig:rmdp_best_main} shows both the \textbf{average return} and \textbf{worst return} of our single best variant and MRPO on one environment. Following prior work~\citep{jiang2021monotonic}, we measure the worst return using the average return in the worst $10\%$ testing tasks.
Despite using the average-case objective, our method achieves better worst-case performance than MRPO, which directly optimizes this worst-case objective. Surprisingly, our method even outperforms the oracle, which has access to hidden state information. Our method is also over 80\% more sample-efficient than these alternative approaches.
Our limitation of recurrent model-free RL is its slow wall-clock time; our implementation is 17.5x slower than MRPO, given the same simulation steps (see App.~\ref{sec:code}). Nonetheless, we believe that sample efficiency is often a more important factor than computing efficiency.

\begin{figure}
    \centering
    \footnotesize
    \includegraphics[width=\linewidth]{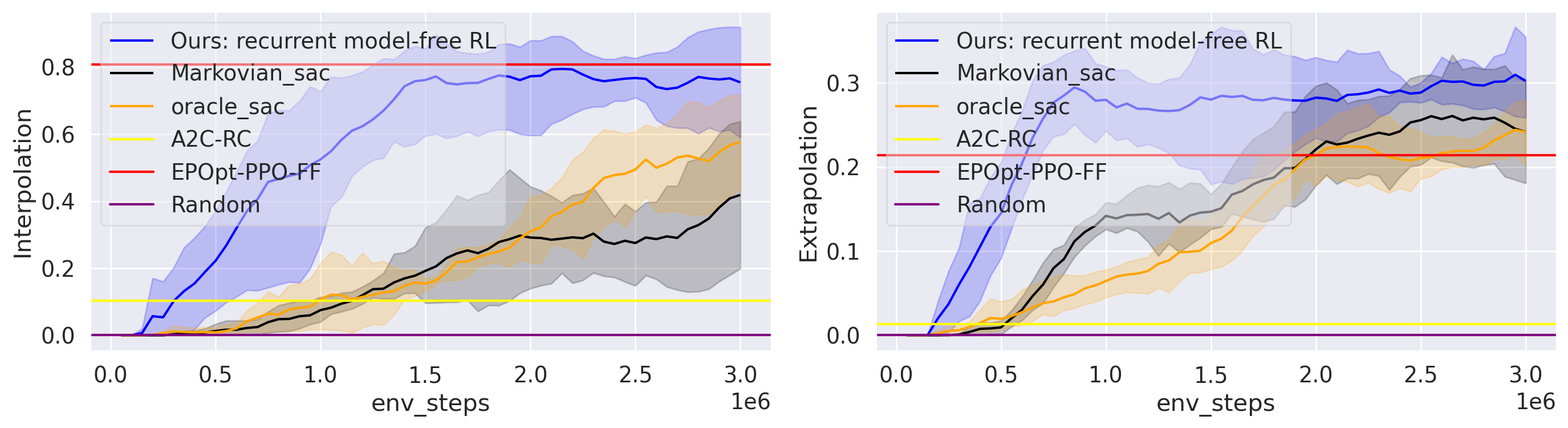}
    \vspace{-5mm}
    \caption{\footnotesize \textbf{Learning curves on RL in generalization in one environment,} Hopper-Generalize.
    We show the interpolation success rates (left figure) and extrapolation success rates (right figure) of each method. The \textbf{single best variant} of our implementation on recurrent model-free RL can be par with the specialized method EPOpt-PPO-FF~\cite{rajeswaran2016epopt} in interpolation and outperform it in extrapolation. The data of EPOpt-PPO-FF and A2C-RC (a recurrent model-free on-policy RL method) are copied from the Table 7 \& 8 in~\citet{packer2018assessing}.}
    \vspace{-2mm}
    \label{fig:generalize_best_main}
\end{figure}

\paragraph{Generalization in RL.} We study generalization in RL using two environments from the \textbf{SunBlaze benchmark}: $\{$Hopper, Cheetah$\}$-Generalize.
Following the evaluation metrics~\citep[Sec.~6]{packer2018assessing}, we report the average success rates in \textbf{interpolation} setting (training and testing on the same POMDP) and \textbf{extrapolation} setting (training on a POMDP with a hidden state distribution of small support, and testing on another POMDP with a hidden state distribution of a disjoint support). 
We pick the best specialized method in the tables of final performance~\citep[Table~7,8]{packer2018assessing}, a Markovian on-policy robust RL method \textbf{EPOpt-PPO-FF}~\cite{rajeswaran2016epopt}.

Fig.~\ref{fig:generalize_best_main} shows interpolation and extrapolation results on one environment. 
In the interpolation setting, our method performs on par with the best prior method, EPOpt-PPO-FF. 
In the more challenging extrapolation setting, our method outperforms it and is comparable to the oracle.
However, unlike EPOpt-PPO-FF and oracle, our method does not require access to the dynamics parameters.

\begin{figure}
    \centering
    \footnotesize
    \includegraphics[width=0.49\linewidth]{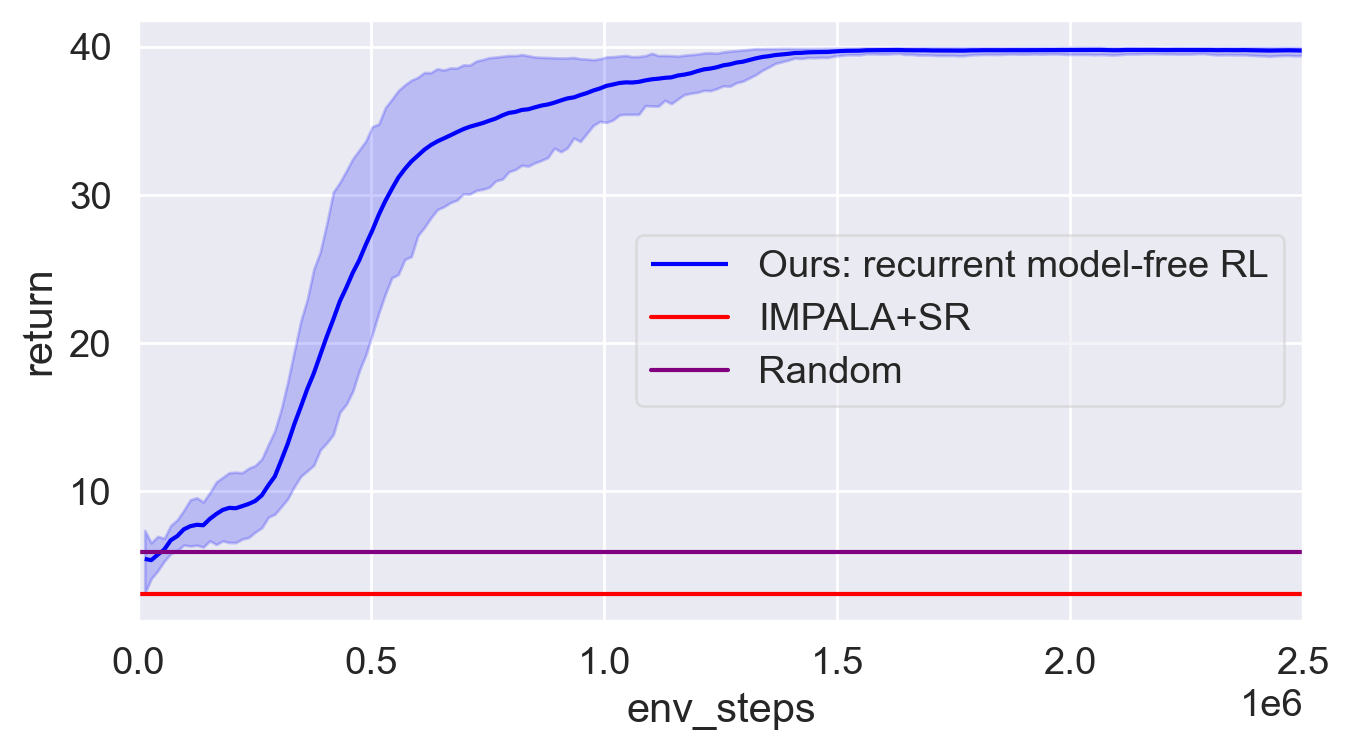}
    \includegraphics[width=0.49\linewidth]{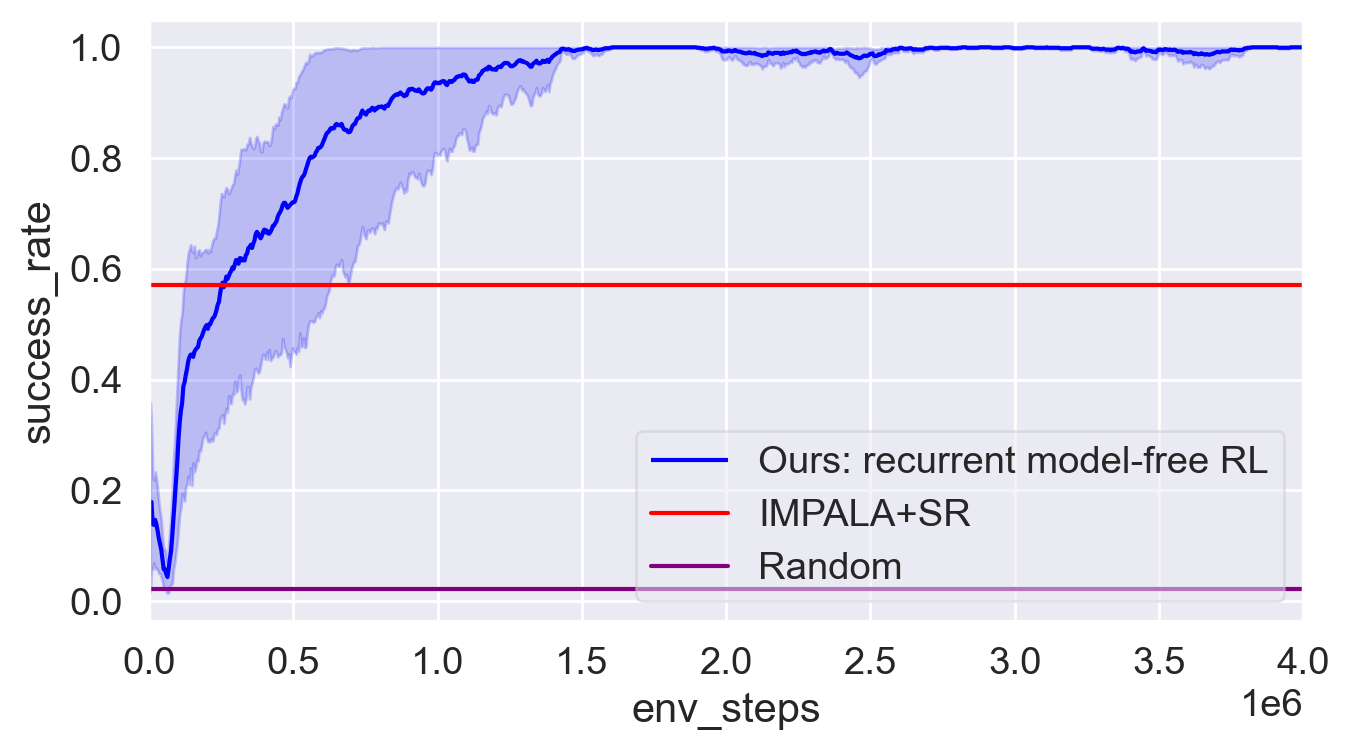}
    \begin{tabular}{P{0.49\linewidth}P{0.49\linewidth}}
Delayed-Catch     &  Key-to-Door
\end{tabular}
    \vspace{-5mm}
    \caption{\footnotesize\textbf{Learning curves on two temporal credit assignment environments.} We show the returns for Delayed-Catch and the success rates of opening the door for Key-to-Door, following the practice of IMPALA+SR~\citep{raposo2021synthetic}. 
    The \textbf{single best variant} of our implementation on recurrent model-free RL is much more sample efficient than the specialized method IMPALA+SR (the horizontal lines show their performance at 2.5M and 4M steps, respectively).}
    \vspace{-5mm}
    \label{fig:credit_main}
\end{figure}

\paragraph{Temporal credit assignment.} Finally we move on to temporal credit assignment. We choose the recent, specialized algorithm IMPALA+SR~\citep{raposo2021synthetic}, and adopt their environments, namely, Delayed-Catch and Key-to-Door. Both tasks have sparse rewards that depend on the whole trajectory, thus the optimal value function should be memory-based, and we did not compare Markovian RL methods. 
As both tasks are discrete control with pixel input, we adapt the observation embedder into a simple CNN, and select SAC-Discrete~\citep{christodoulou2019soft} as the RL algorithm, which is the discrete version of SAC. We follow IMPALA+SR to use LSTM as encoder and set context length as full episode length. We tune the entropy temperature of SAC-Discrete and find that $0.1$ works well on both tasks. 

Fig.~\ref{fig:credit_main} shows this single best variant can not only solve the tasks, but also requires $100\times$ fewer samples than IMPALA+SR~\citep[Fig.~7b, Fig.~5b]{raposo2021synthetic}, which is also a recurrent off-policy method.

\paragraph{Discussion on the performance of oracle policies.} One seemingly surprising result is that the oracle policies often \textit{underperform} our implementation of recurrent model-free RL. We believe that this result is caused by using the same hyperparameters for the oracle policies as for our recurrent model-free implementation. We expect that further tuning of the hyperparameters for the oracle policies would allow them to surpass all the alternative approaches.

In summary, recurrent model-free RL can perform at least as well as more specialized or complex methods on most of their tasks, provided that the implementation is well tuned. The good performance of this baseline across a wide range of tasks and problem types bodes well for its performance on other problems.

\subsection{What Matters in Recurrent Model-Free RL Algorithms?}
\label{sec:ablation}

To study what factors explain the good performance of recurrent model-free RL, we will ablate the five important design decisions introduced in Sec.~\ref{sec:method}: the actor-critic architecture (\texttt{Arch}), the agent input space (\texttt{Inputs}), the underlying model-free RL algorithm (\texttt{RL}), the RNN encoder (\texttt{Encoder}), and the RNN context length (\texttt{Len}).
See Table~\ref{tab:baselines} for a summary of how prior work made these design decisions. 
Due to the space limit, we show the \textbf{``single factor analysis"} plots for each decision factor by averaging the performance over the other factors in App.~\ref{sec:single_factor}.

\begin{figure}
\centering
    \includegraphics[width=\linewidth]{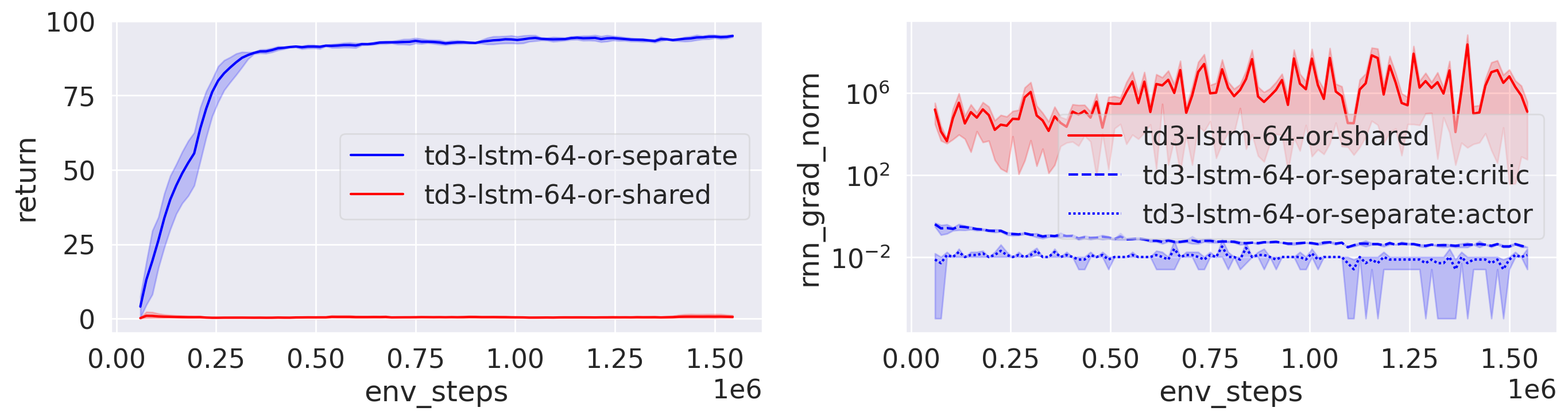}
    \vspace{-3mm}
    \caption{\footnotesize{\textbf{Comparison between \textit{shared} and \textit{separate} recurrent actor-critic architecture} with all the other hyperparameters same, on Semi-Circle, a toy meta-RL environment. We show the performance metric (left) and also the average squared $\ell_2$-norm of the gradient  \wrt RNN encoder(s)  (right, in \textbf{log-scale}). 
    For the separate one, \texttt{:critic} and \texttt{:actor} refer to the separate RNN in critic and actor networks, respectively.}}
    \vspace{-5mm}
    \label{fig:separate_vs_shared_main}
\end{figure}

\paragraph{Recurrent off-policy actor-critic architecture.} We study the choice of shared/separated architectures in two simple POMDP environments. 
The results, shown in Fig.~\ref{fig:separate_vs_shared_main} and App.~\ref{sec:results_arch}, show that the shared architecture failed to learn either of these tasks.
The different scale of RNN gradient norms \wrt actor and critic in the shared architecture suggests that the gradient of critic loss may dominate the actor's.
Our results echo prior work~\citep{fakoor2019meta,meng2021memory,sun2021safe} that use \textit{separate} RNN encoders and achieve high asymptotic rewards, and also echo that \cite{han2019variational} shows poor results in the \textit{shared} architecture of SAC-LSTM. 

\begin{table*}[t]
    \centering
    \footnotesize
\renewcommand{\arraystretch}{1.2} %
    \caption{\footnotesize{\textbf{Ablation results in our implementation of recurrent model-free RL.} This table shows how a single change in one decision factor from the variant that is best on average in that subarea/benchmark, could significantly increase the performance. 
    The first column shows how we change the single decision factor, and the last column shows the performance comparison between \textbf{the best variant in that benchmark} (left) and \textbf{the ablated one} (right). 
    We choose the environments where the ablation makes the \textit{largest} performance difference.
    For robust RL and generalization in RL, we show the performance metric in worst returns and extrapolation success rates, respectively.}}
    \begin{tabular}{c|cccc}
    \toprule
     Change in one decision factor &  Subarea / Benchmark & Environment   & Performance comparison \\
     \midrule
     \texttt{Inputs}: \texttt{oa} $\to$ \texttt{oar} & ``Standard" POMDP & Walker-P & 981.6 $\to$ 1345.0 ({\color{green}$1.3\times$})\\ %
     \texttt{RL}: TD3 $\to$ SAC & ``Standard" POMDP & Ant-P & 310.7 $\to$ 2123.5  ({\color{green}$6.8\times$}) \\ %
     \texttt{Encoder}: LSTM $\to$ GRU & Robust RL & Walker-Robust & 765.9   $\to$ 931.3 ({\color{green}$1.2\times$}) \\ %
     \texttt{Len}: 64 $\to$ 400 & Meta-RL & Cheetah-Vel & -85.2  $\to$ -74.6  ({\color{green}+14\%}) \\  %
     \texttt{Len}: 64 $\to$ 5  & Generalization & Hopper-Generalize & 0.292  $\to$ 0.415 ({\color{green}$1.4\times$}) \\ %
     \texttt{Len}: 64 $\to$ 5  & ``Standard" POMDP & Walker-V & 121.4 $\to$ 264.3 ({\color{green}$2.2\times$}) \\ %
     \bottomrule
    \end{tabular}
\vspace{-4mm}
    \label{tab:ablation}
\end{table*}

\paragraph{Agent inputs.} 
We next study the choice of agent inputs using the Walker-P task. As shown in Table~\ref{tab:ablation} (row 1), additionally conditioning the agent on past rewards increases performance by $1.3\times$. The reward signals could help reveal the missing information of the velocity of the robot base, which is occluded in Walker-P. 
On other tasks where velocity is occluded, we similarly find that conditioning on past rewards improves performance (see \texttt{o} vs. \texttt{or} in Fig.~\ref{fig:pomdp_ablation_P}). 

\paragraph{Model-free RL algorithms.} Table~\ref{tab:baselines} shows that recurrent model-free RL implementations using TD3 dominate in 4 out of 5 benchmarks. This finding might be partially explained by the fact that most environments have relatively easy dynamics. 
However, on environments with harder dynamics such as Ant and Humanoid, SAC performs better than TD3. For instance, the 2nd row of Table~\ref{tab:ablation} shows the effect of RL algorithm in a POMDP environment Ant-P. SAC is significantly better than TD3 (increase by 6.8$\times$, surpassing the PPO-GRU~\cite{kostrikov2018pytorch} in Fig.~\ref{fig:trailer}). 
Two exceptions to this rule are Walker-V (Fig.~\ref{fig:pomdp_ablation_V}) and Ant-Dir (Fig.~\ref{fig:meta_best_main}) where on-policy algorithms (PPO-GRU and RL2) outperform off-policy ones as used in our implementation.

\paragraph{RNN variants and context length.} Generally, there is no significant difference between LSTM and GRU (see the single factor analysis in App.~\ref{sec:single_factor}). However, the 3rd row of Table~\ref{tab:ablation} shows the effect of RNN encoder in a robust RL environment. We can see replacing LSTM with GRU can increase the worst-case metric in Walker-Robust.
For the context length in RNNs, a medium length (64) dominates in all the best variants in most benchmarks (see Table~\ref{tab:baselines}), which could be viewed as a trade-off between memory capacity and computation costs. 
However, the remaining rows of Table~\ref{tab:ablation} show the mixed effects of context length in RNNs. Both increasing and decreasing the context length can boost the performance in different environments. Specifically, decreasing the length from 64 to 5 makes our implementation surpass VRM in Walker-V (increase by 2.2$\times$). 
This result might explain why the prior methods adopt a wide range of context lengths from 1 to 2048 (see Table~\ref{tab:baselines}). Therefore, the choice of context length seems to be problem-specific and may require tuning.

\paragraph{Summary.}
We now summarize the main findings of our experiments based on the benchmarks:
\vspace{-3mm}
\begin{enumerate}[itemsep=0mm, left=1em]
    \item Using separate weights for the recurrent actor and recurrent critic can boost performance, likely because it avoids gradient explosion (Fig.~\ref{fig:separate_vs_shared_main} and Fig.~\ref{fig:separate_vs_shared}).
    \item Using state-of-the-art off-policy RL algorithms as the backbone in recurrent model-free RL can improve asymptotic performance and sample efficiency in most environments (Fig.~\ref{fig:trailer} and Figures in App.~\ref{sec:learning_curves}).
    \item The context length for the recurrent actor and critic has a large influence on task performance, but the optimal length seems to be task-specific. Starting with a medium length is a good strategy (Rows 4--6 in Table~\ref{tab:ablation} and Figures in App.~\ref{sec:single_factor}).
    \item It is important that the inputs to the recurrent actor and critic, such as past observations and past returns, contain enough information to infer the POMDP hidden states (Row 1 in Table~\ref{tab:ablation} and Figures in App.~\ref{sec:single_factor}).
\end{enumerate}
\vspace{-3mm}
These findings may provide a useful initialization for researchers to study recurrent model-free RL.

\vspace{-2mm}
\section{Conclusion and Future Work}

This paper shows that a carefully-designed implementation of recurrent model-free RL can perform well across a range of benchmarks corresponding to different types of POMDPs. In most cases, our implementation performs on par with (if not significantly better than) prior methods that are specifically designed for the corresponding types of POMDPs. Our ablation experiments demonstrate the importance of key design decisions, such as the underlying RL algorithm and the RNN context length. While the best choices for some decisions (such as using separate RNNs for the actor and the critic) seem to be consistent across domains, the best choices for other decisions (such as RNN context length) are problem-dependent. We encourage future work to study automated mechanisms for selecting these crucial design decisions. In releasing our code, we hope to aid future research into the design of stronger POMDP algorithms.

{\footnotesize 
\subsection*{Acknowledgement}
We thank Pierre-Luc Bacon, Murtaza Dalal, Paul Pu Liang, Sergey Levine,  Evgenii Nikishin, Hao Sun, and Maxime Wabartha for their constructive feedback on the draft of this paper.
TN thanks Pierre-Luc Bacon for suggesting temporal credit assignment experiments and Michel Ma and Pierluca D'Oro for their help on the environments. 
We thank Luisa Zintgraf for sharing the learning curves of on-policy variBAD.
TN thanks CMU cluster and Mila cluster for compute resources. 
This work is supported by the Facebook CIFAR AI Chair, the Fannie and John Hertz Foundation and NSF GRFP (DGE1745016).
}

{\small
\bibliography{citation}
\bibliographystyle{icml2022}
}

\newpage
\appendix
\onecolumn

\section{Code-Level Details}
\label{sec:code}
In this section, we first introduce the outline of code design, especially the replay buffer for sequences, and then compare the system usage, including computing speed, RAM, and GPU memory, with previous POMDP methods. 

\subsection{Code Design}

\paragraph{Easy to use.} Our code can be used either as an API to call the recurrent model-free RL class or a framework to tune the details in the class. 
The recurrent model-free RL class takes the hyperparameters of RNN encoder type, shared or separate actor-critic architecture, and whether include previous observations, and/or actions, and/or rewards into the inputs, to generate different instances. The details of the hyperparameter tuning set are shown in Sec.~\ref{sec:setting}.

\paragraph{Memory-efficient replay buffer for sequence data.} Moreover, we design an efficient replay buffer for off-policy RL methods to cope with sequential inputs. Previous methods~\cite{han2019variational,yang2021recurrent} mainly use a \textit{three-dimensional} replay buffer to store sequential inputs, with the dimensions of \texttt{(num episodes, max episode length, observation dimension)}, taking observation storage as an example. 
This kind of implementation becomes memory-inefficient if the actual episode length is far smaller than the max episode length (\eg in VRM's occlusion benchmark, the shortest episode length can be 5, while the max episode length is 1000, which can cause 200x waste in RAM).
Instead, we manage to implement a two-dimensional replay buffer of shape \texttt{(num transitions, observation dimension)} for observation storage, which also records the locations where each stored episode ends. 
In case of actual episodes that are shorter than the provided context length, the buffer also generates \textit{on-the-fly masks} to indicate if the corresponding transitions are valid, so that we do not need to save zero-padded observations in the buffer. 
This enables the agent to receive a batch of previous experiences in a three-dimensional tensor of \texttt{(batch size, context length, observation dimension)} when sampling from the replay buffer. 
To sum up, our replay buffer can support varying-length sequence inputs and subsequence sampling without zero padding in the buffer.

\paragraph{Flexible training speed.} Finally, our code supports flexible training speed by controlling the ratio of the numbers of gradient updates in RL w.r.t. the environment rollout steps (called \texttt{num\_updates\_per\_iter} in the code).
The training speed is approximately proportional to the ratio if the simulator speed is much faster than the policy gradient update. 
Typically, the ratio is less than or equal to 1.0 to enjoy higher training speed.

\subsection{System Usage}

Table~\ref{tab:system} shows the typical system usage of our implementation and the compared specialized methods on different environments. 
The time cost for our implementation and off-policy variBAD depends on how many processes in parallel are run on a single GPU -- our implementation can be run with 8 processes on a single GPU while off-policy variBAD is run with one process due to large GPU memory usage.
From the results we can see that our implementation is memory-efficient in both RAM and GPU, and has an acceptable training speed with default hyperparameters. 
The computer system we used during the experiments includes a GeForce RTX 2080 Ti Graphic Card (with 11GB memory) and Intel(R) Xeon(R) Gold 6148 CPU @ 2.40GHz (with 250GB RAM and 80 cores).

\begin{table}[h]
    \centering
    \vspace{-1em}
    \caption{\textbf{Comparison between our implementation and specialized methods in system usage.} 
    The time costs are evaluated within 1M environment steps. 
    Both VRM and MRPO are run on CPUs and MRPO does not have a replay buffer (shown in N/A). Off-policy variBAD requires the assumption of fixed episode length for the RAM cost.}
    \vspace{2mm}
    \begin{tabular}{cc|ccc}
    \toprule
     Method    & Environment &  Time cost& RAM & GPU memory  \\
     \midrule
     \textbf{Ours} & Hopper-V & 22.5 h  & O(1) &  1.2 GB \\ 
     VRM~\cite{han2019variational} & Hopper-V &  102 h & O(200) & N/A \\ 
     \textbf{Ours} & Semi-Circle &  12 h  & O(1) & 1 GB \\ 
    Off-policy variBAD~\cite{dorfman2020offline} & Semi-Circle & 2.3 h  & O(1)* &  9.5 GB \\ 
    \textbf{Ours} & Cheetah-Robust &  7 h & O(1) & 1.1 GB \\ 
    MRPO~\cite{jiang2021monotonic} & Cheetah-Robust &  0.4 h & N/A & N/A \\ 
    \bottomrule
    \end{tabular}
    \label{tab:system}
\end{table}

\subsection{Our Hyperparameter Tuning Set}
\label{sec:setting}

Our proposed implementation has the following decision factors (introduced in Sec.~\ref{sec:method}) to tune in the experiments with the following options (the names in brackets are abbreviated ones):
\begin{itemize}
    \item Actor-Critic architecture (\textbf{Arch}): share the encoder weights between the recurrent actor and recurrent critic or not, namely \texttt{shared} and \texttt{separate}.
    \item Model-free RL algorithms (\textbf{RL}): \texttt{td3}~\cite{fujimoto2018addressing} and \texttt{sac}~\cite{haarnoja2018soft2} (\ie automated tuning of the entropy temperature) 
    \item Encoder architecture (\textbf{Encoder}): \texttt{lstm}~\cite{hochreiter1997long} and \texttt{gru}~\cite{cho2014learning}.
    \item Agent inputs (\textbf{Inputs}): \texttt{o}, \texttt{oa}, \texttt{or}, \texttt{oar}, \texttt{oard} (the notation is introduced in Sec.~\ref{sec:method}; depending on the POMDPs, see ``Agent input space'' row in Table~\ref{tab:hparams}).
    \item Context length (\textbf{Len}): short (5), medium (64), long (larger than 100, depending on the POMDPs).
    \item Entropy temperature of SAC-Discrete (SAC-D)~\citep{christodoulou2019soft} (used in temporal credit assignment tasks): 0.001, 0.01, 0.1, 1.0. 
\end{itemize}

For each instance, we label it with the names of all the hyperparameters it used in lowercase as notation. For example, \texttt{td3-lstm-64-or-separate} in Fig.~\ref{fig:separate_vs_shared_main} refers to the instance that uses the separate actor-critic architecture, TD3 RL algorithm, LSTM encoder, the agent input space of previous observations and reward sequences, and RNN context length of 64.

\section{Training Details}
\label{sec:training}

\begin{figure}[h]
    \centering
    \includegraphics[width=\linewidth]{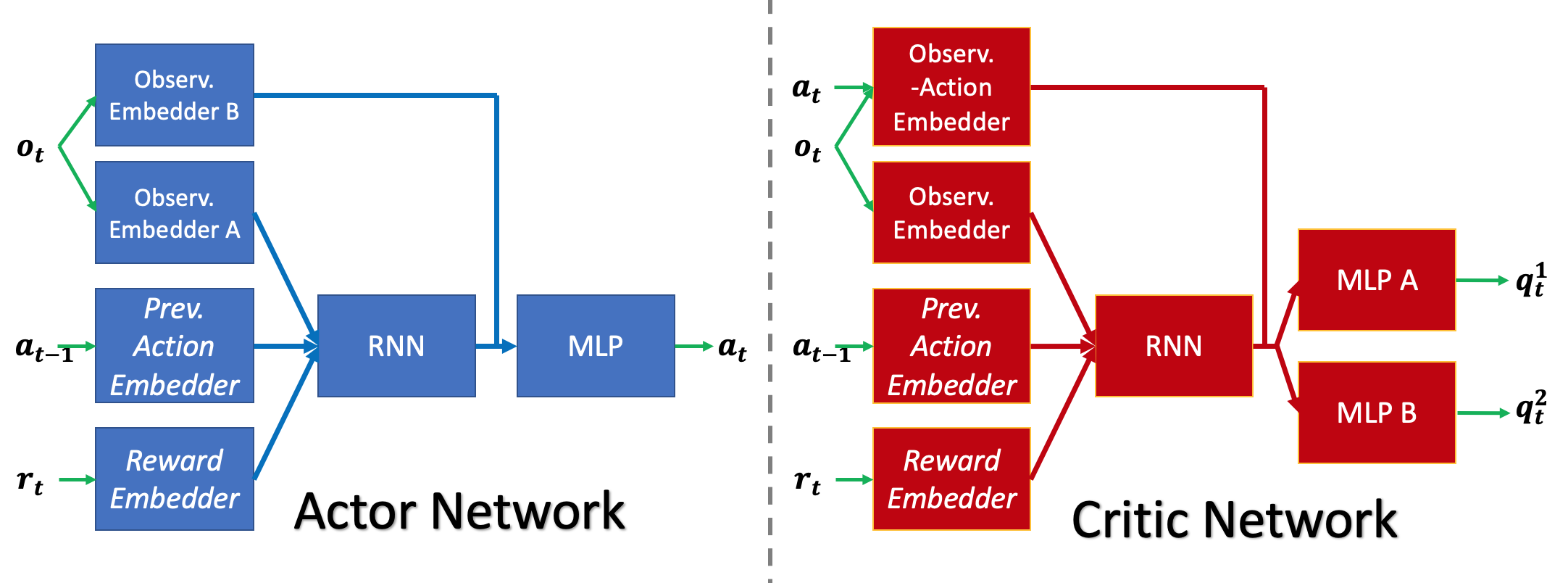}
    \caption{\textbf{The network architecture of our implementation on recurrent model-free RL with \textit{separate} RNNs.} The left part shows the actor network, and the right shows the critic network. Each block shows a trainable module, with independent weights. We \textit{italicize} the previous action and reward embedders as they are optional. By default, each embedder has one hidden layer, each RNN is one-layer LSTM or GRU, each MLP has two hidden layers.}
    \label{fig:arch}
\end{figure}

\begin{figure}[h]
    \centering
    \includegraphics[width=0.6\linewidth]{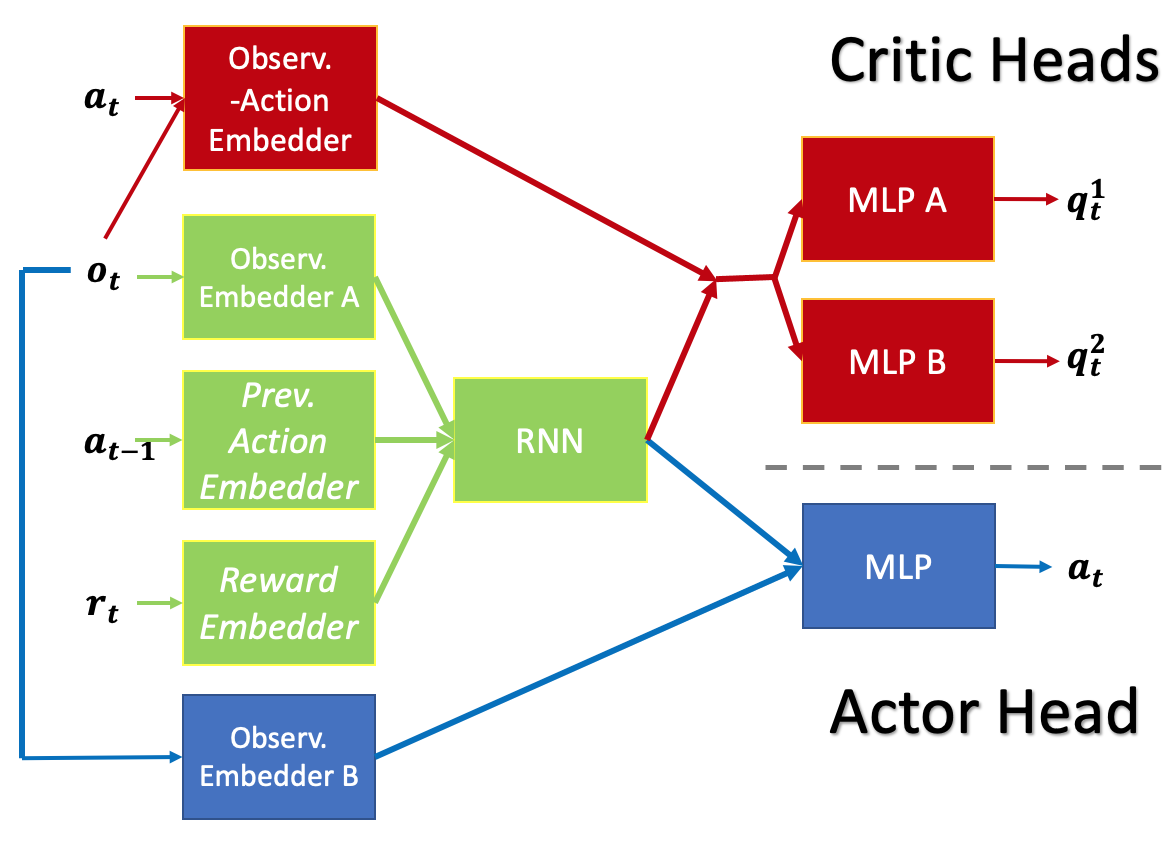}
    \caption{\textbf{The network architecture of our implementation on recurrent model-free RL with \textit{shared} RNN.} The upper right part shows the critic heads, and the bottom right shows the actor head. Both take the inputs from the same RNN. Notation is same as Fig.~\ref{fig:arch}.}
    \label{fig:arch_shared}
\end{figure}

\begin{table}[t]
    \centering
    \caption{\textbf{Hyperparameter summary in our implementation of model-free recurrent RL.} For each benchmark, we report the hidden layer size of each module, RL and training hyperparameters.
    For meta-RL, we take the model on Cheetah-Vel as example, which follows the architecture design of off-policy variBAD~\cite{dorfman2020offline}. The hidden size of observation-action embedder is the sum of that of observation embedder, previous action embedder (if exists), and reward embedder (if exists).}
    \vspace{2mm}
    \begin{tabular}{c|c|ccccc}
    \toprule
     \multirow{2}{*}{}  & \multirow{2}{*}{}  &  \multirow{2}{*}{Meta-RL} & ``Standard''   & \multirow{2}{*}{Robust RL} & Generalization   & Temporal \\
     & & & POMDP & & in RL & credit assignment \\  \midrule
     \multirow{5}{*}{\vtop{\hbox{\strut Hidden}\hbox{\strut layer}\hbox{\strut size}}} &  Observ. embedder   & \multicolumn{4}{c}{[32]}  & 2-layer CNN\\ 
      & Prev. Action embedder  & \multicolumn{4}{c}{[16]} & not used \\
     & Reward embedder  &  \multicolumn{4}{c}{[16]} & not used \\
     & RNN & \multicolumn{5}{c}{[128]} \\
     & MLP & [128,128,128]  & \multicolumn{3}{c}{[256, 256]} & [128, 128] \\ \midrule
  \multirow{7}{*}{\vtop{\hbox{\strut RL}\hbox{\strut hparams}}} 
  & Optimizer & \multicolumn{5}{c}{Adam~\citep{kingma2014adam}} \\
  & Learning rate & \multicolumn{5}{c}{3e-4} \\
  & Discount factor $\gamma$ & \multicolumn{5}{c}{0.99} \\
  & Smoothing coef $\tau$ & \multicolumn{5}{c}{0.005} \\
  & SAC(D) temperature & \multicolumn{4}{c}{automatically updated by~\citet{haarnoja2018soft2}} & 0.1 \\
  & TD3 noises & \multicolumn{4}{c}{default values from~\citet{fujimoto2018addressing}} & N/A \\
  & Replay buffer size & \multicolumn{5}{c}{1e6} \\
   & Batch size & 32 & \multicolumn{3}{c}{64} & 32 \\ \midrule
  RNN & Weight initialization & \multicolumn{5}{c}{Orthogonal matrices~\citep{saxe2013exact}} \\ \midrule
  \multirow{2}{*}{\vtop{\hbox{\strut Training}\hbox{\strut hparams}}} & Environment steps & 5M & 1.5M & \multicolumn{2}{c}{3M} & 5M \\ 
  & Gradient steps & 0.1M & 1.5M & \multicolumn{2}{c}{0.6M}  & 1.25M \\ \midrule
   \multirow{2}{*}{\vtop{\hbox{\strut Agent}\hbox{\strut inputs}}}  & Largest input space & \texttt{oard} & \texttt{oar} & \texttt{oa} & \texttt{oar} & \texttt{o} \\  
   & Best input space & \texttt{oard} & \texttt{oa} & \texttt{o} & \texttt{o} & \texttt{o} \\  
  \bottomrule
    \end{tabular}
    \label{tab:hparams}
\end{table}

\begin{table}[t]
    \centering
    \caption{\textbf{Settings of the specialized methods we compared in the main paper.}  
    For off-policy variBAD, we take the model on Cheetah-Vel as example.}
    \vspace{2mm}
    \begin{tabular}{c|cccccc}
    \toprule
    \multirow{2}{*}{}   & \multirow{2}{*}{Meta-RL} & \multirow{2}{*}{Meta-RL} & ``Standard''   & \multirow{2}{*}{Robust RL} & Generalization & Temporal \\ 
    & & & POMDP & & in RL & credit assignment \\
    \midrule
 \multirow{2}{*}{Approach} & Off-policy  & On-policy & \multirow{2}{*}{VRM} & \multirow{2}{*}{MRPO} & \multirow{2}{*}{EPOPT} & \multirow{2}{*}{IMPALA+SR} \\ 
 & variBAD & variBAD & & & &  \\ 
  Memory-based? & \cmark & \cmark &  \cmark  & \xmark & \xmark & \cmark  \\ 
  Off-policy? & \cmark & \xmark & \cmark & \xmark & \xmark & \cmark \\ 
  Input space & \texttt{oard} & \texttt{oard} & \texttt{oar} & \texttt{s} & \texttt{s} & \texttt{oar} \\
  Access to hidden states? & \xmark & \xmark  & \xmark & \cmark & \cmark & \xmark \\ 
  \bottomrule
    \end{tabular}
    \label{tab:compared_methods}
\end{table}

Fig.~\ref{fig:arch} shows our (separate) recurrent actor-critic architecture (except for temporal credit assignment tasks).
The shortcut from current observation embedding to the MLP may reduce the burden of accurately memorizing it in RNN, and is widely used in prior memory-based architectures~\citep{zintgraf2019varibad,dorfman2020offline,ding2019rlalgorithms,hung2019optimizing}. 
For temporal credit assignment tasks (image-based observations, discrete actions), we adjust the network architectures: replace the MLP observation embedders in actor and critic with two-layer CNNs, remove the observation-action embedder in critic. 

Table~\ref{tab:hparams} shows the main hyperparameters we adopt for each benchmark. We did not tune these hyperparameters, except that we adjusted the number of gradient steps so that all the experiments could be completed in 72 hours.

We store the observed trajectories in the replay buffer we designed (see App.~\ref{sec:code}). 
Each time we sample a (sub)trajectory given the context length. If the actual episode length is smaller than the context length, we zero-pad the (sub)trajectory. 
We use the zero start state strategy~\cite{hausknecht2015deep,kapturowski2018recurrent} for simplicity, \ie use zeros as the initial hidden state of RNNs.

For \textbf{Markovian} policies (SAC and TD3), we remove the embedders and RNNs from the actor-critic architecture, and train them with same hyperparameters as those of recurrent policies. For each task, we report the results of \emph{either} SAC or TD3, whichever achieves higher returns.
For \textbf{oracle} policies, we use the well-tuned results from Table 1 (``SAC w/ unstructured row'') in~\citet{raffin2021smooth} based on Stable Baseline3~\citep{rl-zoo3}, for ``standard" POMDPs. 
For the other benchmarks, we have to run the Markovian policies (SAC and TD3) with access to the hidden states, using the same training hyperparameters as those of recurrent policies. But these oracle policies might be not well-tuned given the same environment and gradient steps, especially in robust RL and generalization in RL. 
In temporal credit assignment benchmark, as the optimal value function is history-dependent, we do not run Markovian policies or oracle policies. 

We also show the settings of the specialized methods we compared in the main paper in Table~\ref{tab:compared_methods}. 
Note that our recurrent model-free RL share exactly the same settings as off-policy variBAD~\cite{dorfman2020offline} and VRM~\cite{han2019variational}.
For MRPO~\cite{jiang2021monotonic} and EPOPT~\cite{rajeswaran2016epopt}, they adopt totally different settings, \ie on-policy Markovian approaches to MDPs (with access to the ground-truth state (\texttt{s}) of environment). 
Thus, in fact, MRPO and EPOPT should be viewed more as \textit{oracle} policies as upper bounds of recurrent model-free RL.

\section{Evaluation Details}
\label{sec:eval_details}

Throughout the experiments, we run each instance/variant in our implementation and each compared method with 4 random seeds. 

There are two steps to select the best single variant of our implementation in each benchmark.
First, we calculate the final performance of each variant by the average performance of the last 20\% environment steps across the 4 seeds. 
Then we select the best variant in terms of the normalized returns, calculated by $\frac{R - R_{\min}}{R_{\max} - R_{\min}} \in [0,1]$, where $R$ is the raw average return of that variant and $R_{\max}$ and $R_{\min}$ are the maximum and minimum of all the methods including oracle policy and random policy.

The bar charts in Fig.~\ref{fig:trailer} and~\ref{fig:pomdp_bar} and Table~\ref{tab:ablation} show the final normalized performance of each method / variant.

\section{Benchmark Details}
\label{sec:env}

We conduct our experiments on 6 benchmarks with 21 environments in total. 

\subsection{``Standard" POMDP Benchmark from VRM}
\label{sec:env_pomdp}

We adopt the occlusion benchmark proposed by VRM, replace the deprecated roboschool with PyBullet~\cite{coumans2016pybullet} as suggested by the official github repository\footnote{\url{https://github.com/openai/roboschool\#deprecated-please-use-pybullet-instead}}.
We follow the practice in VRM~\cite{han2019variational} in the other aspects of environment design, \ie we remove all the position/angle-related entries in the observation space for ``-V" environments and velocity-related entries for ``-P" environments, to transform the original MDP into POMDP.

We also consider the classic Pendulum environment for sanity check in App.~\ref{sec:results_arch}.

\paragraph{$\{$Pendulum, Ant, Cheetah, Hopper, Walker$\}$-P.} The ``-P" stands for the environments that keep position-related entries by removal of velocity-related entries. Thus, the observed state $s^o$ includes positions $p$, while the hidden state $s^h$ is the velocities $v$. 

\paragraph{$\{$Pendulum, Ant, Cheetah, Hopper, Walker$\}$-V.} The ``-V" stands for the environments that keep velocity-related entries by removal of position-related entries. Thus, the observed state $s^o$ includes positions $v$, while the hidden state $s^h$ is the velocities $p$. 

\subsection{Meta-RL Benchmark from Off-Policy VariBAD}
\label{sec:env_meta_off}

For a fair comparison with the same training setting, we directly use the benchmark adopted in off-policy variBAD~\cite{dorfman2020offline}, and limit the number of training tasks as it does. 

\paragraph{Semi-Circle.} The observed state $s^o$ includes the agent's 2D position $p$, and the hidden state $s^h$ is referred to the goal state $p_g$. The goal state only appears in reward function: $R(s^o_{t}, s^o_{t+1}, a_t, s^h) \defeq R(p_{t+1},p_g) =  \mathbbm{1} (\|p_{t+1} - p_g\|_2 \le r)$.  The dynamic function $T$ is independent of the goal state.

\paragraph{Wind.} We modified the parameters of Wind environment in \citet{dorfman2020offline} to make it harder to solve. The agent must navigate to a fixed (but unknown) goal $p_g$ within a distance of $D=1$ from its fixed initial state. 
Similarly to Semi-Circle, the reward function is goal conditioned but without hidden state: $R(s^o_{t}, s^o_{t+1}, a_t, s^h) \defeq R(p_{t+1},p_g) =  \mathbbm{1} (\|p_{t+1} - p_g\|_2 \le r)$. The hidden state $s^h$ appears in the deterministic dynamics as a noise term, \ie  $ s^o_{t+1} = s^o_t +  a_t + s^h $, where $s^h$ is sampled from $U[-0.08,0.08]$ at the initial time-step and then kept fixed.

\paragraph{Cheetah-Vel.} It uses MuJoCo~\cite{todorov2012mujoco} simulator of \texttt{HalfCheetah-v2}. The hidden state $s^h$ is the target \textit{speed} $v_{g}\in \R$ and the observed state $s^o$ includes the velocity $v \in \R$. Reward function includes both the hidden state and action: $R(s^o_{t}, s^o_{t+1}, a_t, s^h) \defeq R(v_t,v_g,a_t) =  -\|v_t - v_g\|_1 - 0.05 \|a_t\|_2^2$. The dynamic function $T$ is independent of the goal state.

\subsection{Meta-RL Benchmark from On-Policy VariBAD}
\label{sec:env_meta_on}

For a fair comparison with the same training setting, we directly use the benchmark adopted in on-policy variBAD~\citep{zintgraf2019varibad}, and \textit{do not} limit the number of training tasks as it does. 

\paragraph{\{Ant, Cheetah, Humanoid\}-Dir.} It uses MuJoCo~\cite{todorov2012mujoco} simulator of \texttt{Ant-v2, HalfCheetah-v2, Humanoid-v2}. The hidden state $s^h$ is the target velocity \textit{direction} $v_g \in \R^2$, and the observed state $s^o$ includes the velocity $v \in \R^2$. The reward function takes both the hidden state and action as inputs: $R(s^o_{t}, s^o_{t+1}, a_t, s^h) \defeq R(v_t,v_g,a_t) =  \langle v_t, v_g \rangle - \alpha \|a_t\|_2^2$ where $\alpha > 0$ is the penalty constant. The dynamic function $T$ is independent of the goal state. Ant-Dir and Cheetah-Dir have only 2 tasks (forward or backward), while Humanoid-Dir samples tasks uniformly from the unit circle.

\subsection{Robust RL Benchmark from MRPO}
\label{env:rmdp}

\paragraph{$\{$Hopper, Walker, Cheetah$\}$-Robust.} We directly adopt the environments used in MRPO~\cite{jiang2021monotonic}. In each environment, the hidden state is the dynamics parameters including the density and friction coefficients of the simulated robot in roboschool, adapted from the SunBlaze~\cite{packer2018assessing}. The exact ranges of the hidden states in each environment can be found in~\citet[Table~1]{jiang2021monotonic}. We evaluate the algorithms with 100 tasks in each environment, and use the average of them as average returns, and the average of the worst $10\%$ of them as worst returns, following the MRPO paper.

\subsection{Generalization in RL Benchmark from SunBlaze}
\label{env:gen}

\paragraph{$\{$Hopper, Cheetah$\}$-Generalize.} We directly adopt the environments used in SunBlaze~\cite{packer2018assessing}. In each environment, the hidden state is the dynamics parameters including the density, friction coefficients, and the power of the simulated robot in roboschool. The exact ranges of both interpolation and extrapolation in the hidden state distribution for each environment can be found in~\citet[Table~1]{packer2018assessing}. 
We follow the practice of SunBlaze to evaluate the interpolation and extrapolation success rates.

\subsection{Temporal Credit Assignment Benchmark from IMPALA+SR}

\paragraph{Delayed-Catch, Key-to-Door.} We directly adopted the two environments from IMPALA+SR paper~\citep[Sec.~3.2 and 3.3]{raposo2021synthetic}. In both environments, they have discrete action spaces (3 and 4 actions) and pixels as observations ($1\times 7 \times 7$ and $3\times 5 \times 5$). The reward functions are trajectory-level and sparse. 

In Delayed-Catch, there are 40 runs in each episode, with a total length of around 280. 
The agent will only receive a non-zero reward at the end of each episode, which is the total number of successful runs, thus the optimal terminal reward is 40. 

In Key-to-Door, there are three phases in one episode. In the first phase, the agent can pick up a key, but no reward will be given. In the second phase, the agent can pick up apples to get rewards. In the third phase, the agent can open the door, only if it has picked up the key in the first phase (the agent cannot see the key after the first phase), to get a reward bonus. Thus the final reward bonus depends on the agent's action that happens in the distant past. We follow the prior work to report the success rate of opening the door as the evaluation metric.

\section{Full Experimental Results}
\label{sec:full_results}

\subsection{Learning Curves of All the Compared Methods}
\label{sec:learning_curves}
\begin{figure}[h]
    \centering
    \includegraphics[width=0.49\textwidth]{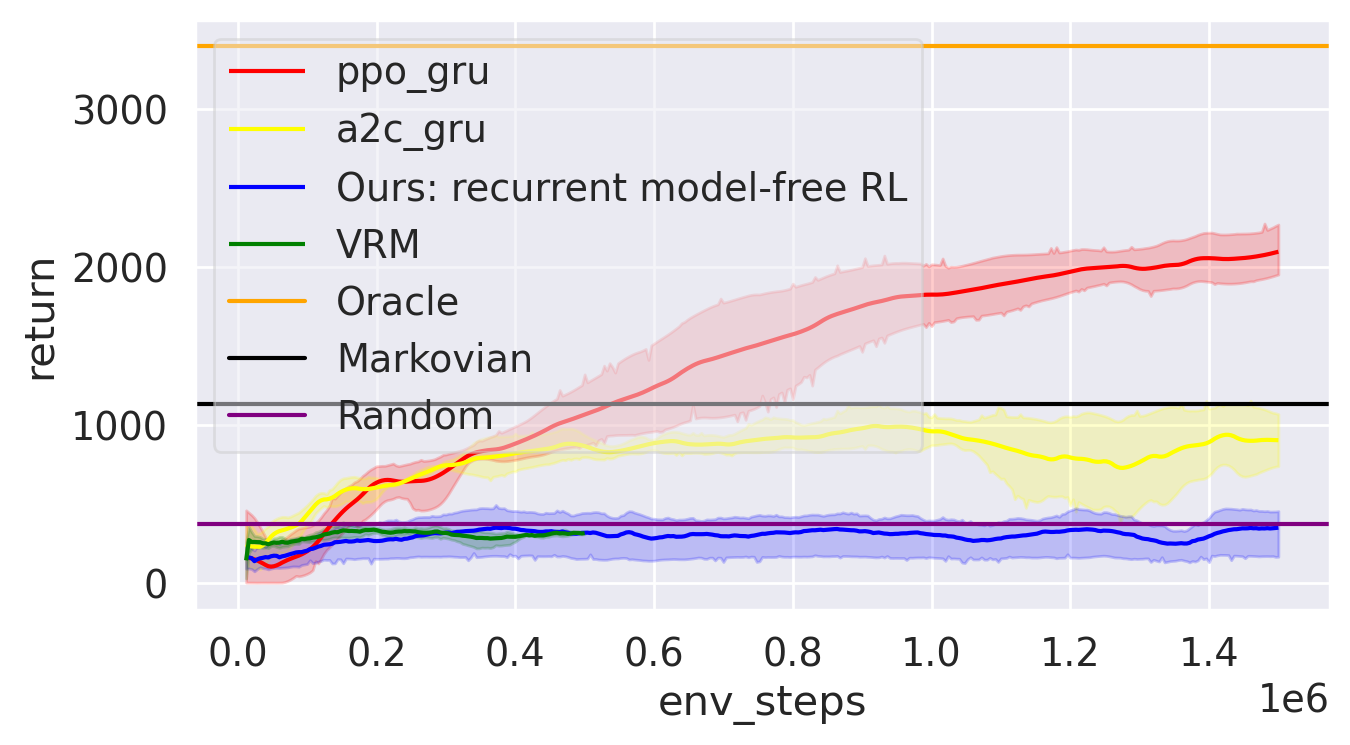} 
    \includegraphics[width=0.49\textwidth]{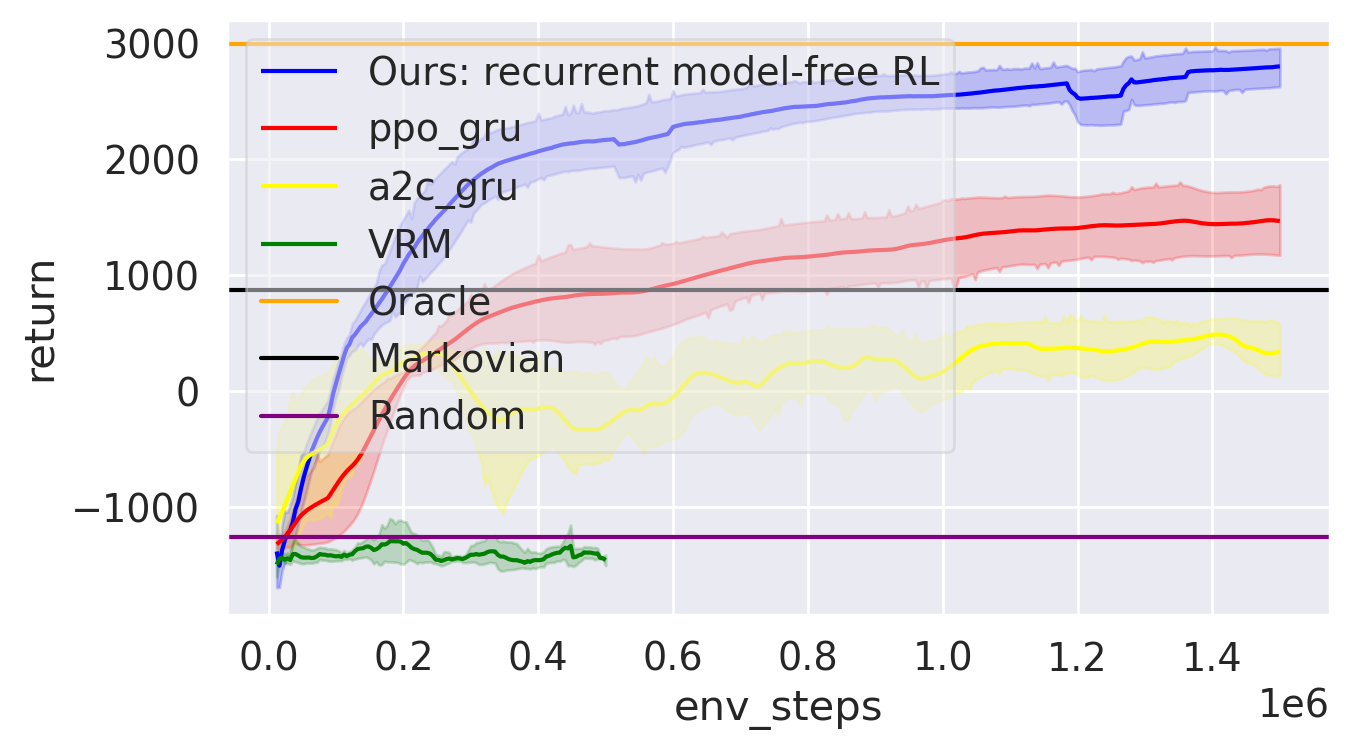} 
    \begin{tabular}{P{0.49\linewidth}P{0.49\linewidth}}
    Ant-P &
    Cheetah-P
    \end{tabular}
    \includegraphics[width=0.49\textwidth]{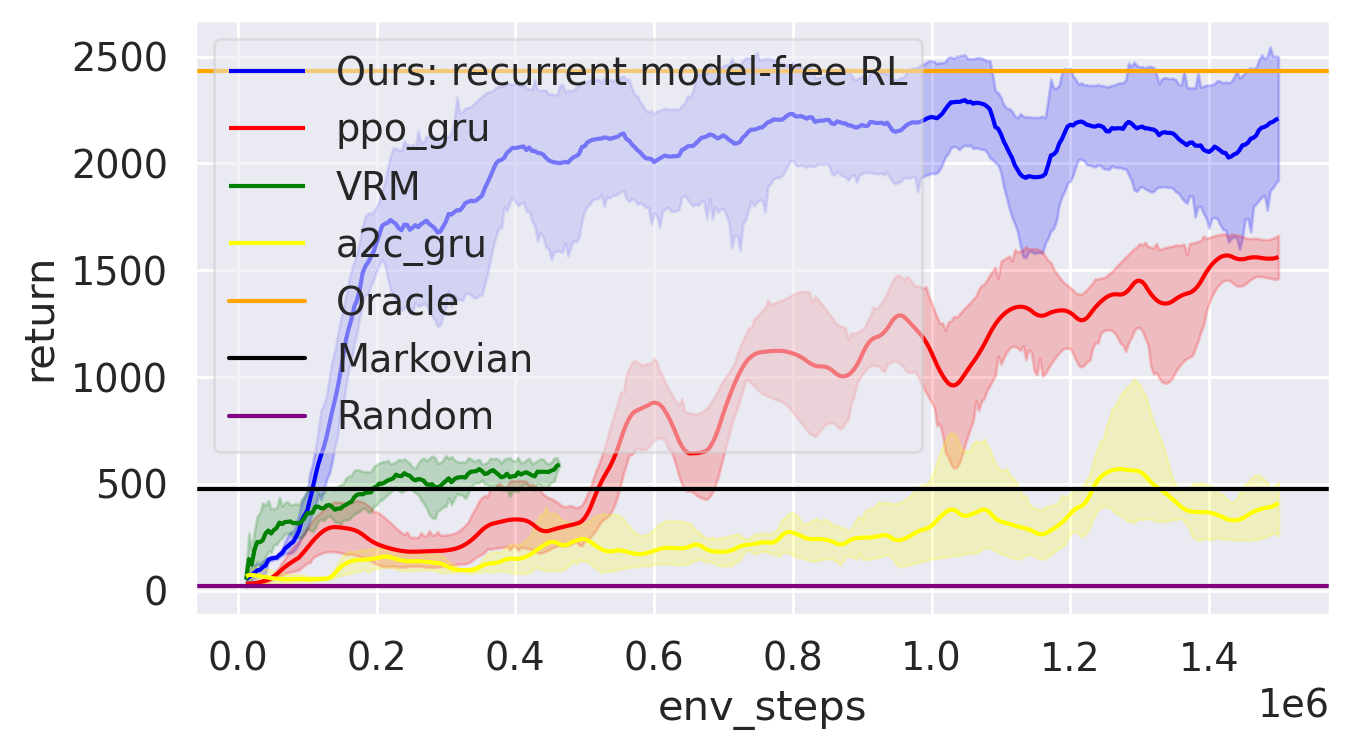} 
    \includegraphics[width=0.49\textwidth]{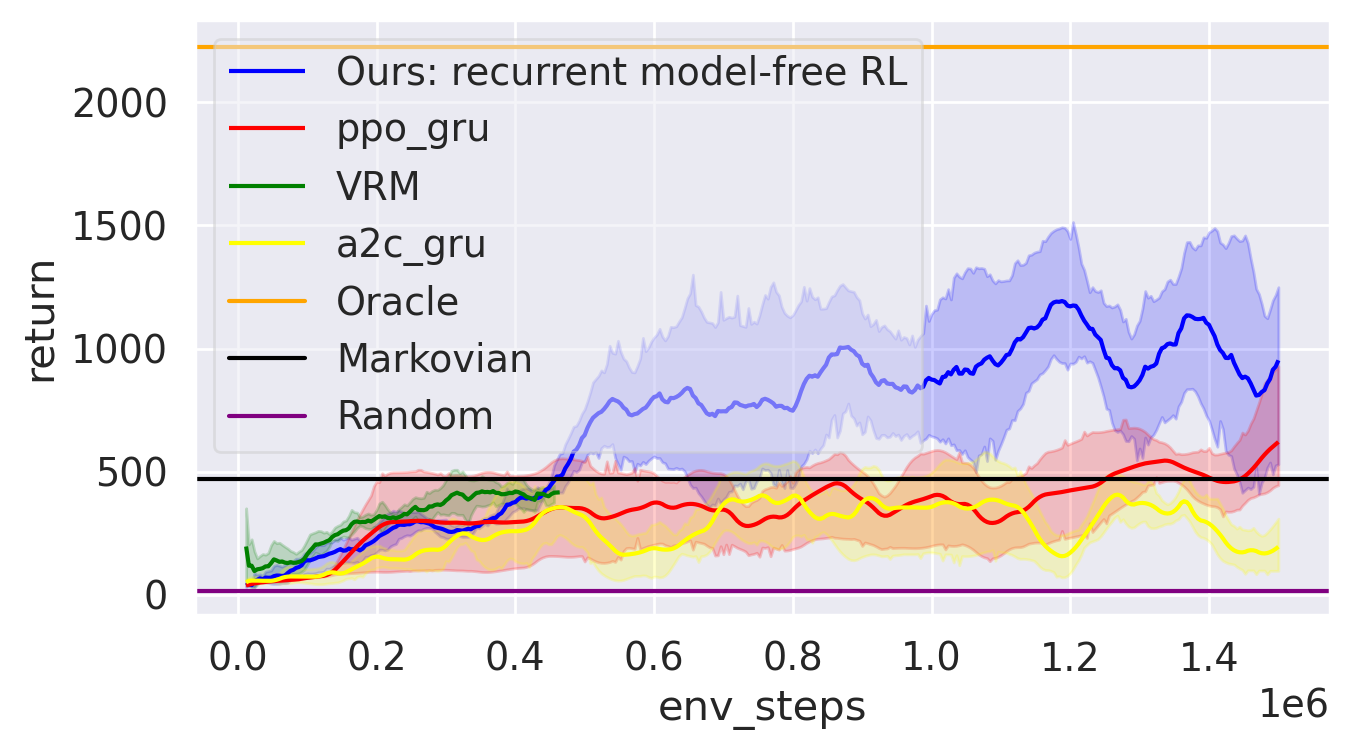} 
    \begin{tabular}{P{0.49\linewidth}P{0.49\linewidth}}
    Hopper-P &
    Walker-P 
    \end{tabular}
    \vspace{-3mm}
    \caption{\textbf{Learning curves on ``standard" POMDP benchmark that preserves positions \& angles  but occludes velocities  in the states (namely ``-P").} 
    We show the results from the \textbf{single best variant} of our implementation on recurrent model-free RL, the popular recurrent  model-free  on-policy implementation (PPO-GRU, A2C-GRU)~\cite{kostrikov2018pytorch}, and also model-based method VRM~\cite{han2019variational}. Note that VRM is around 5x slower than ours, so we have to run 0.5M environment steps for it. 
    Given 0.5M steps budget, our implementation is at least comparable to (if not greatly surpasses) the specialized method VRM \textbf{on all the 4 environments}.}
    \label{fig:pomdp_best_P}
\end{figure}

\begin{figure}[h]
    \centering
    \includegraphics[width=0.49\textwidth]{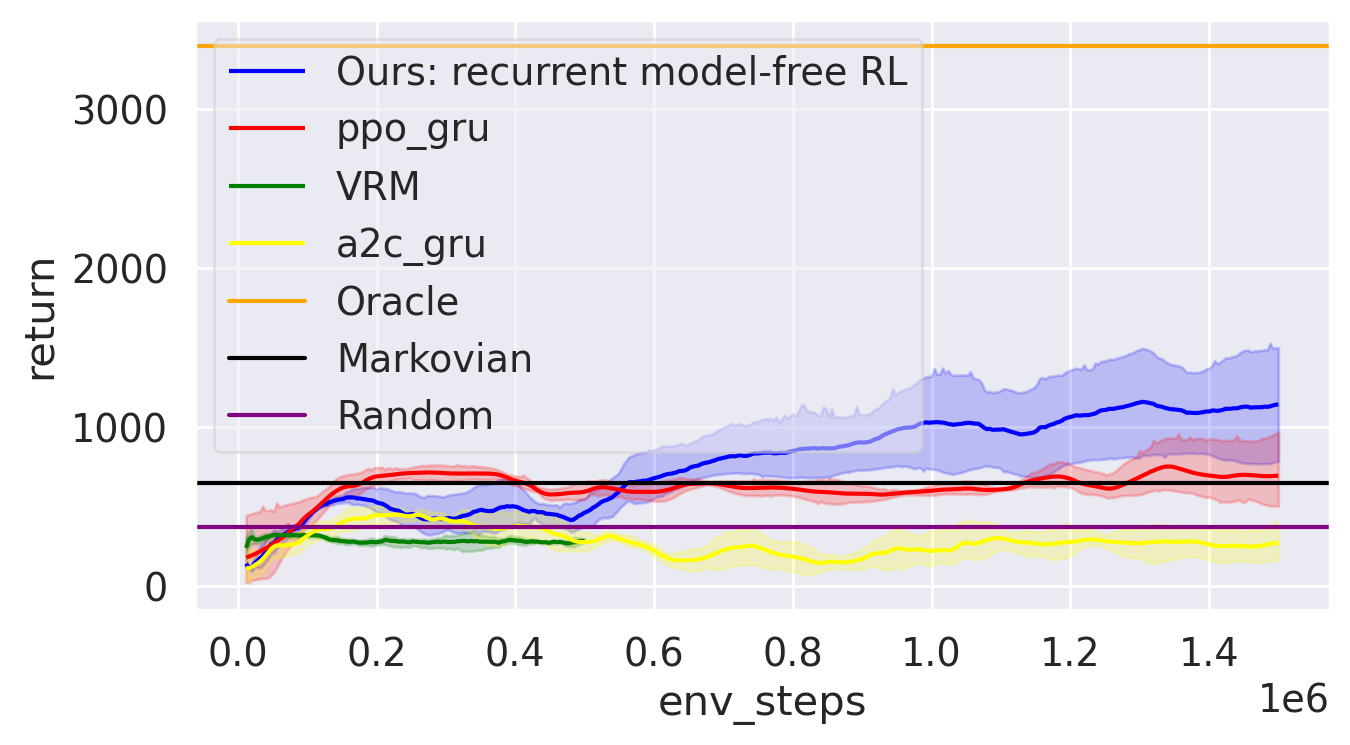} 
    \includegraphics[width=0.49\textwidth]{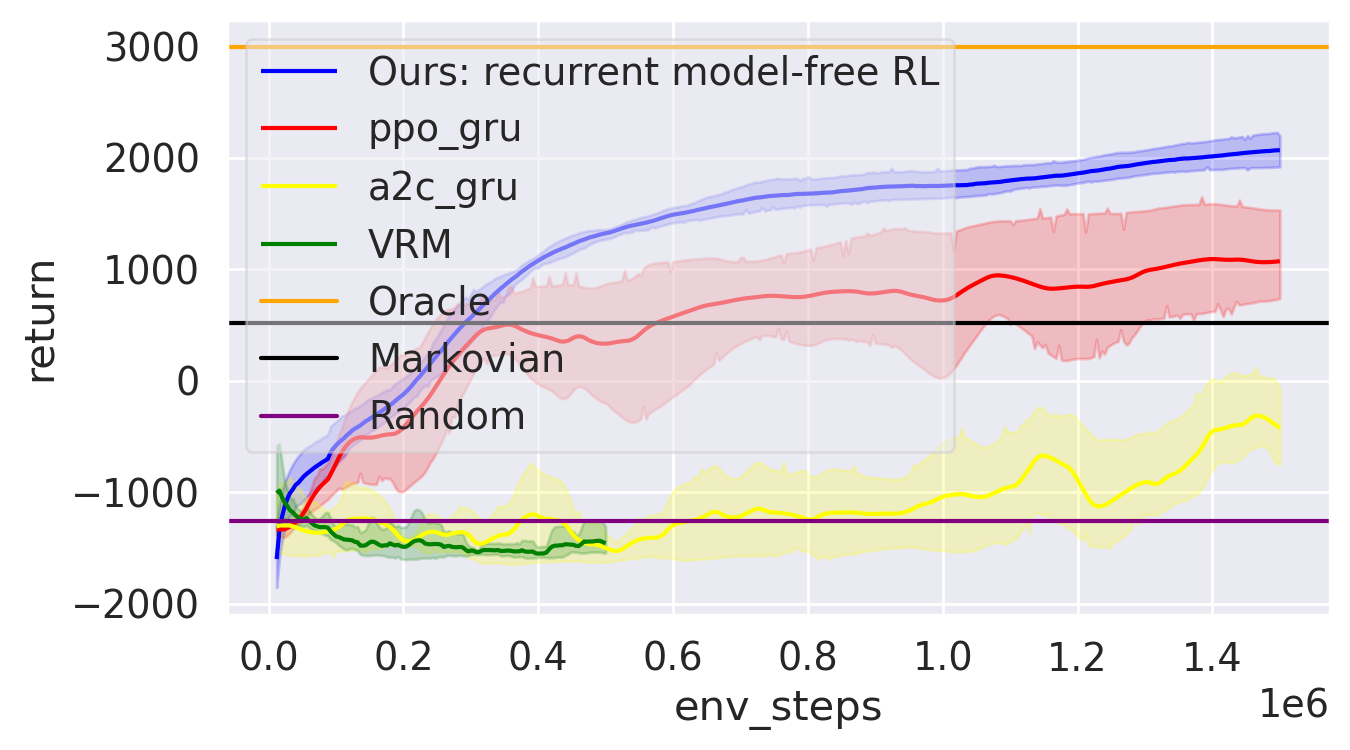} 
    \begin{tabular}{P{0.49\linewidth}P{0.49\linewidth}}
    Ant-V &
    Cheetah-V 
    \end{tabular}
    \includegraphics[width=0.49\textwidth]{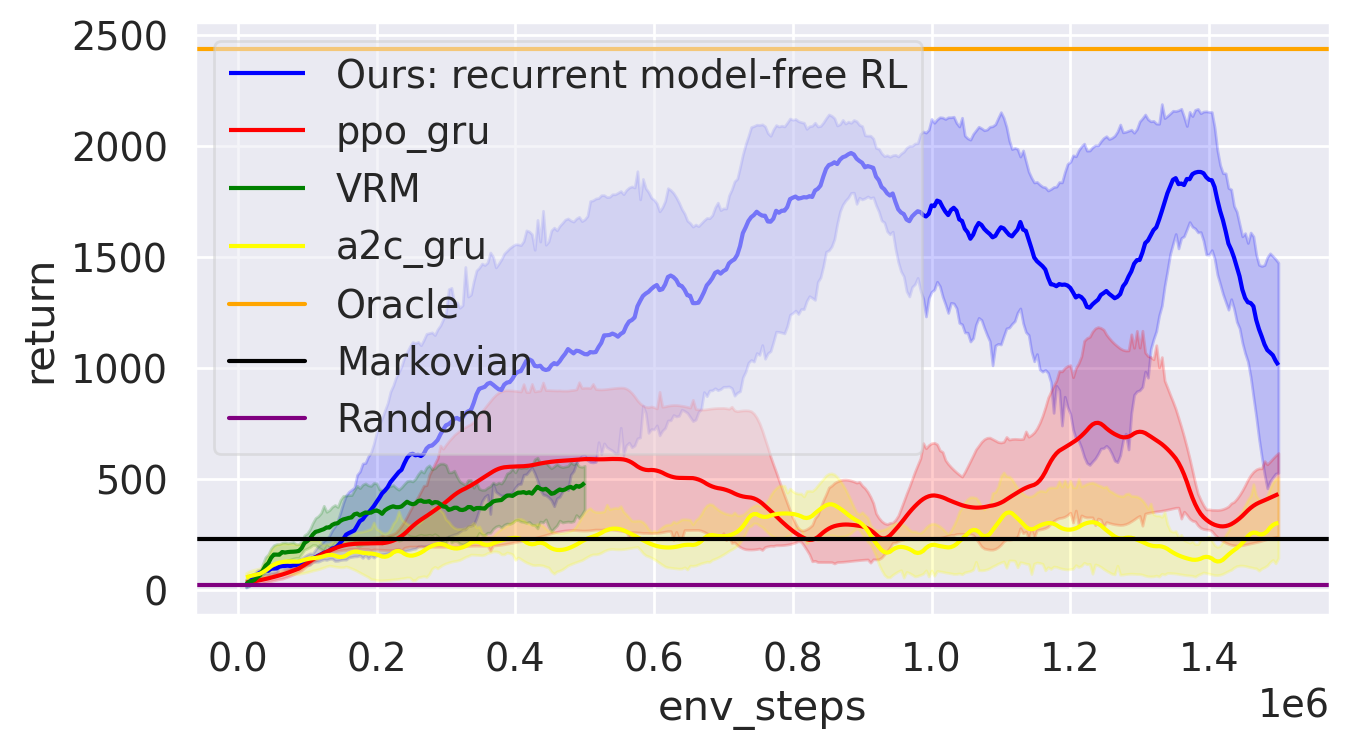} 
    \includegraphics[width=0.49\textwidth]{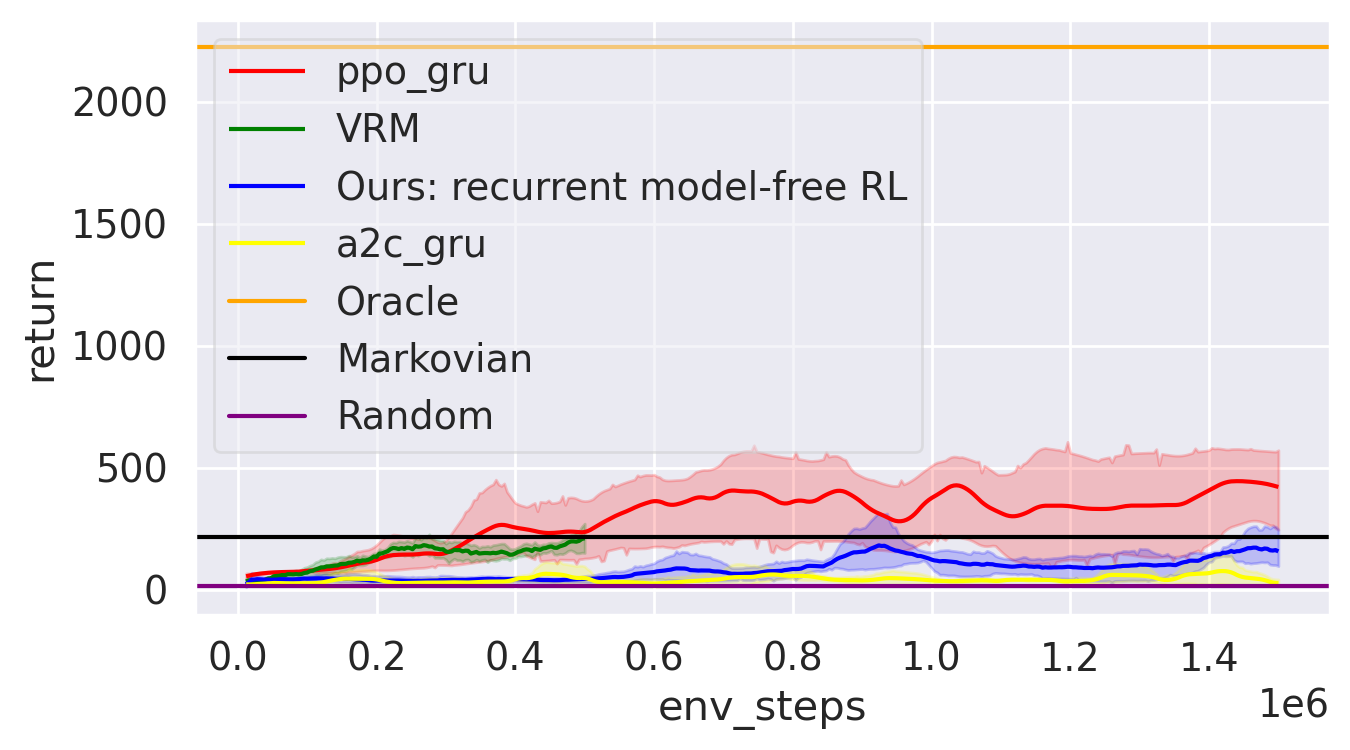} 
    \begin{tabular}{P{0.49\linewidth}P{0.49\linewidth}}
    Hopper-V &
    Walker-V 
    \end{tabular}
    \vspace{-3mm}
    \caption{\textbf{Learning curves on ``standard" POMDP benchmark that preserves velocities but occludes positions \& angles  in the states (namely ``-V").} 
    We show the results from the \textbf{single best variant} of our implementation on recurrent model-free RL, the popular recurrent  model-free  on-policy implementation (PPO-GRU, A2C-GRU)~\cite{kostrikov2018pytorch}, and also model-based method VRM~\cite{han2019variational}. Note that VRM is around 5x slower than ours, so we have to run 0.5M environment steps for it. 
    Given 0.5M steps budget, our implementation is at least comparable to (if not greatly surpasses) the specialized method VRM \textbf{on 3 out of the 4 environments}.}
    \label{fig:pomdp_best_V}
\end{figure}

\begin{figure}[h]
    \centering
    \includegraphics[width=0.49\textwidth]{figs/results/meta/PointRobotSparse-v0/instance-td3-lstm-64-or-separate-max_xNone-last0.8-window10-othersoffpolicy-varibad_oracle_sac_Markovian_sac.png} 
    \includegraphics[width=0.49\textwidth]{figs/results/meta/Wind-v0/instance-td3-lstm-64-oa-separate-max_xNone-last0.8-window10-othersoffpolicy-varibad_oracle_td3_Markovian_sac.png} 
    \begin{tabular}{P{0.49\linewidth}P{0.49\linewidth}}
    Semi-Circle & Wind 
    \end{tabular}
    \includegraphics[width=0.49\textwidth]{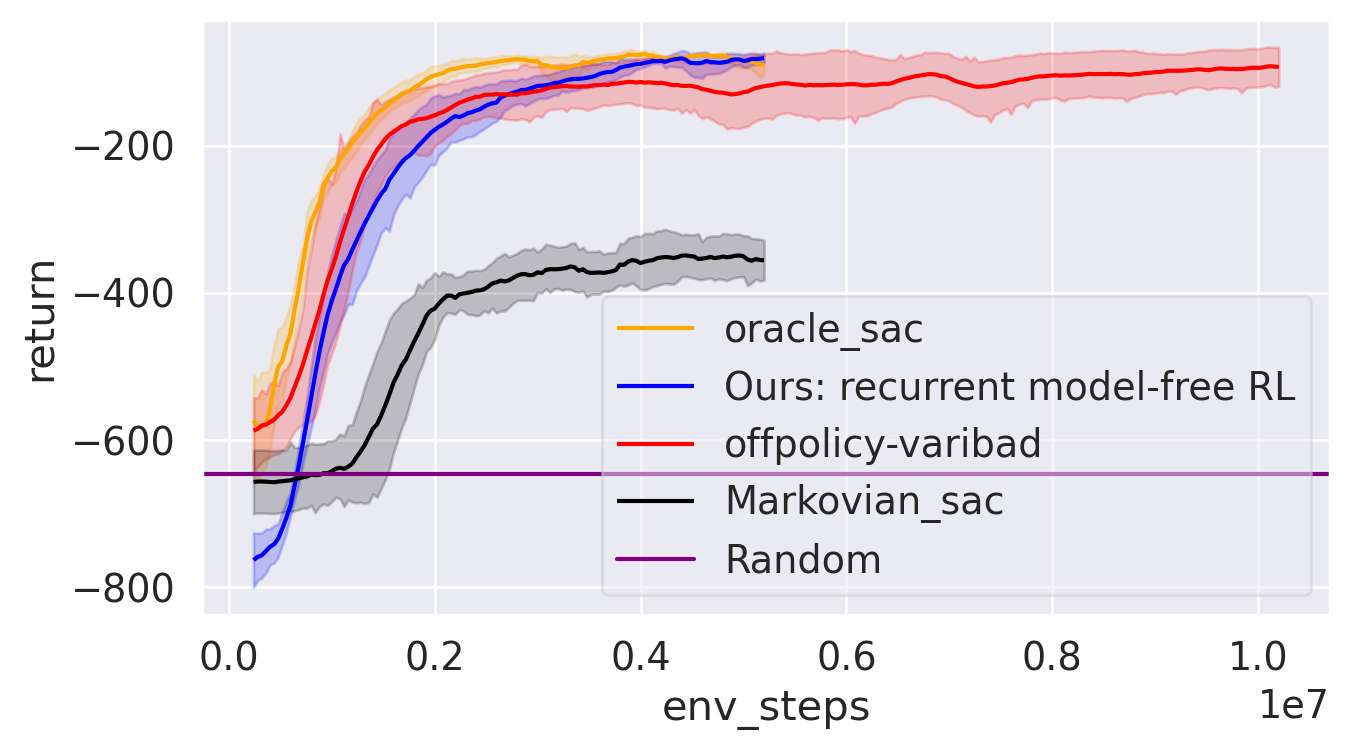} \\
    Cheetah-Vel 
    \caption{\textbf{Learning curves on meta-RL benchmark adopted in \textit{off}-policy variBAD paper~\citep{dorfman2020offline}.} 
    We show the results from the \textbf{single best variant} of our implementation on recurrent model-free RL, and the specialized meta-RL method off-policy variBAD~\citep{dorfman2020offline}.
    With better sample efficiency, our implementation is at least comparable to (if not greatly surpasses) the specialized method off-policy variBAD \textbf{on all the 3 environments}.}
    \label{fig:meta_off}
\end{figure}

\begin{figure}[h]
    \centering
    \includegraphics[width=0.49\textwidth]{figs/results/meta/Cheetah-Dir/instance-sac-gru-400-oar-separate-max_x100000000-last0.8-window10-othersonpolicy-varibad_rl2_oracle_ppo_Markovian_sac_oracle_sac.png} 
    \includegraphics[width=0.49\textwidth]{figs/results/meta/AntDir-v0/instance-sac-gru-400-oar-separate-max_x100000000-last0.8-window10-othersonpolicy-varibad_rl2_oracle_ppo_oracle_sac_Markovian_td3.png} 
    \begin{tabular}{P{0.49\linewidth}P{0.49\linewidth}}
      Cheetah-Dir & Ant-Dir  
    \end{tabular}
    \includegraphics[width=0.49\textwidth]{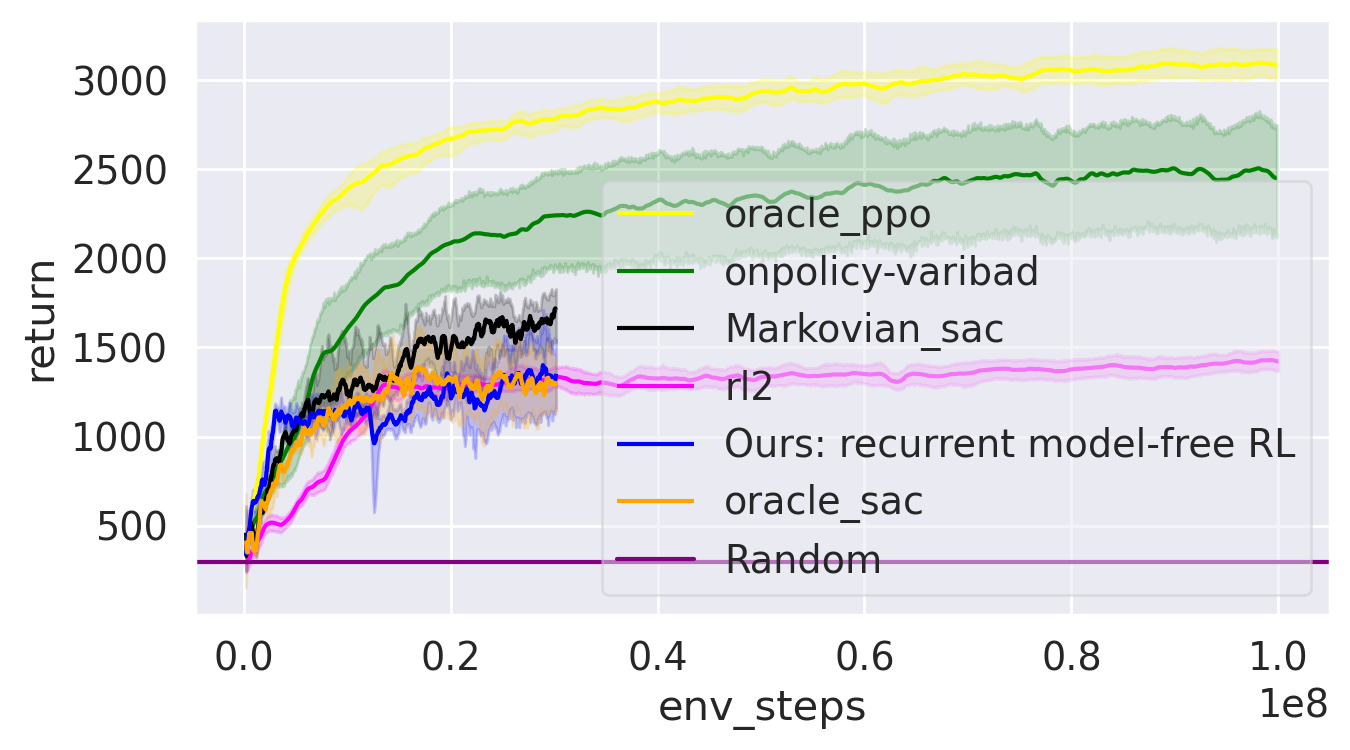} 
     \\
    Humanoid-Dir  \\ 
    \caption{\textbf{Learning curves on meta-RL benchmark adopted in \textit{on}-policy variBAD paper~\citep{zintgraf2019varibad}.} 
    We show the results from the \textbf{single best variant} of our implementation on recurrent model-free RL, RL2~\citep{duan2016rl}, and the specialized meta-RL method on-policy variBAD~\citep{zintgraf2019varibad}. 
    We also show the learning curves of oracle PPO, off-policy oracle, off-policy Markovian policies for reference.
    We directly use the open-sourced learning curve data from~\url{https://github.com/lmzintgraf/varibad\#results} for oracle PPO, RL2, and on-policy variBAD.
    Our implementation is at least comparable to (if not greatly surpasses) the specialized method on-policy variBAD \textbf{on 1 out of the 3 environments}.}
    \label{fig:meta_on}
\end{figure}

\begin{figure}[h]
    \centering
    \includegraphics[width=\textwidth]{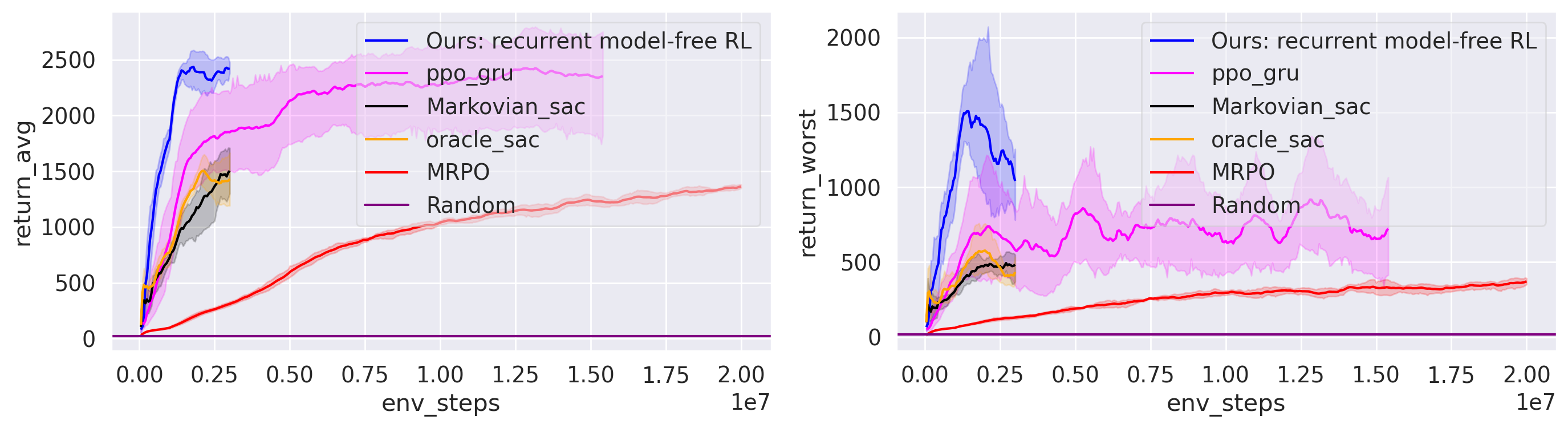} \\ Hopper-Robust \\
    \includegraphics[width=\textwidth]{figs/results/rmdp/cheetah-robust/instance-td3-lstm-64-o-separate-max_x15000000-last0.8-window20-othersMRPO_ppo_gru_oracle_sac_Markovian_sac.png} \\ Cheetah-Robust \\
    \includegraphics[width=\textwidth]{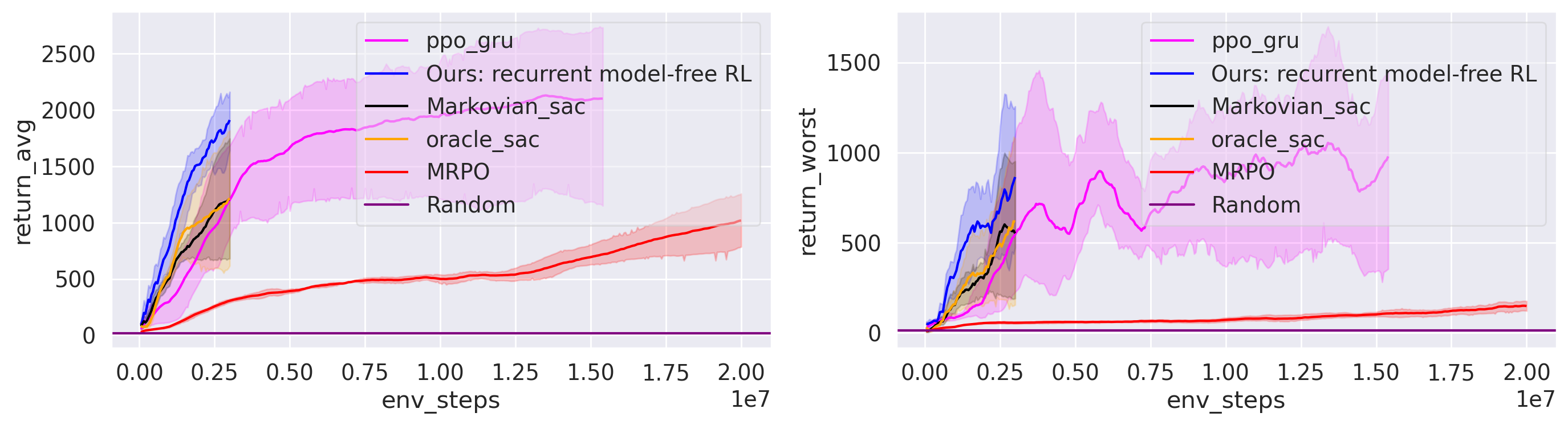} \\ Walker-Robust \\
    \caption{\textbf{Learning curves on robust RL benchmark.} 
    We show the average returns (left figures) and worst returns (right figures) from the \textbf{single best variant} of our implementation on recurrent model-free RL, the specialized robust RL method MRPO~\cite{jiang2021monotonic}, and recurrent PPO. Note that our implementation is much slower than MRPO and recurrent PPO, so we have to run our implementation within 3M environment steps.
    With better sample efficiency, our implementation is at least comparable to (if not greatly surpasses) the specialized method MRPO \textbf{on all the 3 environments}.}
    \label{fig:rmdp_best}
\end{figure}

\begin{figure}[h]
    \centering
    \includegraphics[width=\textwidth]{figs/results/generalize/hopper-generalize/instance-td3-lstm-64-o-separate-max_xNone-last0.8-window10-othersoracle_sac_Markovian_sac.png} \\ Hopper-Generalize \\
    \includegraphics[width=\textwidth]{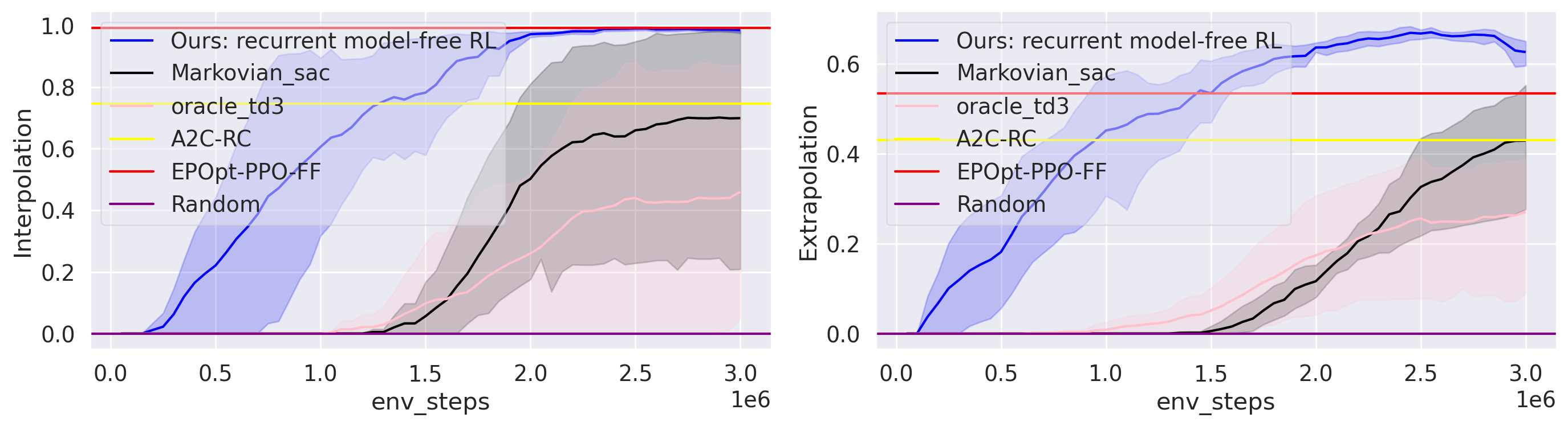} \\ Cheetah-Generalize \\
    \caption{\textbf{Learning curves on generalization in RL benchmark.}
    We show the interpolation success rates (left figures) and extrapolation success rates (right figures) from the \textbf{single best variant} of our implementation on recurrent model-free RL. We also show the final performance of the specialized method EPOpt-PPO-FF~\cite{rajeswaran2016epopt} and another recurrent model-free (on-policy)  RL method (A2C-RC) copied from the Table 7 \& 8 in \citet{packer2018assessing}.
    Our implementation is at least comparable to (if not greatly surpasses) the specialized method EPOpt-PPO-FF \textbf{on both the 2 environments}.}
    \label{fig:generalize_best}
\end{figure}

\begin{figure}
    \centering
    \footnotesize
    \includegraphics[width=0.49\linewidth]{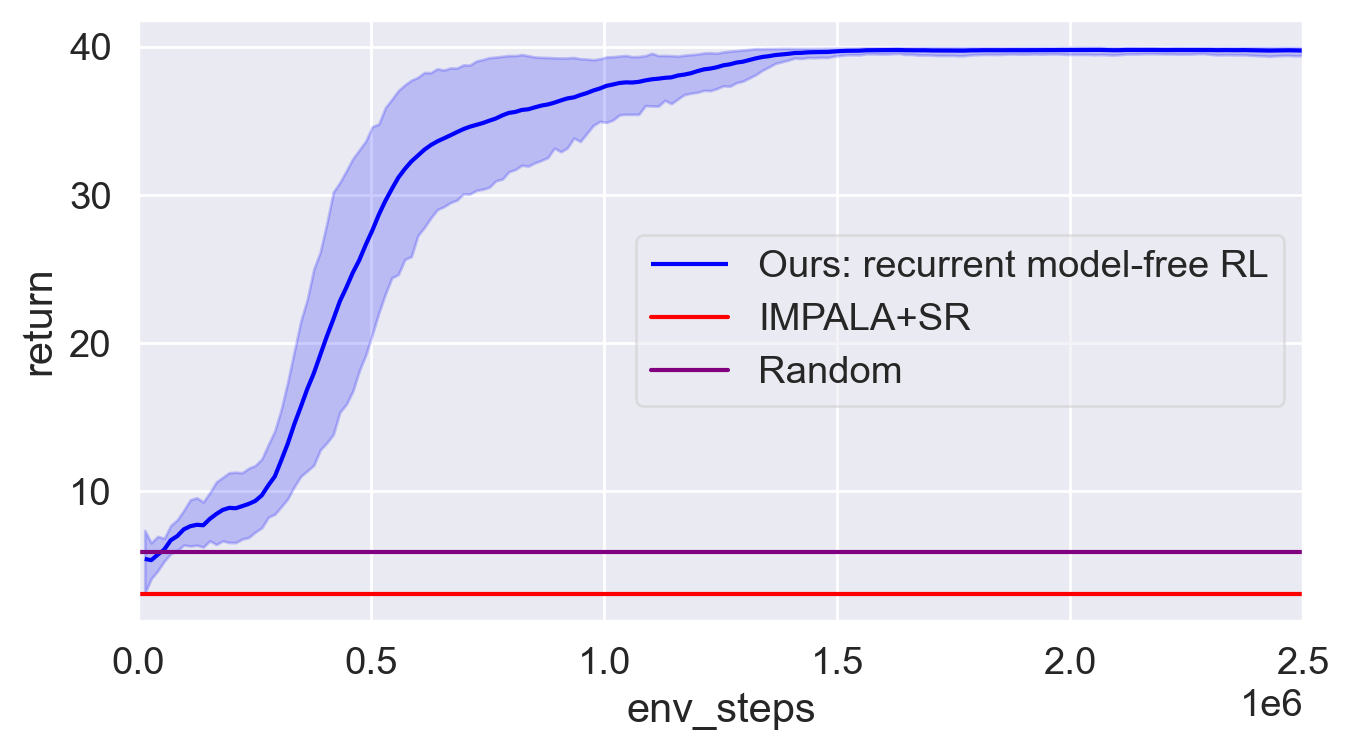}
    \includegraphics[width=0.49\linewidth]{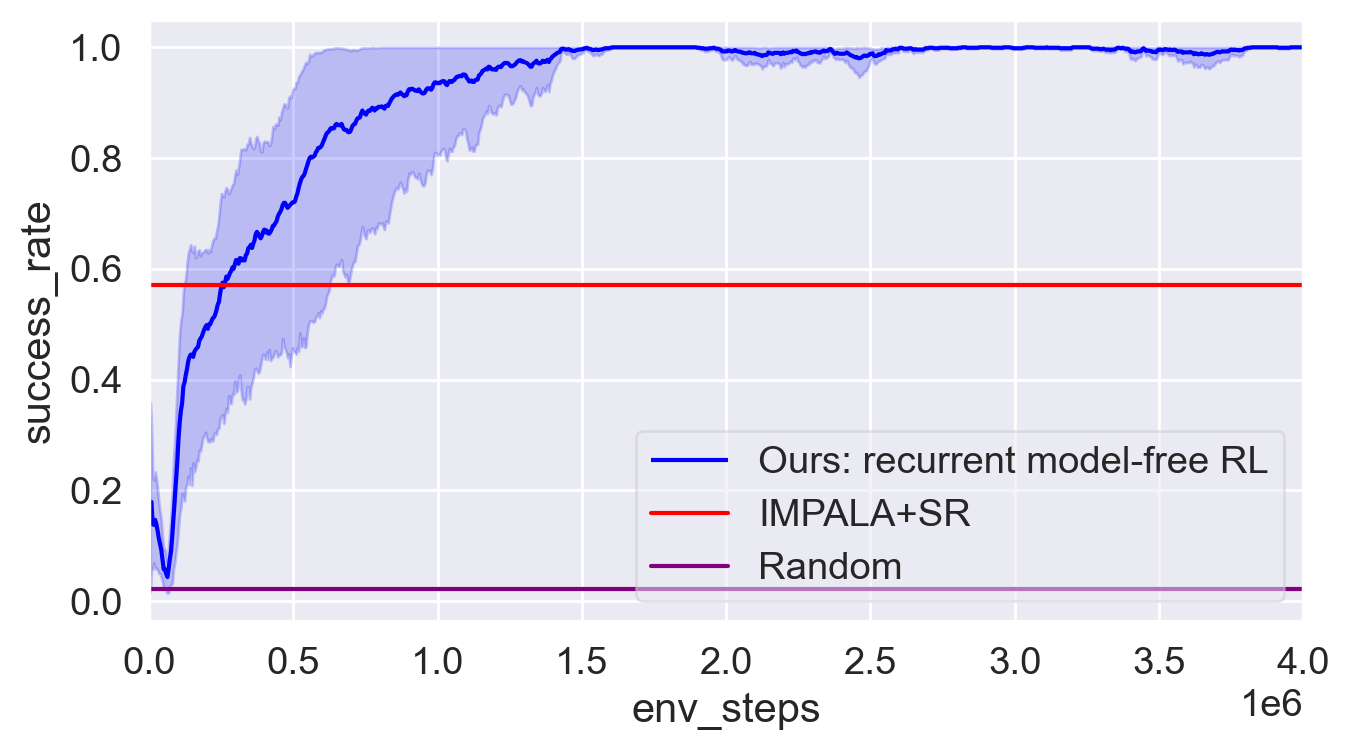}
    \begin{tabular}{P{0.49\linewidth}P{0.49\linewidth}}
Delayed-Catch     &  Key-to-Door
\end{tabular}
    \caption{
\textbf{Learning curves on temporal credit assignment benchmark.}
We show the total rewards for Delayed-Catch and the success rates of opening the door for Key-to-Door, from the \textbf{single best variant} of our implementation on recurrent model-free RL.
We also show the performance of the specialized method IMPALA+SR at 2.5M and 4M steps, respectively.
Our implementation is at least comparable to (if not greatly surpasses) the specialized method IMPALA+SR \textbf{on both the 2 environments}.}
    \label{fig:credit}
\end{figure}

In this subsection, we show all the learning curves of all the compared methods (including oracle policy as upper bound, Markovian and random policies as lower bounds) in each benchmark, namely ``standard" POMDPs (Fig.~\ref{fig:pomdp_best_P} and Fig.~\ref{fig:pomdp_best_V}), meta-RL (Fig.~\ref{fig:meta_off} and Fig.~\ref{fig:meta_on}), robust RL (Fig.~\ref{fig:rmdp_best}), generalization in RL (Fig.~\ref{fig:generalize_best}), and temporal credit assignment (Fig.~\ref{fig:credit}).

\begin{table}[t]
    \centering
    \small
    \caption{\textbf{Numerical results of our final performance.} The best single variant follows the notation in App.~\ref{sec:setting}.
    The performance column shows the mean and standard deviation of the metric (averaged at the last 20\% of the total environment steps) across the 4 seeds.}
    \vspace{2mm}
    \begin{tabular}{cccccc}
    \toprule
       Benchmark  & Best single variant & Environment & Env steps & Metric & Performance \\ \midrule
      \multirow{8}{*}{\vtop{\hbox{\strut ``Standard''}\hbox{\strut  POMDP}}}   & {\footnotesize\multirow{8}{*}{\texttt{td3-gru-64-oa-separate}}} & Ant-P & \multirow{8}{*}{1.5M} & \multirow{8}{*}{Avg return} & 348 $\pm$ 282 \\
      &&Ant-V &  &&  1113 $\pm$ 360  \\
      &&Cheetah-P &&  & 2693 $\pm$ 219  \\
      &&Cheetah-V &&  & 1980 $\pm$ 143  \\
      &&Hopper-P & && 2133 $\pm$ 326  \\
      &&Hopper-V & && 1495 $\pm$ 381  \\
      &&Walker-P & && 982 $\pm$ 339  \\
      &&Walker-V & && 121 $\pm$ 52  \\ \midrule
      
      \multirow{6}{*}{Meta-RL} & {\footnotesize\texttt{td3-lstm-64-ord-separate}} & Semi-Circle & 1.5M & \multirow{6}{*}{Avg return} & 94.2 $\pm$ 0.6 \\ 
     &{\footnotesize\texttt{td3-lstm-64-oad-separate}} & Wind & 0.75M &  & 62.8 $\pm$ 0.5 \\ 
    &{\footnotesize\texttt{td3-lstm-64-oard-separate}} & Cheetah-Vel & 5M & & -84.7 $\pm$ 11.1 \\ 
    &\multirow{3}{*}{{\footnotesize\texttt{sac-gru-max-oard-separate}}} & Ant-Dir & 30M & & 1886 $\pm$ 177  \\ 
    & & Cheetah-Dir & 20M & & 4189 $\pm$ 282  \\ 
    & & Humanoid-Dir & 30M & & 1322 $\pm$ 257  \\ \midrule
    
    \multirow{6}{*}{Robust RL} & \multirow{6}{*}{{\footnotesize\texttt{td3-lstm-64-o-separate}}} & \multirow{2}{*}{Cheetah-Robust} &  \multirow{6}{*}{3M} & Avg return &  2278 $\pm$ 454 \\
     &  &  &   & Worst return &  1587 $\pm$ 355 \\
    & &\multirow{2}{*}{Hopper-Robust} &  & Avg return &  2392 $\pm$ 127 \\
     & & &    & Worst return &  1169 $\pm$ 304 \\
    & &\multirow{2}{*}{Walker-Robust} &  & Avg return &  1807 $\pm$ 347 \\
     & & &    & Worst return &  766 $\pm$ 504 \\ \midrule

    \multirow{4}{*}{\vtop{\hbox{\strut Generalization}\hbox{\strut in RL}}}  & \multirow{4}{*}{{\footnotesize\texttt{td3-lstm-64-o-separate}}} & \multirow{2}{*}{Cheetah-Generalize} &  \multirow{4}{*}{3M} & Interpolation &  0.989 $\pm$ 0.008 \\
     &  &  &   & Extrapolation &  0.656 $\pm$ 0.011 \\
    & &\multirow{2}{*}{Hopper-Generalize} &  & Interpolation  &  0.757 $\pm$ 0.138 \\
     & & &    & Extrapolation &  0.299 $\pm$ 0.029 \\ \midrule

    \multirow{2}{*}{\vtop{\hbox{\strut Temporal}\hbox{\strut credit assignment}}}  & \multirow{2}{*}{{\footnotesize\texttt{sacd-lstm-max-o-separate}}} & Delayed-Catch&  2.5M & Avg return &  39.8 $\pm$ 0.4 \\
    & &Key-to-Door & 4M & Success rate  &  0.996 $\pm$ 0.009 \\ %
        \bottomrule
    \end{tabular}
    \label{tab:numerics}
\end{table}

\subsection{Single Factor Analysis on Our Implementation}
\label{sec:single_factor}

Our analysis will focus on ablating the important design decisions: the actor-critic architecture (\texttt{Arch}), the agent input space (\texttt{Inputs}), the underlying model-free RL algorithm (\texttt{RL}), the RNN encoder (\texttt{Encoder}), and the RNN context length (\texttt{Len}).

\begin{figure}[h]
    \centering
    \includegraphics[width=0.49\textwidth]{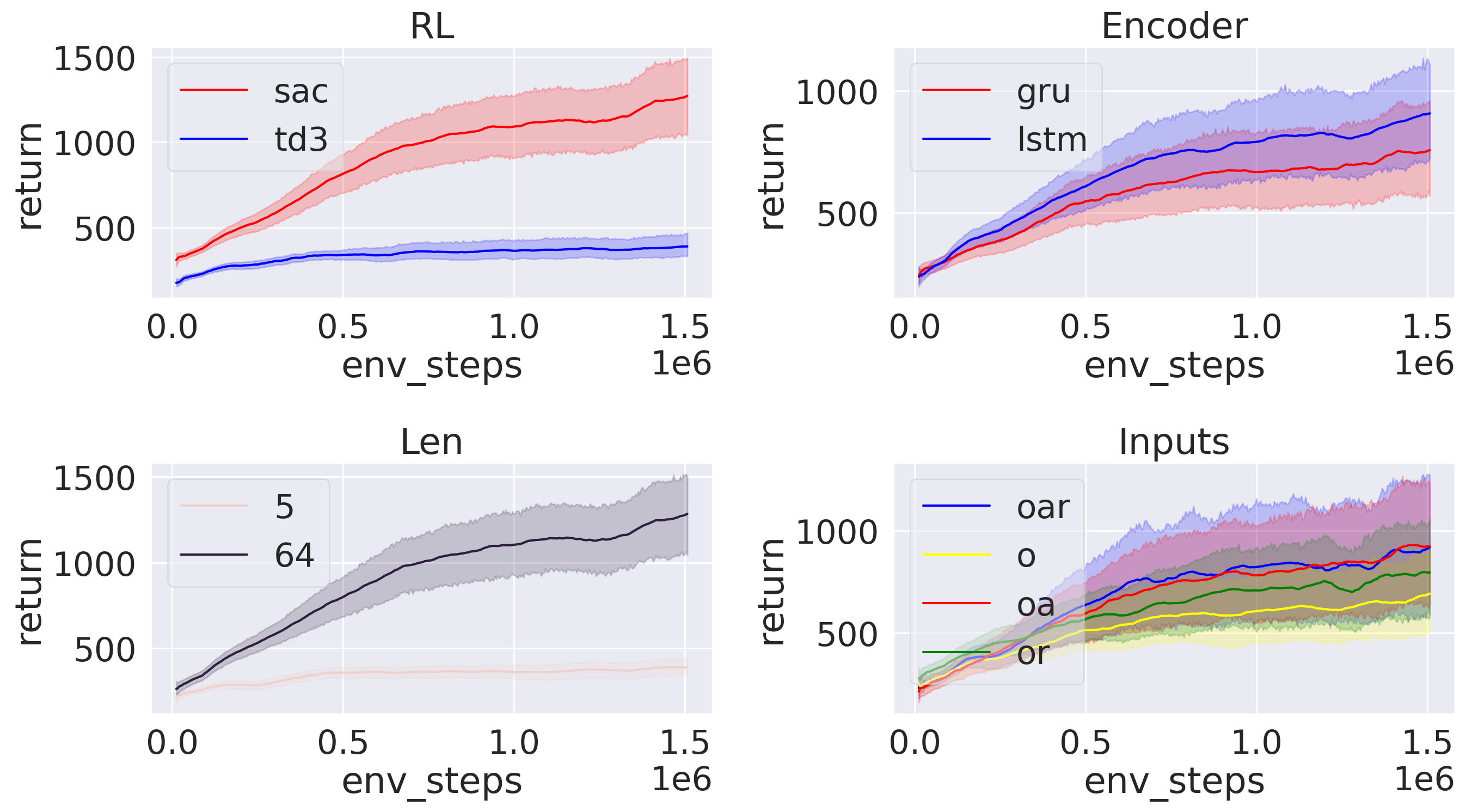} 
    \includegraphics[width=0.49\textwidth]{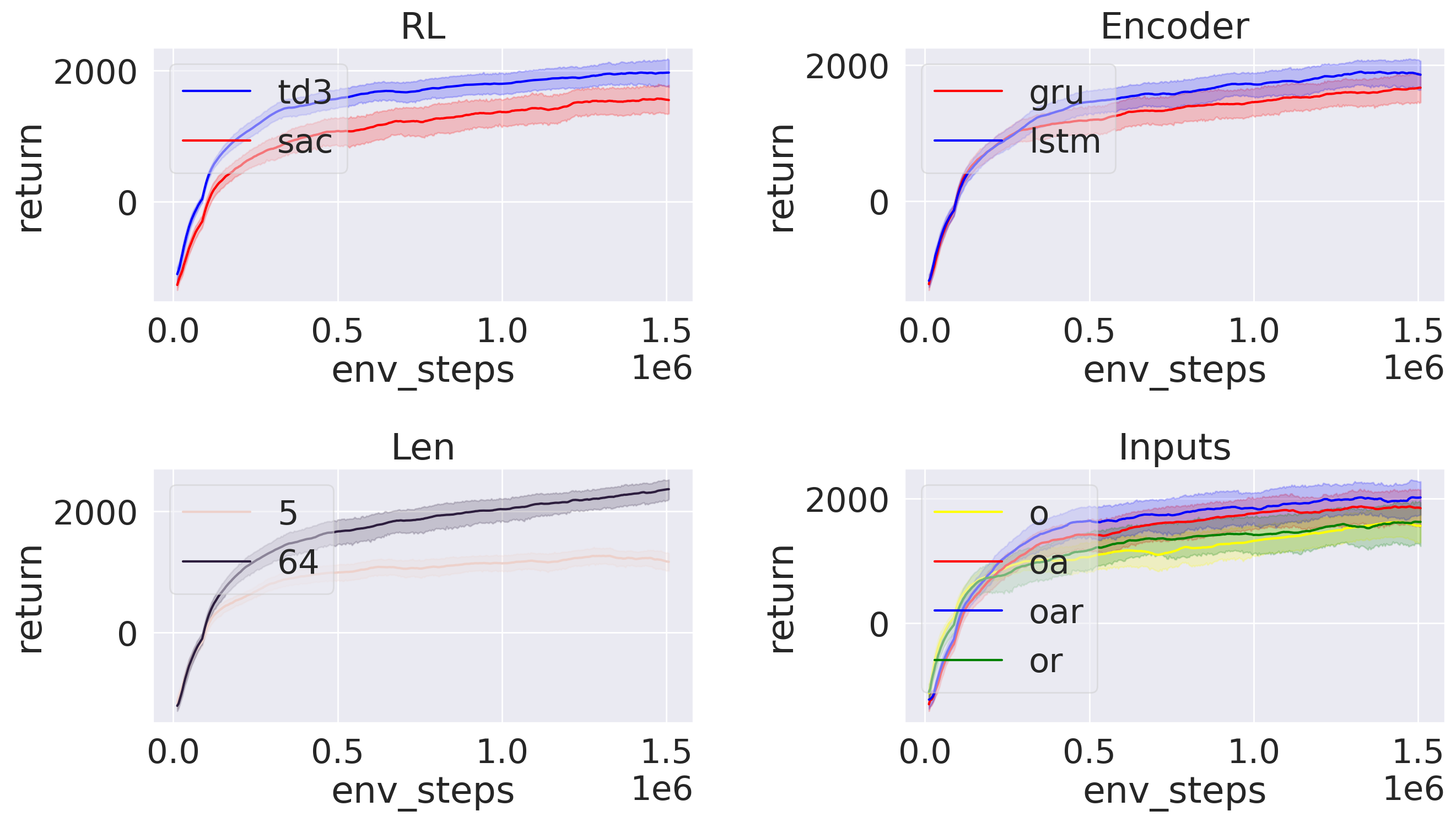} 
    \begin{tabular}{P{0.49\linewidth}P{0.49\linewidth}}
    Ant-P &
    Cheetah-P
    \end{tabular}
    \includegraphics[width=0.49\textwidth]{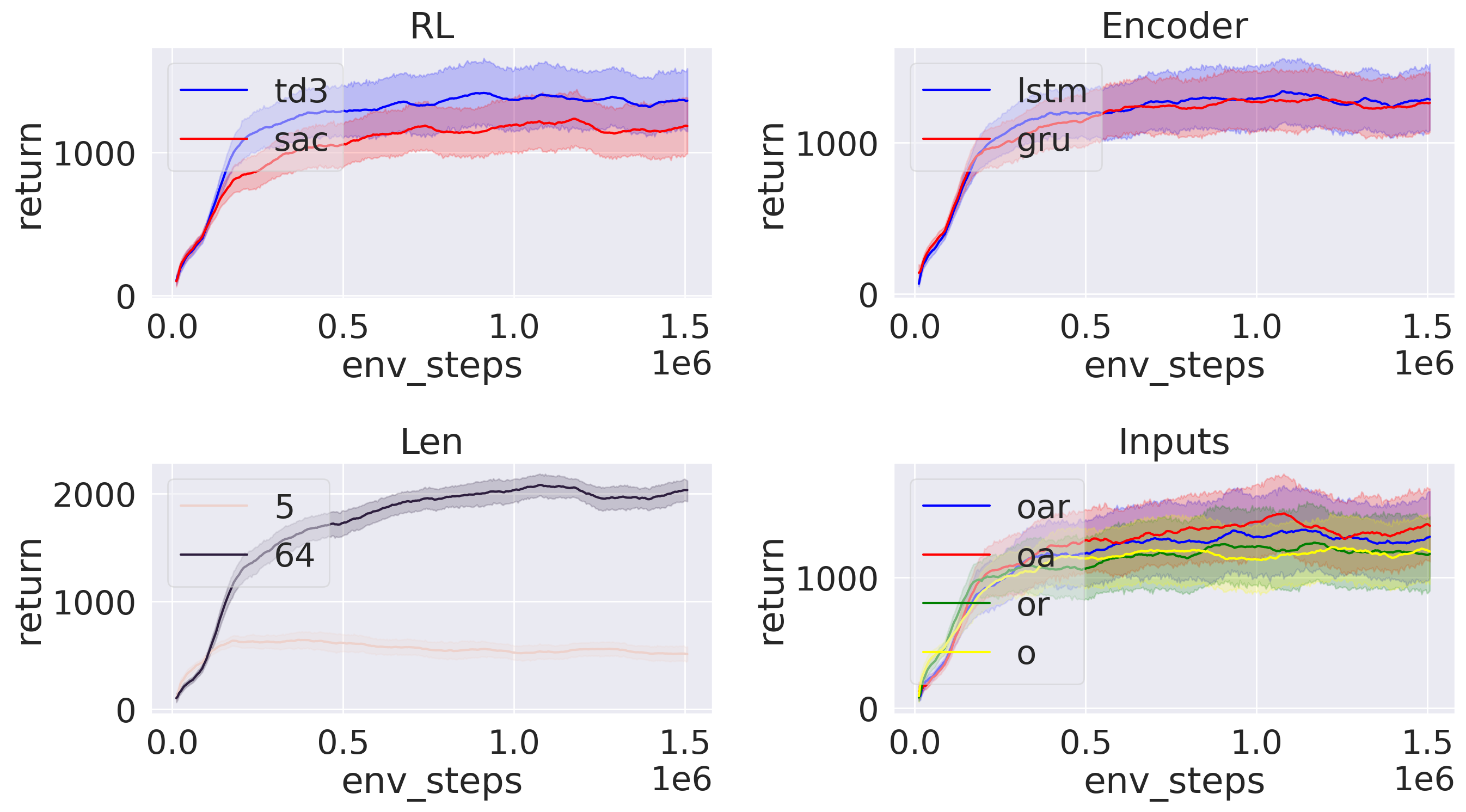} 
    \includegraphics[width=0.49\textwidth]{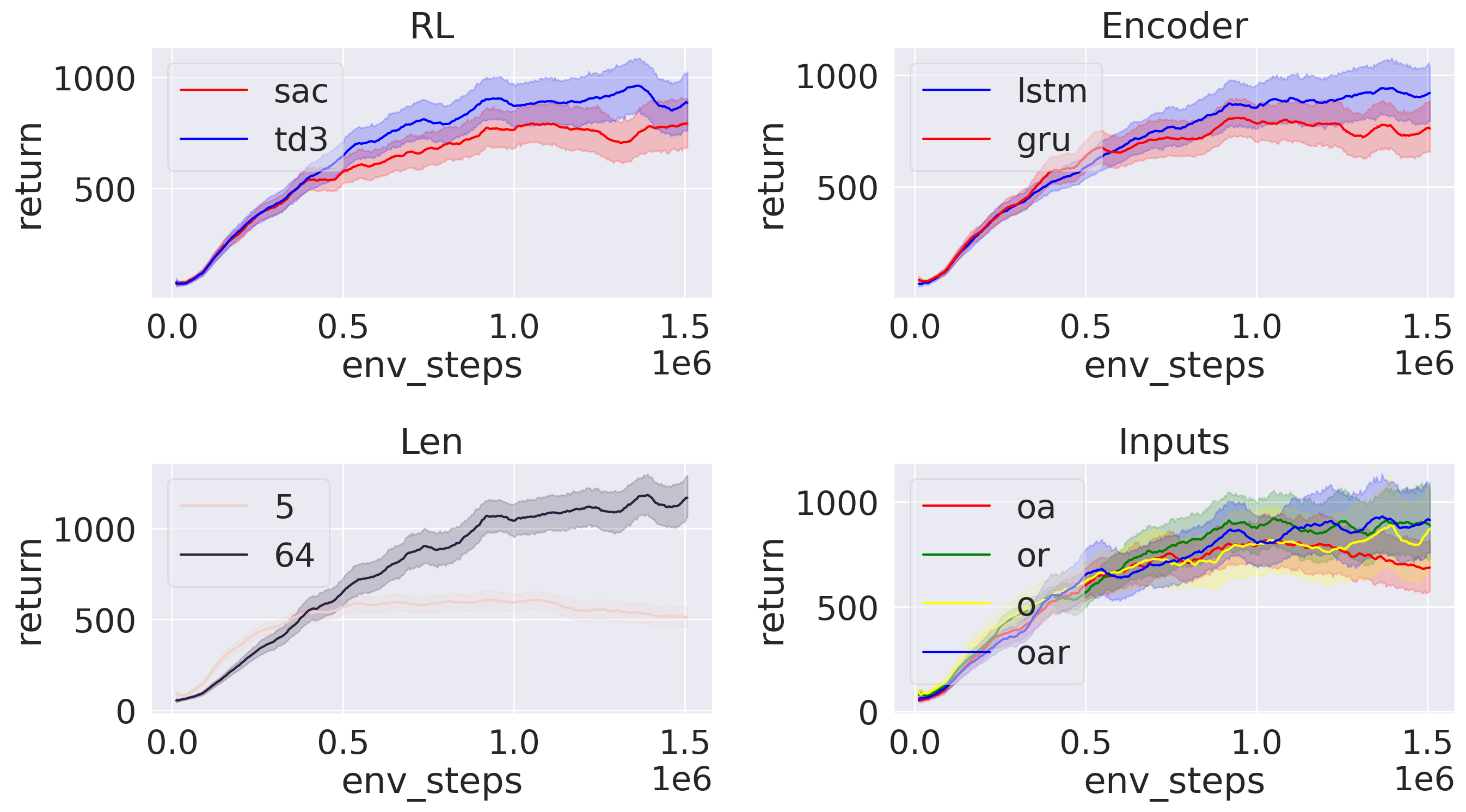} 
    \begin{tabular}{P{0.49\linewidth}P{0.49\linewidth}}
    Hopper-P &
    Walker-P 
    \end{tabular}
    \vspace{-3mm}
    \caption{\textbf{Ablation study of our implementation on ``standard" POMDP benchmark that preserves positions \& angles  but occludes velocities  in the states (namely ``-P").} We show the single factor analysis on the 4 decision factors including RL, Encoder, Len, and Inputs for each environment.}
    \label{fig:pomdp_ablation_P}
\end{figure}

\begin{figure}[h]
    \centering
    \includegraphics[width=0.49\textwidth]{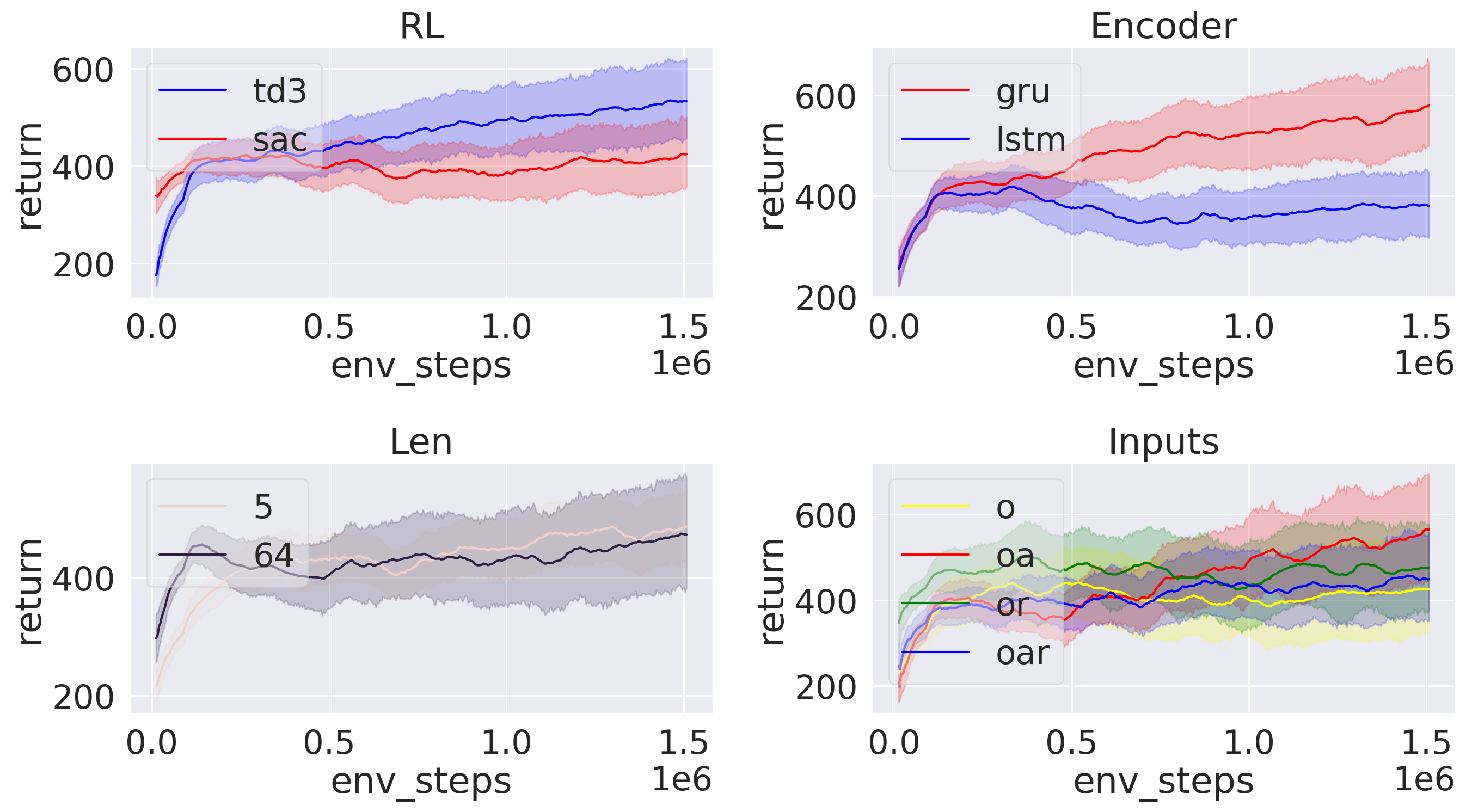} 
    \includegraphics[width=0.49\textwidth]{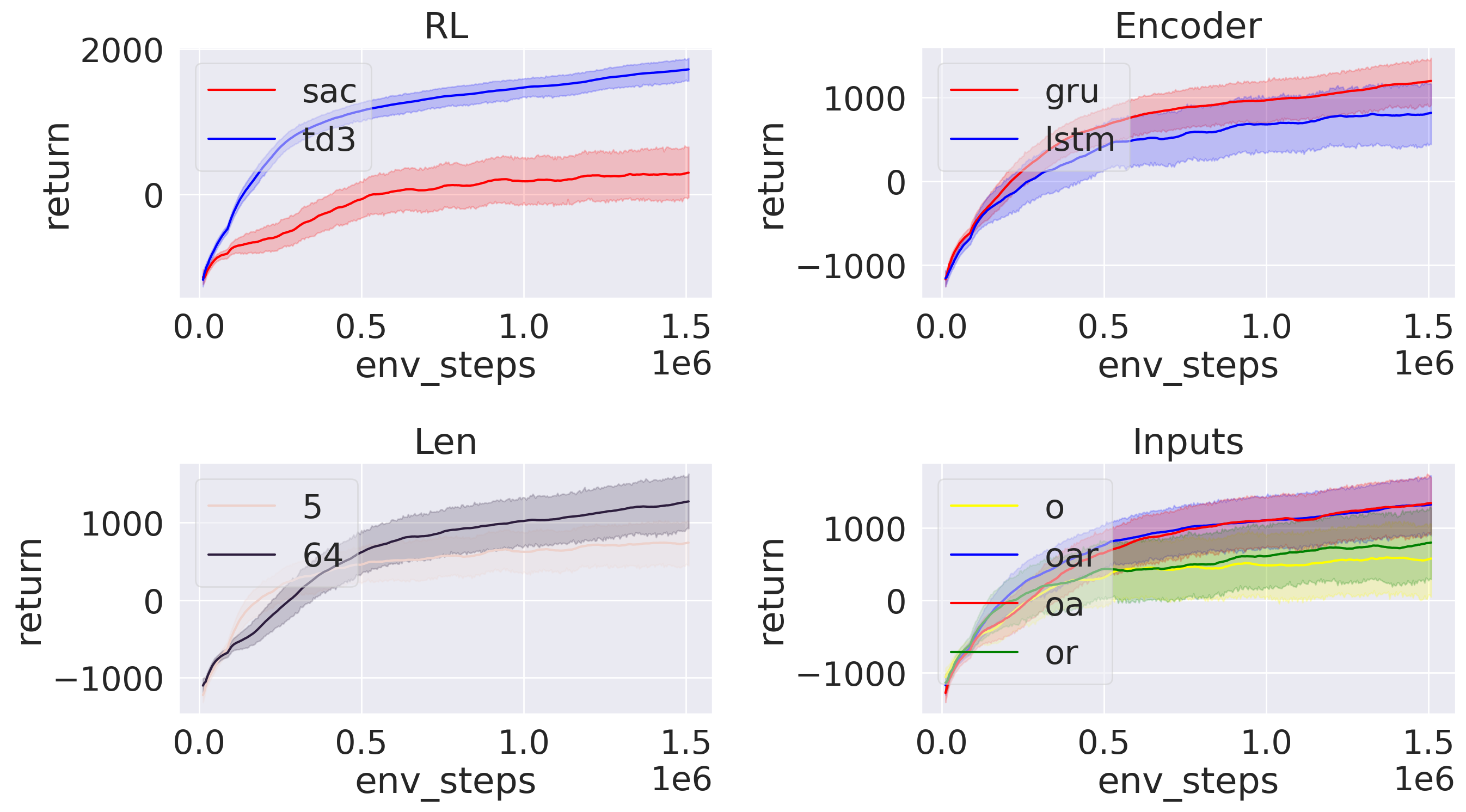} 
    \begin{tabular}{P{0.49\linewidth}P{0.49\linewidth}}
    Ant-V &
    Cheetah-V 
    \end{tabular}
    \includegraphics[width=0.49\textwidth]{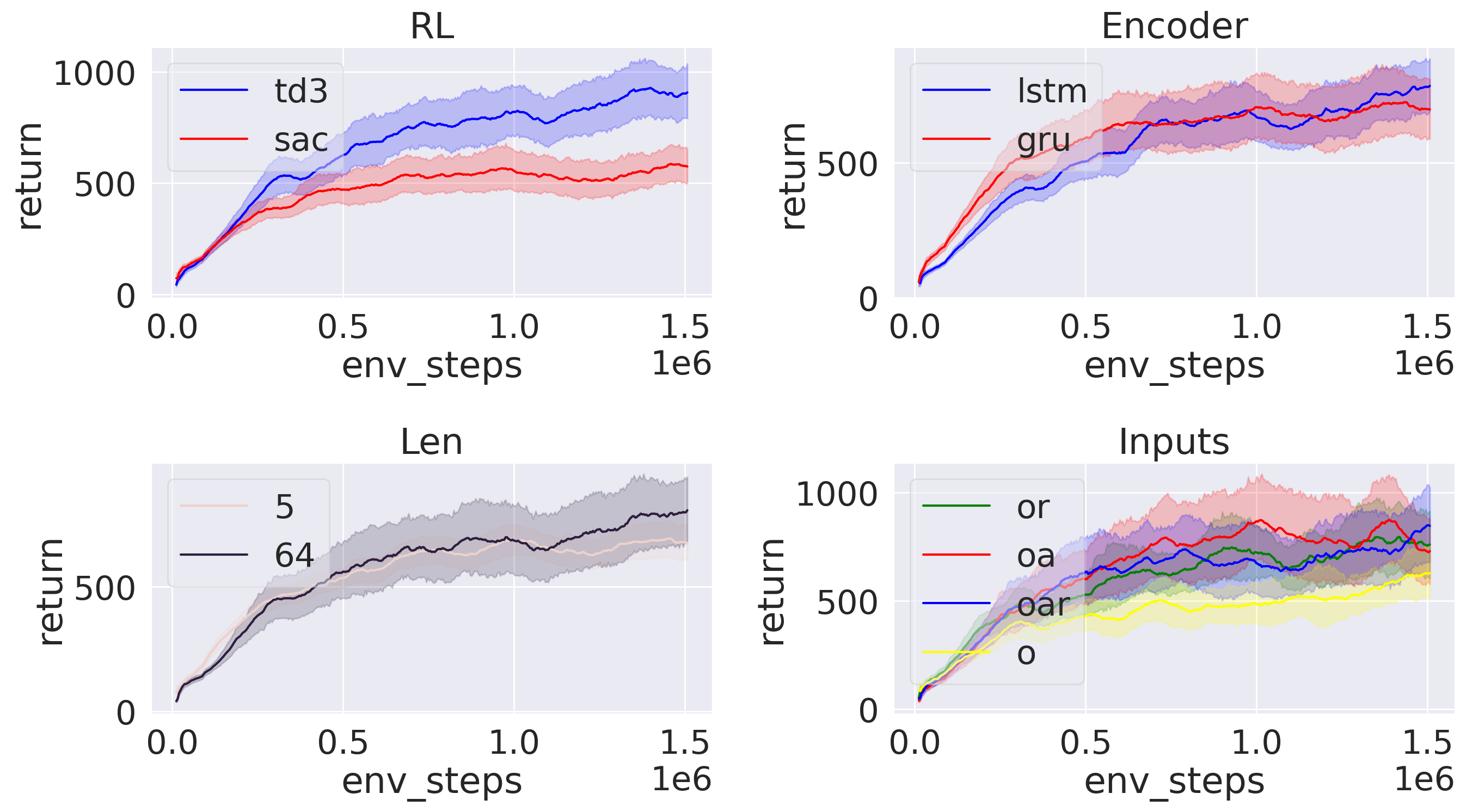} 
    \includegraphics[width=0.49\textwidth]{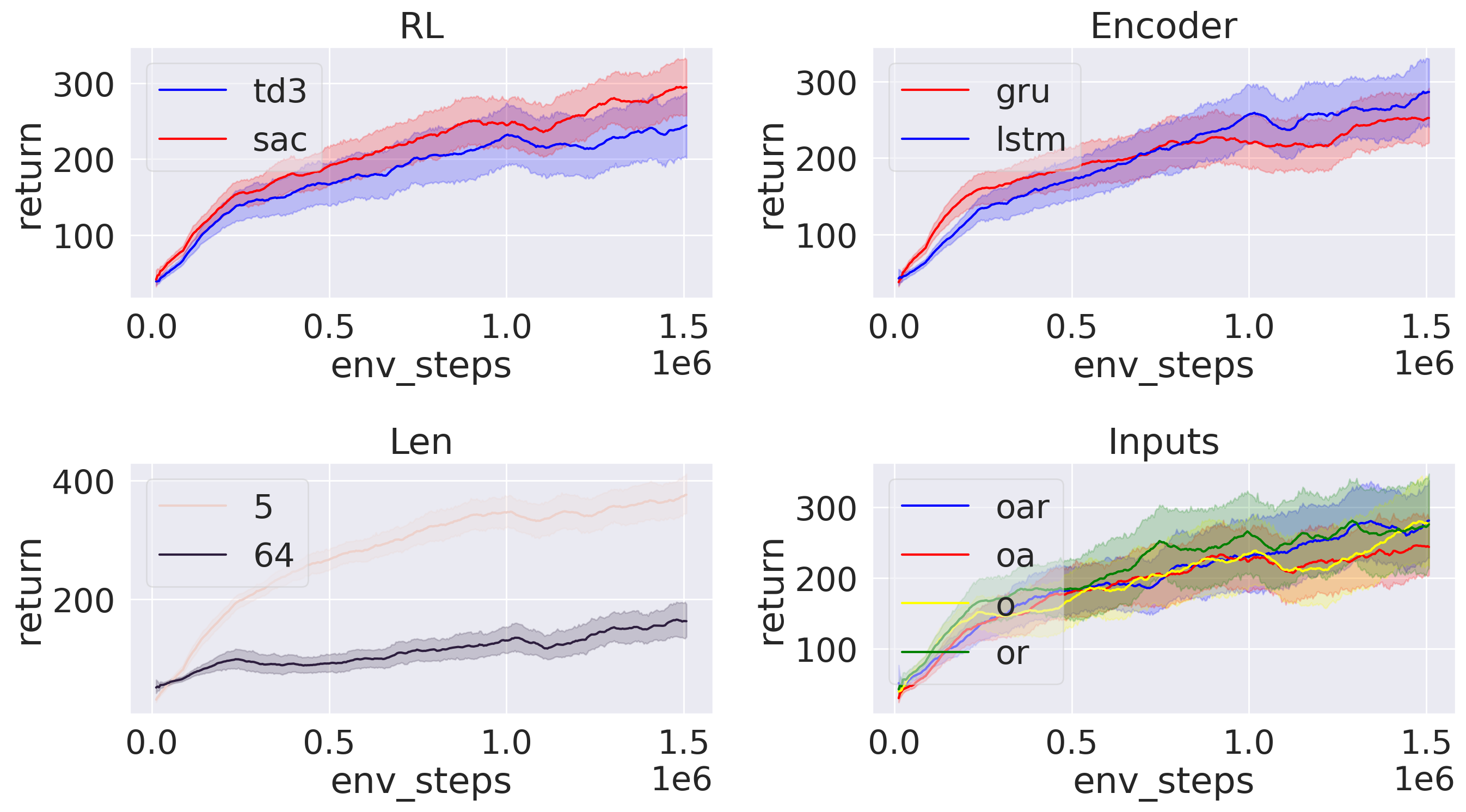}  
    \begin{tabular}{P{0.49\linewidth}P{0.49\linewidth}}
    Hopper-V &
    Walker-V 
    \end{tabular}
    \vspace{-3mm}
    \caption{\textbf{Ablation study of our implementation on ``standard" POMDP benchmark that preserves velocities but occludes positions \& angles  in the states (namely ``-V").} We show the single factor analysis on the 4 decision factors including RL, Encoder, Len, and Inputs for each environment.}
    \label{fig:pomdp_ablation_V}
\end{figure}

\begin{figure}[h]
    \centering
    \includegraphics[width=0.7\textwidth]{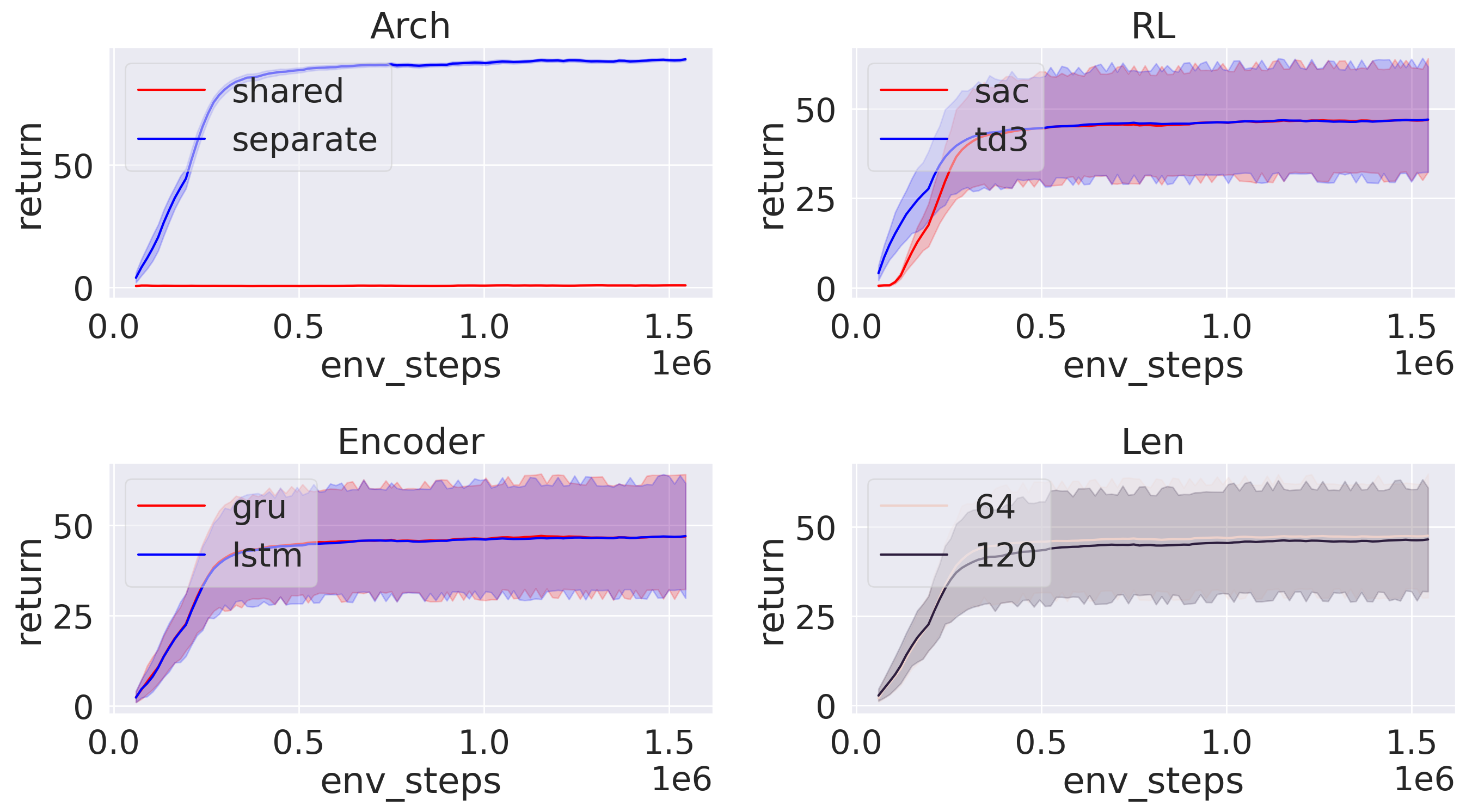} \\ Semi-Circle \\ 
    \includegraphics[width=0.7\textwidth]{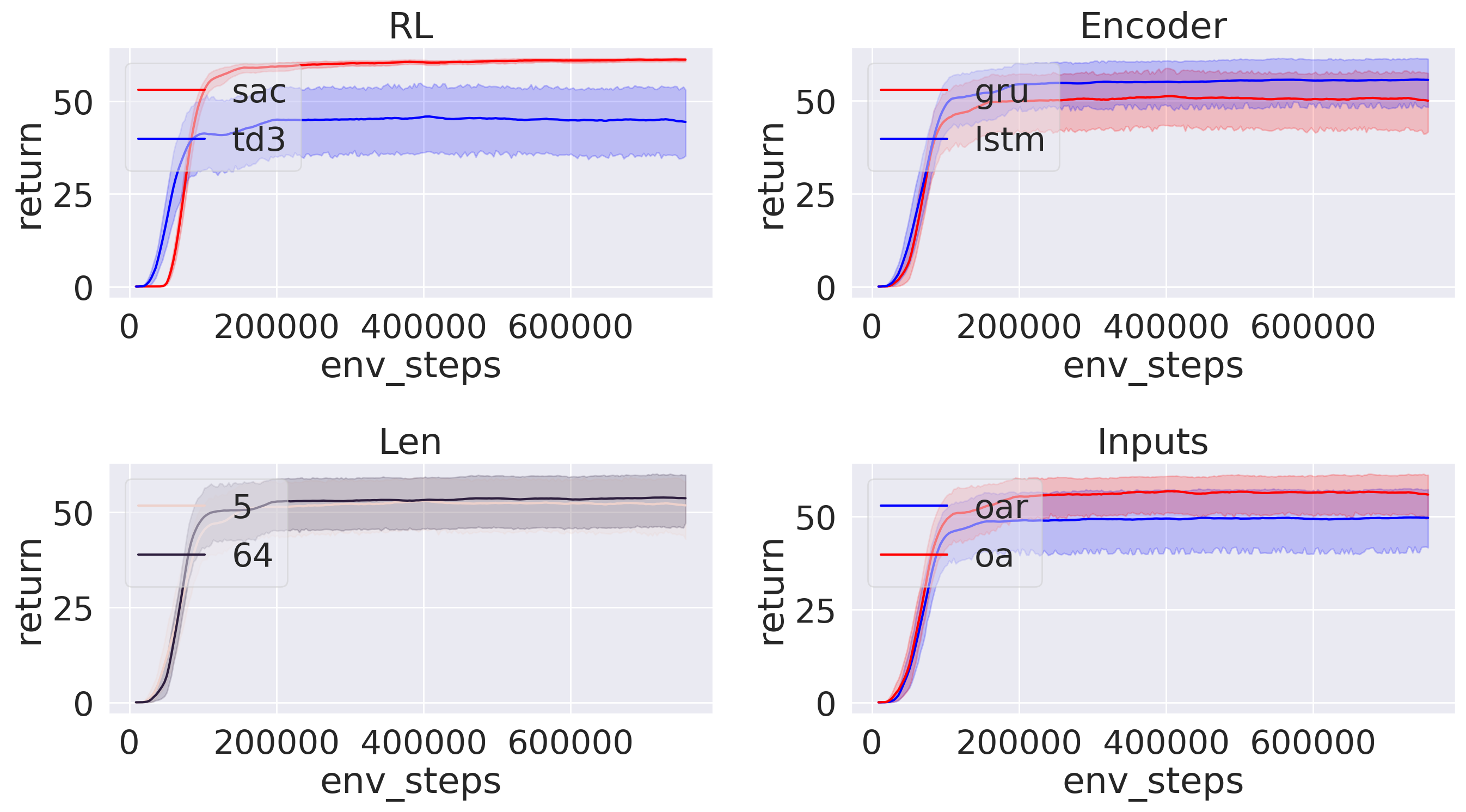} \\ Wind \\
    \includegraphics[width=0.7\textwidth]{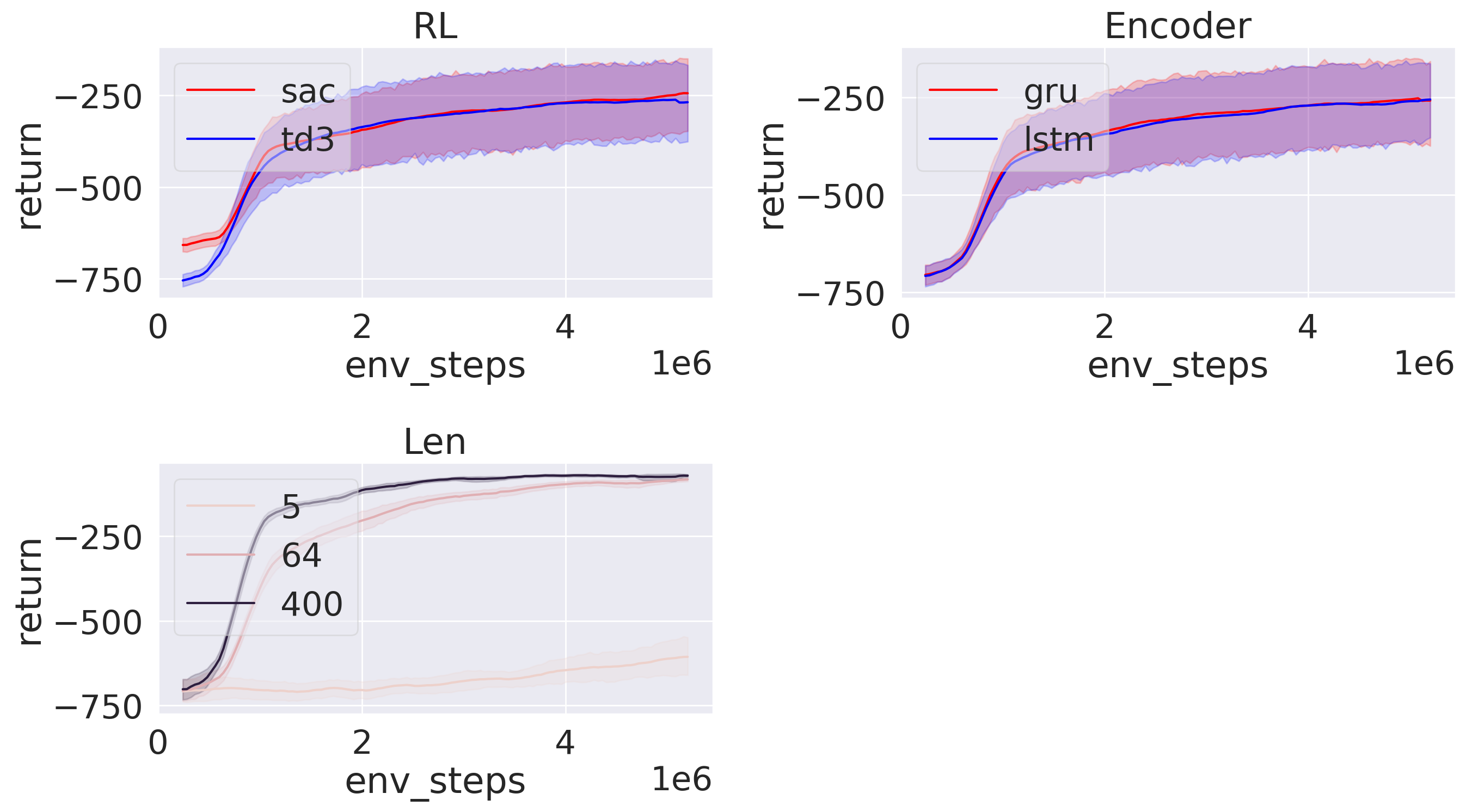} \\  Cheetah-Vel \\
    \caption{\textbf{Ablation study of our implementation on meta-RL benchmark from off-policy variBAD.} We show the single factor analysis on covering all the decision factors.}
    \label{fig:meta_all}
\end{figure}

From these plots, we can see that each decision factor can make a difference in some environments. For example, the choice of RL algorithm is crucial in Ant-P (Fig.~\ref{fig:pomdp_ablation_P}), Cheetah-V (Fig.~\ref{fig:pomdp_ablation_V}), Wind (Fig.~\ref{fig:meta_all}) and Hopper-Generalize (Fig.~\ref{fig:generalize_all}). The context length is essential in all the ``-P" environments (Fig.~\ref{fig:pomdp_ablation_P}), Cheetah-Vel (Fig.~\ref{fig:meta_all}),  and both the generalization environments (Fig.~\ref{fig:generalize_all}). The agent input space can make a difference in most ``-P" environments (Fig.~\ref{fig:pomdp_ablation_P}) possibly because \texttt{oar} contains the information of missing velocities.

\subsection{Additional Results on Separate vs Shared Recurrent Actor-Critic Architecture}
\label{sec:results_arch}

Now we show the result in another POMDP environment, Pendulum-V, which occludes the positions and angles, in Fig.~\ref{fig:separate_vs_shared}. We can see that the shared encoder architecture is also worse than the separate one, possibly due to the different gradient scales in actor and critic losses \wrt the encoder.

\subsection{Additional Results on Comparison with VRM}

Both Fig.~\ref{fig:trailer} and Fig.~\ref{fig:pomdp_bar} shows the final performance of the same single variant of our implementation, but the former shows our results with 1.5M simulation steps while the latter shows our results with 0.5M simulation steps to match with those of VRM due to the time budget.

\begin{figure}[h]
    \centering
    \includegraphics[width=0.7\textwidth]{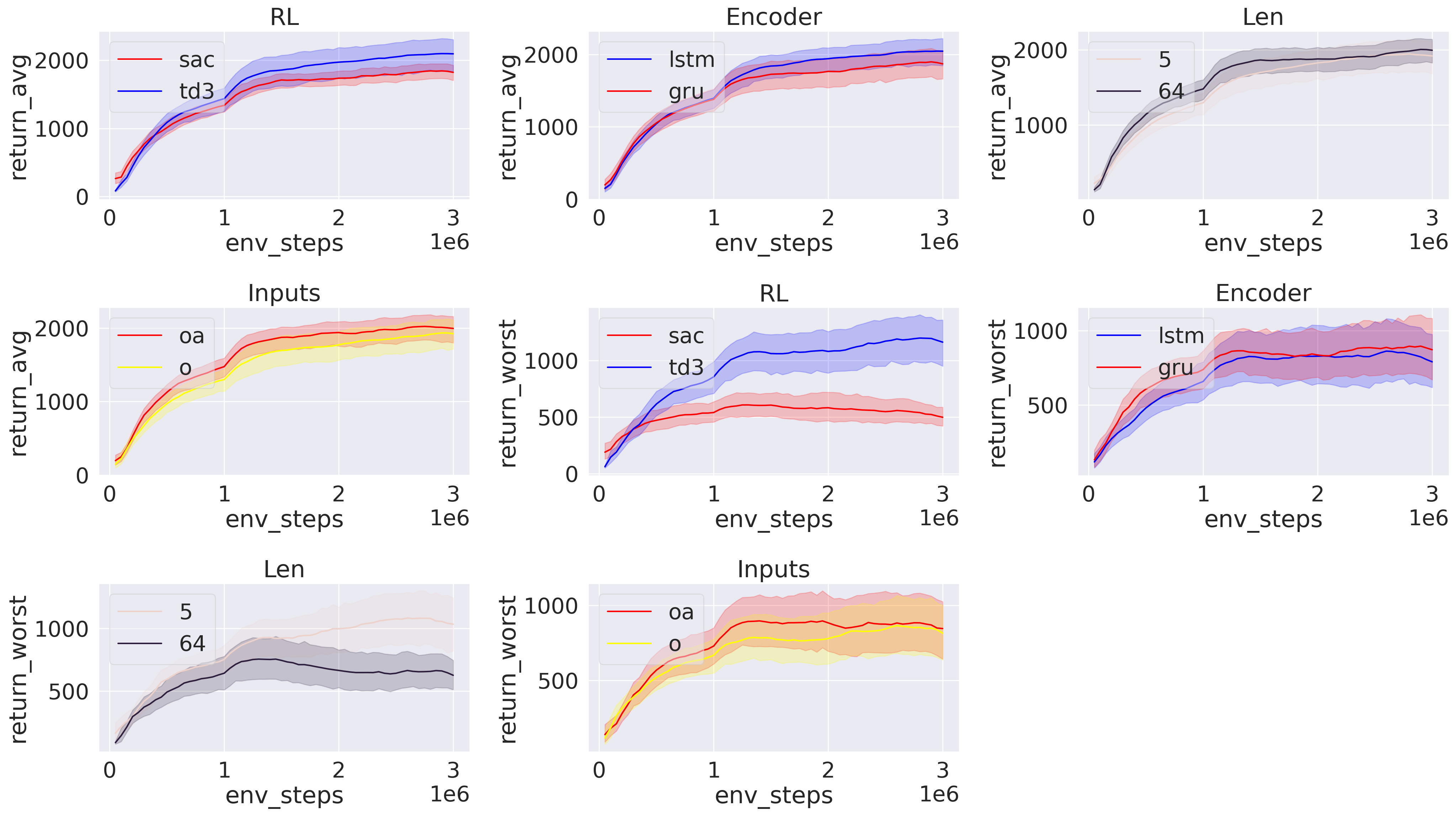} \\ Hopper-Robust \\
    \includegraphics[width=0.7\textwidth]{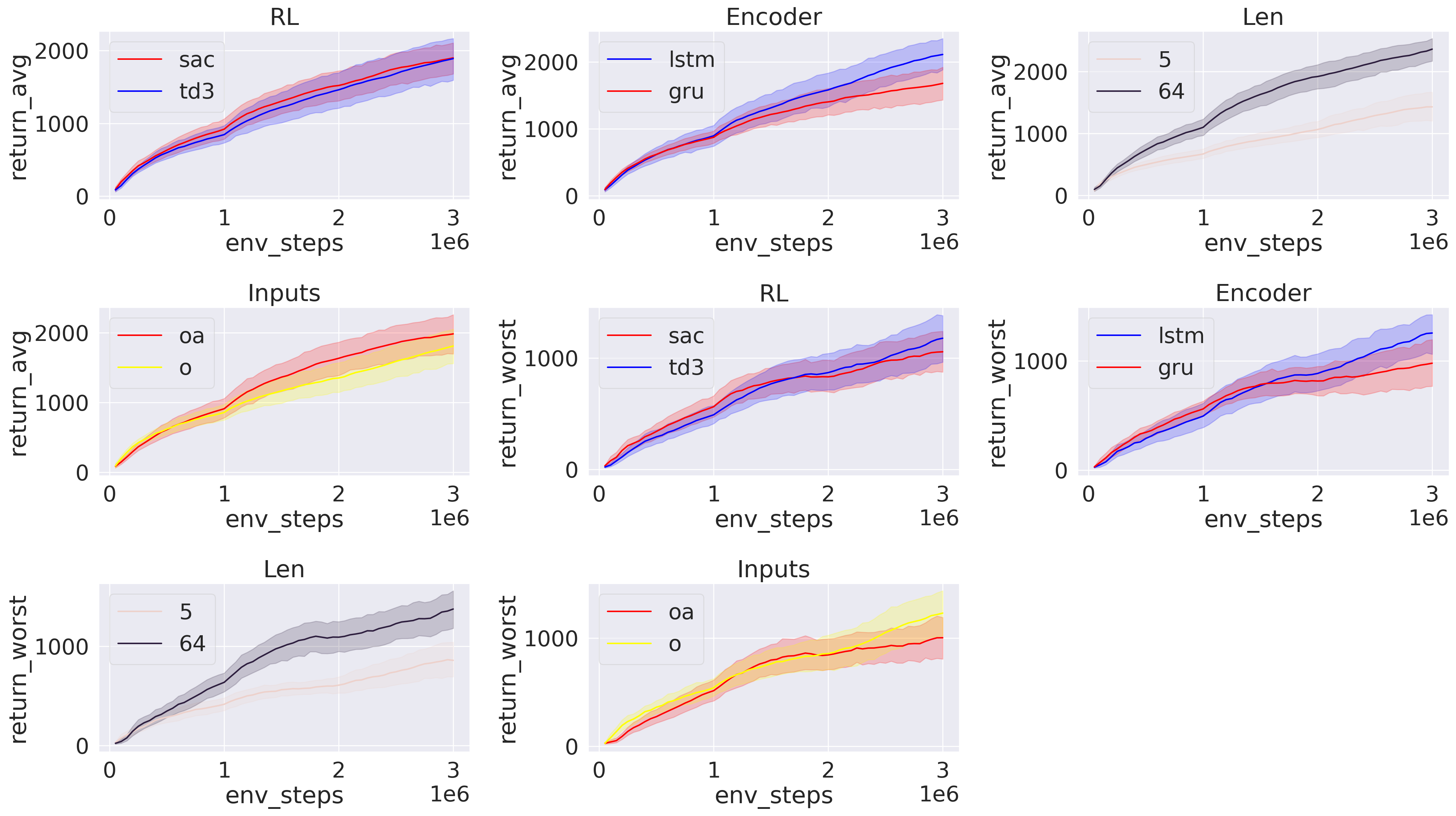} \\ Cheetah-Robust \\
    \includegraphics[width=0.7\textwidth]{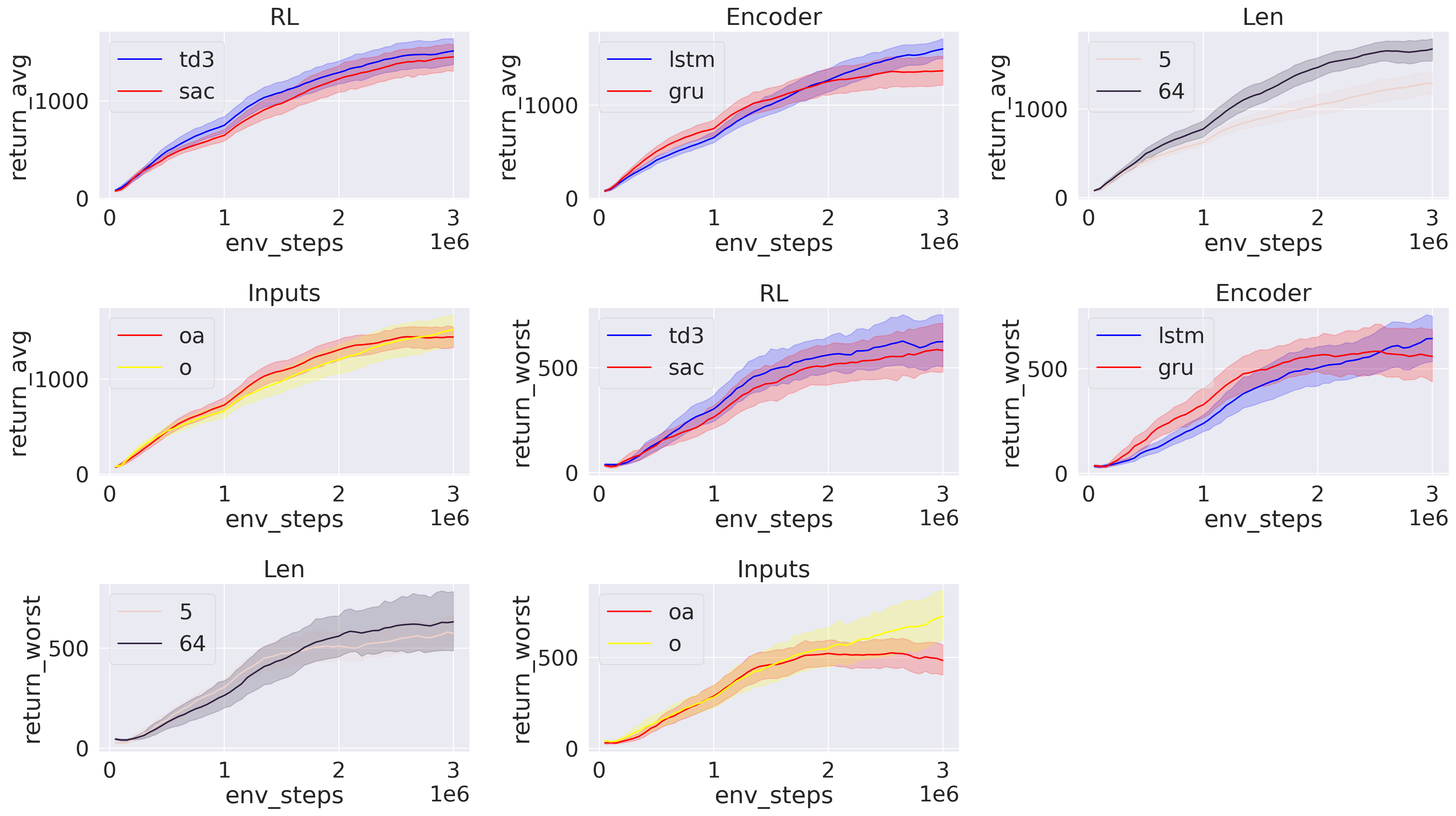} \\ Walker-Robust \\
    \vspace{-1mm}
    \caption{\textbf{Ablation study of our implementation on robust RL benchmark.} We show the single factor analysis on the 4 decision factors including RL, Encoder, Len, and Inputs for each environment in both average returns and worst returns.}
    \label{fig:rmdp_all}
\end{figure}

\begin{figure}[h]
    \centering
    \includegraphics[width=\textwidth]{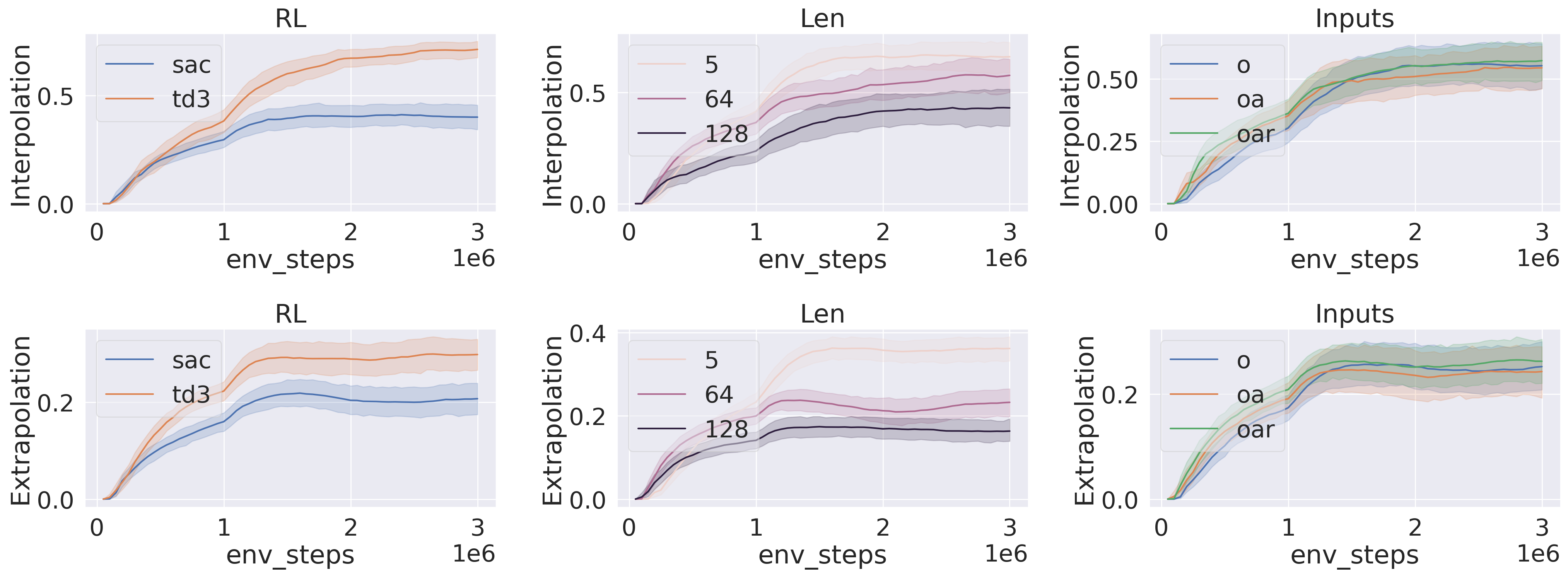} \\ Hopper-Generalize \\
    \includegraphics[width=\textwidth]{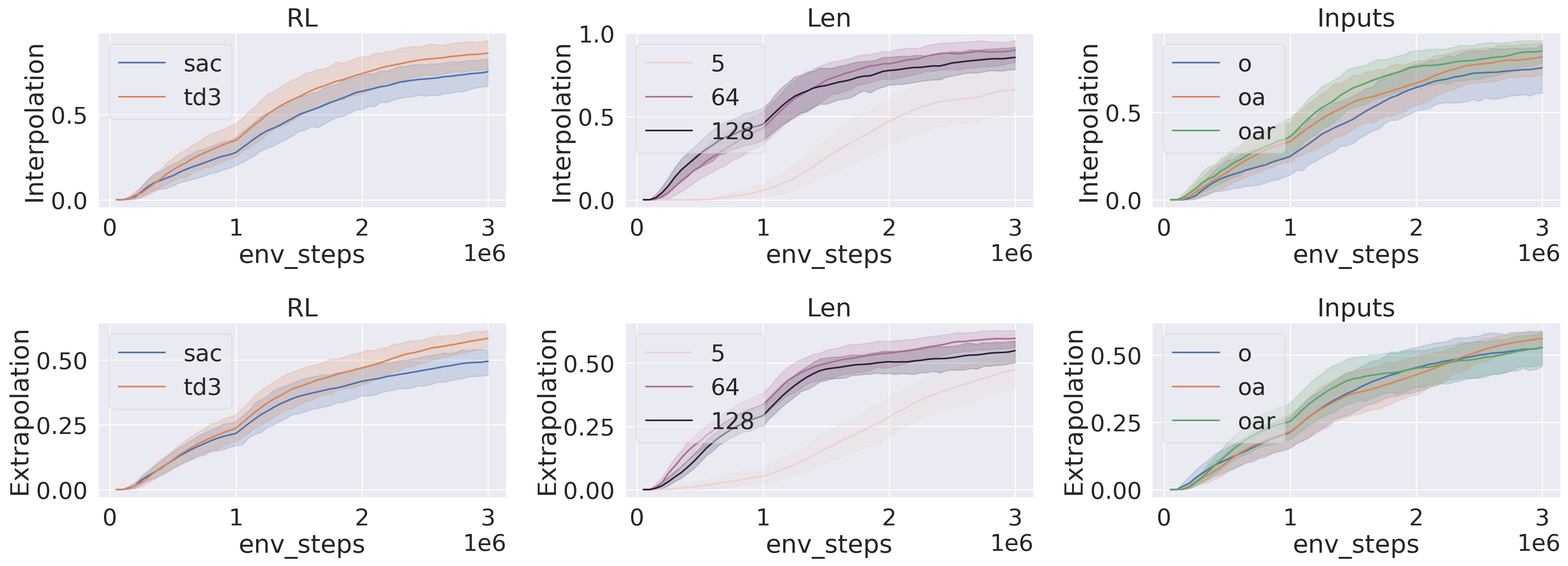} \\ Cheetah-Generalize \\
    \caption{\textbf{Ablation study of our implementation on generalization in RL benchmark.} We show the single factor analysis on the 3 decision factors including RL, Len, and Inputs for each environment in both interpolation and extraploation success rates.}
    \label{fig:generalize_all}
\end{figure}

\begin{figure}[h]
\centering
    \includegraphics[width=\textwidth]{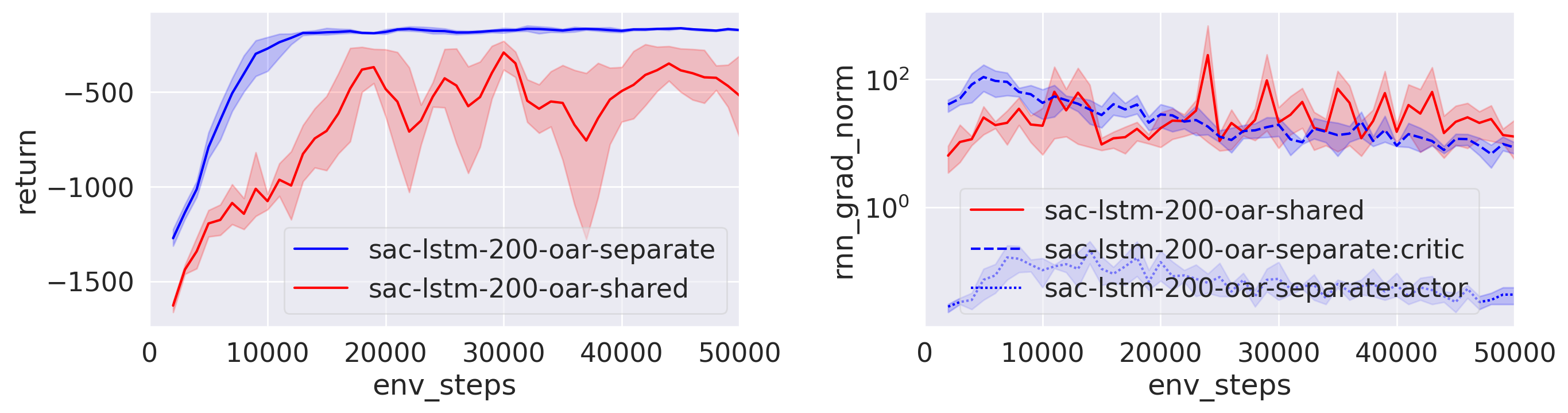}
    \vspace{-3mm}
    \caption{\textbf{Comparison between \textit{shared} and \textit{separate} recurrent actor-critic architecture} with all the other hyperparameters same, on Pendulum-V, a simple ``standard" POMDP environment. We show the performance metric (left) and also the average squared $\ell_2$-norm of the gradient \wrt RNN encoder(s) (right, in \textbf{log-scale}). 
    For the separate architecture, \texttt{:critic} and \texttt{:actor} refer to the separate RNN in critic and actor networks, respectively.}
    \label{fig:separate_vs_shared}
\end{figure}

\begin{figure}[h]
    \centering
    \includegraphics[width=0.8\linewidth]{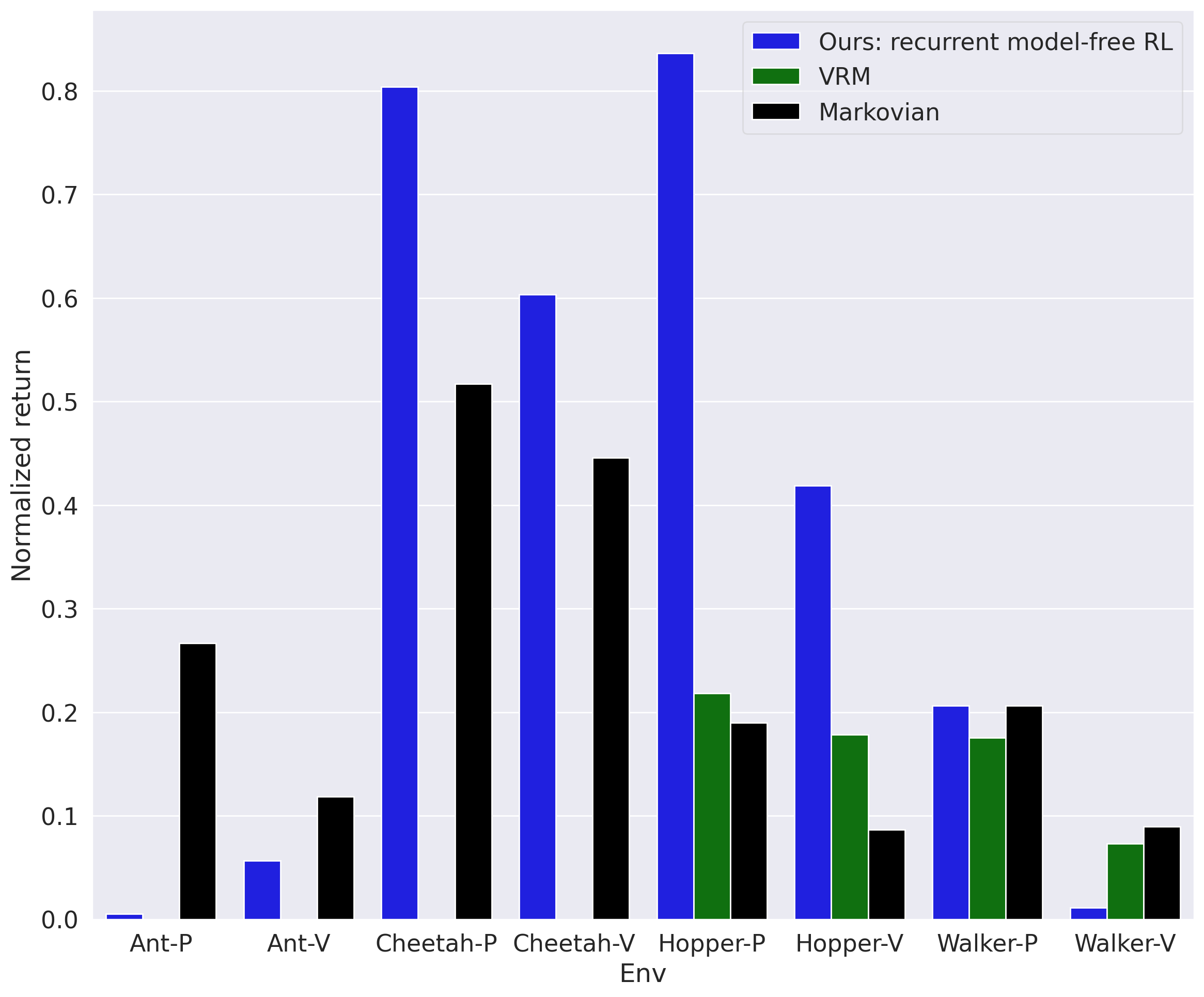}
    \vspace{-3mm}
    \caption{Final \textbf{normalized returns} of our implemented recurrent model-free RL algorithm with the same hyperparameters, and the prior method VRM~\cite{han2019variational} across the eight environments in \textbf{``standard" POMDPs}, each of which trained in 0.5M simulation steps. Our implementation surpasses the specialized method VRM \textbf{on 7 out of 8 environments}. 
    In the figure, we also show Markovian policies as lower bounds for reference, and the y-axis is normalized return given the return of oracle policy from~\citet{raffin2021smooth}.}
    \label{fig:pomdp_bar}
\end{figure}

\end{document}